\documentclass[11pt]{book}
\usepackage{setspace}
\usepackage{amsmath,amssymb}
\usepackage{amsthm}
\usepackage{algorithm, algpseudocode}
\usepackage{url}
\usepackage{fullpage}
\usepackage{makeidx}
\usepackage{graphicx,float,psfrag,epsfig}
\usepackage{color}
\usepackage{xcolor}
\usepackage{graphicx}
\usepackage{tikz}
\usepackage{bbm}
\usepackage{rotating}
\usepackage[font=small,justification=centering]{caption}
\usepackage[font=footnotesize,justification=centering]{subcaption}

\numberwithin{equation}{section}

\usepackage[mathscr]{euscript}

\usepackage{hyperref}
\hypersetup{
    colorlinks,
    linkcolor={blue!80!black},
    citecolor={green!50!black},
}
\colorlet{linkequation}{blue}

\DeclareSymbolFont{rsfs}{U}{rsfs}{m}{n}
\DeclareSymbolFontAlphabet{\mathscrsfs}{rsfs}

\newtheoremstyle{myexample} 
    {\topsep}                    
    {\topsep}                    
    {\rm\small }                   
    {}                           
    {\bf }                   
    {.}                          
    {.5em}                       
    {}  

\newtheoremstyle{myremark} 
    {\topsep}                    
    {\topsep}                    
    {\rm}                        
    {}                           
    {\bf}                        
    {.}                          
    {.5em}                       
    {}  

\newtheorem{claim}{Claim}[section]

\newtheorem{assumption}{Assumption}

\newtheorem{theorem}{Theorem}
\newtheorem{proposition}[claim]{Proposition}
\newtheorem{corollary}[claim]{Corollary}
\newtheorem{definition}[claim]{Definition}

\theoremstyle{myremark}
\newtheorem{remark}{Remark}[section]

\theoremstyle{myremark}

\theoremstyle{myexample}

\def\<{\langle}
\def\>{\rangle}

\def\argmin{{\rm argmin}}

\def\snew{\mbox{\tiny\rm new}}
\def\sinit{\mbox{\tiny\rm init}}

\def\sRF{\mbox{\tiny\sf RF}}
\def\sApp{\mbox{\tiny\sf app}}
\def\sNT{\mbox{\tiny\sf NT}}
\def\stL{\mbox{\tiny\sf L}}
\def\He{{\rm He}}
\def\cV{\mathcal{V}}
\def\Kop{\mathbb{K}}
\def\ratio{{\zeta}}
\def\sNL{\mbox{\tiny\rm NL}}
\def\normf{F}
\def\esssup{{\rm esssup}}
\def\bTheta{{\boldsymbol \Theta}}
\def\hrho{\hat{\rho}}
\def\hbw{\hat{{\boldsymbol w}}}

\def\Trace{{\sf Tr}}

\def\eps{{\varepsilon}}

\def\id{{\boldsymbol I}}

\def\sT{{\sf T}}

\def\cH{{\cal H}}

\def\bLambda{{\boldsymbol \Lambda}}

\def\Z{{\mathbb{Z}}}

\def\S{\mathbb{S}}

\def\normal{{\sf N}}

\def\cA{{\cal A}}

\def\Var{{\sf Var}}

\def\Tr{{\rm {Tr}}}

\def\de{{\rm d}}

\def\Var{{\sf Var}}

\newcommand\norm[1]{\lVert{#1}\rVert}

\def\Z{{\mathcal Z}}

\def\bfzero{{\bf 0}}

\def\bb{{\boldsymbol b}}
\def\br{{\boldsymbol r}}
\def\hbb{\hat{\boldsymbol b}}
\def\hbbeta{\hat{\boldsymbol \beta}}
\def\bbeta{{\boldsymbol \beta}}
\def\cX{{\mathcal X}}

\def\cF{{\cal F}}

\def\bG{{\bf G}}

\def\Var{{\rm Var}}

\def\sB{{\sf B}}

\def\hf{\hat{f}}
\def\err{{\sf err}}
\newcommand{\constantx}{\kappa_{\bx}}

\def\naturals{\mathbb{N}}

\def\reals{\mathbb{R}}

\def\ob{{\overline{b}}}

\def\sQ{{\sf Q}}
\def\orho{\overline{\rho}}
\def\cB{{\mathcal{B}}}

\def\bA{{\boldsymbol A}}

\def\bB{{\boldsymbol B}}

\def\bG{{\boldsymbol G}}

\def\bI{{\boldsymbol I}}
\def\bR{{\boldsymbol R}}
\def\bW{{\boldsymbol W}}
\def\bU{{\boldsymbol U}}

\def\bD{{\boldsymbol D}}

\def\bX{{\boldsymbol X}}
\def\bY{{\boldsymbol Y}}
\def\bZ{{\boldsymbol Z}}

\def\bK{{\boldsymbol K}}
\def\bff{{\boldsymbol f}}

\def\ba{{\boldsymbol a}}
\def\beps{{\boldsymbol \varepsilon}}

\def\bfe{{\boldsymbol e}}
\def\bu{{\boldsymbol u}}

\def\bx{{\boldsymbol x}}

\def\by{{\boldsymbol y}}
\def\tby{\tilde{\boldsymbol y}}
\def\bz{{\boldsymbol z}}

\def\bSigma{{\boldsymbol \Sigma}}
\def\bLambda{{\boldsymbol \Lambda}}

\def\btheta{{\boldsymbol \theta}}

\def\bdelta{{\boldsymbol \delta}}

\def\bv{{\boldsymbol v}}

\def\bF{{\boldsymbol F}}

\def\bw{{\boldsymbol w}}
\def\bS{{\boldsymbol S}}

\def\bPhi{{\boldsymbol \Phi}}
\def\bphi{{\boldsymbol \phi}}
\def\bpsi{{\boldsymbol \psi}}

\def\hbtheta{\hat{\boldsymbol \theta}}
\def\prob{{\mathbb P}}
\def\E{{\mathbb E}}

\def\Treg[#1]{T^{{\rm reg},#1}}
\def\GW[#1]{{\rm GW}(#1)}
\def\MGW[#1]{{\rm MGW}(#1)}

\def\P{{\rm P}}

\def\sexc{{\mbox{\rm\tiny exc}}}
\def\slin{{\mbox{\rm\tiny lin}}}
\def\RF{{\mbox{\sf\tiny RF}}}
\def\NT{{\mbox{\sf\tiny NT}}}
\def\KRR{{\mbox{\sf\tiny KRR}}}
\def\PRR{{\mbox{\sf\tiny PRR}}}

\def\slat{{\mbox{\rm\tiny lat}}}
\def\Unif{{\sf Unif}}

\def\dim{{\rm dim}}

\def\diag{{\rm diag}}

\def\rP{{\rm P}}
\def\op{{\rm op}}
\def\ERM{{\rm ERM}}

\def\cuB{\mathscrsfs{B}}
\def\cuV{\mathscrsfs{V}}
\def\cuP{\mathscrsfs{P}}

\def\cuE{\mathscrsfs{E}}

\def\hR{\hat{R}}
\def\oR{\bar{R}}

\def\Bias{{\sf B}}
\def\Variance{{\sf V}}
\def\proj{{\sf P}}

\newcommand{\constantsig}{\mathsf{d}_{\bSigma}}
\newcommand{\prn}[1]{\left({#1}\right)} 

\title{Six Lectures on\\ Linearized Neural Networks} 
\author{Theodor Misiakiewicz\thanks{Department of Statistics, Stanford University} and Andrea~Montanari\thanks{Department of Electrical 
Engineering and Department of Statistics, Stanford University}}

\makeindex

\begin{document}

\frontmatter

\maketitle
\tableofcontents

\vfill

\section*{Acknowledgments}

These lecture notes are based on two series of lectures given by Andrea Montanari,
 first at the ``Deep Learning Theory Summer School’’ at Princeton from July 27th to August 4th 2021, 
 and then at the summer school ``Statistical Physics \& Machine Learning’’, that took place at Les Houches School of Physics 
 in France from  4th to 29th July 2022.  Theodor Misiakiewicz was Teaching Assistant to 
 AM’s lectures at the ``Deep Learning Theory Summer School’’. We thank Boris Hanin for 
 organizing the summer school in Princeton, and Florent Krzakala and Lenka Zdeborov\'a from EPFL for 
 organizing the summer school in Les Houches.

AM was supported by the NSF through award DMS-2031883, the Simons Foundation through
Award 814639 for the Collaboration on the Theoretical Foundations of Deep Learning, the NSF
grant CCF-2006489 and the ONR grant N00014-18-1-2729

\mainmatter

\chapter{Models and motivations}

\label{ch:intro}

This tutorial examines what can be learnt about the behavior of multi-layer neural
networks from the analysis of linear models. While there are important gaps between
neural networks and their linear counterparts, many useful lessons can be learnt by
studying the latter. 

A few preliminary remarks, before diving into the math:
\begin{itemize}
\item We will not assume specific background in machine learning, let alone neural networks. 
On the other hand, we will assume some graduate-level mathematics, in particular probability
theory (however, we will refer to the literature for complete proofs.)
\item Some of the notations that are used throughout the text will be summarized in Appendix 
\ref{app:Notations}.
\item We will keep bibliographic references in the main text to a minimum.
A short guide to the literature is given in Appendix \ref{app:Bibliography}.
\end{itemize}

This chapter is devoted to describing the correspondence between nonlinear and linear
models via the so-called neural tangent model.

\section{Setting}

We will focus on supervised learning. We are given data  $\{(y_i,\bx_i)\}_{i\le n}\sim_{iid}\prob$
where $\bx_i\in\reals^d$ is a vector of covariates, $y_i\in\reals$ is a response
or label, and $\prob$ a probability distribution on $\reals\times\reals^d$.
We denote the space of such probability distributions by $\cuP(\reals\times\reals^d)$.

We want to learn a model, that is a function $f:\reals^d\to \reals$ that,
given a new covariate vector $\bx_{\snew}$ allows to predict the corresponding response
$y_{\snew}$.  The quality of a prediction is measured by the test error 
$R(f;\prob) =  \E\big\{\ell(y_{\snew},f(\bx_{\snew}))\big\}$ where
 $\ell:\reals\times\reals\to \reals$ is a loss function. We will focus on the simplest 
 choice for the latter, square loss. Namely
 \begin{align}
 R(f;\prob)=  \E\big\{(y_{\snew}-f(\bx_{\snew}))^2\big\}\, , \;\;\;\;\;\; 
 (y_{\snew},\bx_{\snew})\sim\prob \, .
 \end{align}
 Depending on the context, we will  modify the notation for the arguments of $R$.
 In particular, we will be mostly interested in parametric models 
 $f(\bx)= f(\bx;\btheta)$, where $\btheta\in\reals^p$ is a vector of parameters 
 (e.g. the network weights), and therefore the test error can be thought of as a 
 function of these parameters. With a slight abuse of notation, we might therefore write
 \begin{align}
 R(\btheta;\prob)=  \E\big\{(y_{\snew}-f(\bx_{\snew};\btheta))^2\big\} \, .
 \end{align}
 Often we will drop the argument $\prob$ or replace it by a proxy. Also, it is sometimes
 convenient to subtract from the test error the minimum error achieved by any predictor 
 $f$ (also known as the `Bayes risk'). The result is the so called `excess risk'
 \begin{align}
 R_{\sexc}(f;\prob)&= R(f;\prob) - R_{\sB}(\prob)\label{eq:Excess}\\
&:= \E\big\{(y_{\snew}-f(\bx_{\snew}))^2\big\} -\inf_{f_0:\reals^d\to\reals}
 \E\big\{(y_{\snew}-f_0(\bx_{\snew}))^2\big\}\nonumber
\, ,
 \end{align}
 where the infimum is taken over all measurable functions. In the case of
 square loss treated here, the Bayes risk is just the conditional variance:
 $R_{\sB}(\prob) = \E\{(y-\E(y|\bx))^2\}$. 
 
 The main approach to learn the parametric model $f(\,\cdot\,;\btheta)$
 consists in minimizing the empirical risk 
 \begin{align}
 \hR_n(\btheta):=  \frac{1}{n}\sum_{i=1}^n\big(y_i-f(\bx_i;\btheta)\big)^2\, .
 \label{eq:EmpiricalRisk}
 \end{align}
 Modern machine learning systems attempt to achieve approximate minimization 
 of this objective via  first order methods.
 This term is used to refer to algorithms that access the cost function $\hR_n(\btheta)$
 only by obtaining its gradient $\nabla \hR_n(\btheta^i)$ and value $\hR_n(\btheta^i)$
 at query points $\btheta^0,\dots ,\btheta^k$. The next query point $\btheta^{k+1}$
 is computed from this information.
 
 The simplest example is, of course, gradient descent (GD):
 \begin{align*}
  \btheta^{k+1} = \btheta^k - \eps_k \bS\nabla_{\btheta}\hR_n(\btheta^k)\, .
 \end{align*}
 Here $\bS\in\reals^{p\times p}$ is a scaling matrix that allows us to choose different 
 step sizes for different groups of parameters.
 In practice stochastic gradient descent (SGD) is preferred for a number of reasons. 
 In its simplest implementation SGD takes a gradient step with respect to the loss 
 incurred on a single, randomly chosen, sample
 \begin{align*}
  \btheta^{k+1} = \btheta^k +2 \tilde\eps_k  \big(y_{I(k)}-f(\bx_{I(k)};\btheta^k)\big)
  \bS\nabla_{\btheta}f(\bx_{I(k)};\btheta^k)\, .
 \end{align*}
 Both GD and SGD can sometimes be well approximated by gradient flow (GF) for the
 sake of analysis. GF corresponds to the vanishing stepsize limit, and operates in continuous time
 \begin{align*}
  \dot{\btheta}(t) = -\bS\nabla\hR_n(\btheta(t))\, .
  \end{align*}
  Extra care should be paid ---in general--- when working with such continuous
  time dynamics, as they do not necessarily correspond to practical algorithms.
  However in the case of GD and SGD the correspondence is relatively straightforward:
  these algorithms are often well approximated by GF for reasonable choices of the stepsize.
  (Namely, stepsize that is an inverse polynomial in the dimension $d$.)
  
For future reference, it is useful to note that the empirical risk 
\eqref{eq:EmpiricalRisk} only depends on the model $f(\bx;\btheta)$ through its evaluation 
at the $n$ datapoints:
\begin{align}
f_n(\btheta) = \big(f(\bx_1;\btheta),f(\bx_2;\btheta),\dots,f(\bx_n;\btheta)\big)^{\sT}
\end{align}
This define the \emph{evaluation map} $f_n:\reals^p\to\reals^n$. The empirical risk can 
then be written as
 \begin{align}
 \hR_n(\btheta):=  \frac{1}{n}\big\|\by-f_n(\btheta)\big\|^2_2\, ,
 \end{align}
 with $\by = (y_1,\dots,y_n)^{\sT}$.
 
These lectures are mainly devoted to two-layer (one-hidden layer) networks.
In this case the parametric model reads
\begin{align}
f(\bx;\btheta) = \alpha\sum_{i=1}^Na_i\sigma(\<\bw_i,\bx\>)\, ,\;\;\;\;
\btheta = (a_1,\dots,a_N;\bw_1,\dots,\bw_N)\in\reals^{N(d+1)}\, ,\label{eq:TwoLayer}
\end{align}
 where the activation function $\sigma:\reals\to\reals$ is fixed.
 (The scaling factor $\alpha$ will be useful in the following.)
   
\section{The optimization question}

\begin{figure}
\begin{center}
 \includegraphics[width=7.5cm,angle=0]{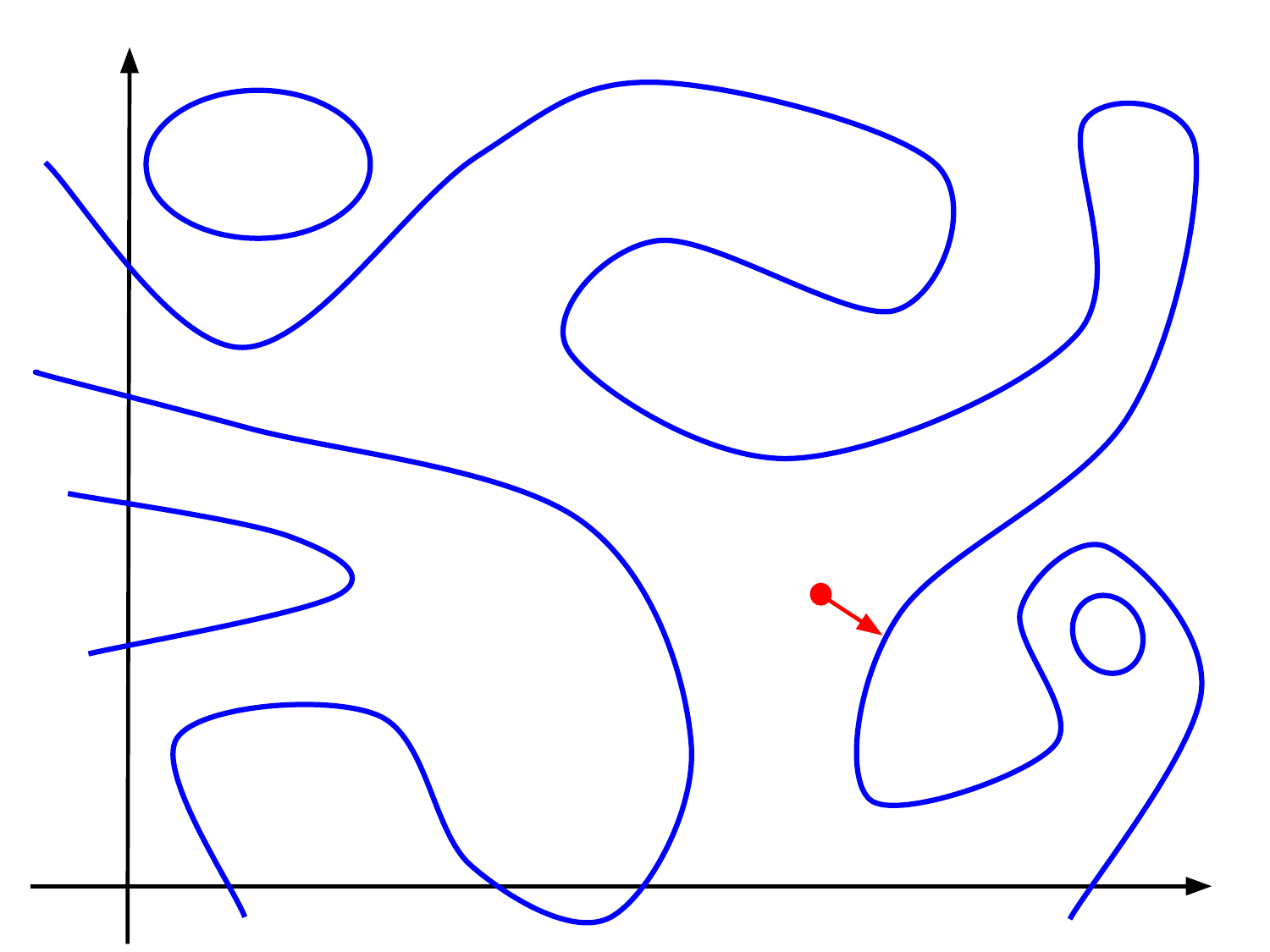}
  \put(-3,7){$\reals^p$}
  \put(-90,60){{\small {\color{red} $\btheta_0$}}}
  \put(-60,45){{\small {\color{red} $\hbtheta$}}}
  \put(-8,100){{\small {\color{blue} $\ERM_0:=\{\btheta:\, \hR_n(\btheta)=0\}$}}}
  \end{center}		
  \caption{}\label{fig:ERM0}
\end{figure}

For nonlinear models such as the two-layer network \eqref{eq:TwoLayer},
the empirical risk $\hR_n(\btheta)$ of Eq.~\eqref{eq:EmpiricalRisk} is highly 
non-convex. Despite this, GD or SGD (and their variants) appear to be able to 
optimize the empirical risk of real neural networks to near global optimality. 

This leads to the following
\begin{quote}
\emph{Optimization question:} How is it possible that simple first order methods
optimize the empirical risk of neural networks to near global optimality?
\end{quote}

Over the last few years, an hypothesis has emerged to explain this puzzle:
tractability
of empirical risk minimization is due to the fact that the network is overparametrized.
Let us briefly describe informally this scenario, which is essentially conjectural at the moment.
Since the number of parameters $p$ is larger than the sample size $n$,
we expect there to be many global empirical risk minimizers that achieve zero risk.
We denote the set of such global minimizers by
\begin{align}
\ERM_0 = \Big\{\btheta\in\reals^p:\; f_n(\btheta) =\by\Big\}\, .
\end{align}
These form a sub-manifold in $\reals^p$, see Fig.~\ref{fig:ERM0} for a cartoon. 
Points of this manifold correspond to models that perfectly interpolate 
the data, i.e. `interpolators'. 

For most initializations $\btheta_0$,  the manifold of interpolators $\ERM_0$
passes close to $\btheta_0$. GD or SGD converge quickly to a specific point on
$\ERM_0$ which is close in a suitable sense.

As emphasized, this scenario is an hypothesis, and we do not know precise
conditions under which it holds. In fact, it is easy to construct counter-examples.
For instance if the function $f(\,\cdot\,;\btheta)$ depends on the parameter $\btheta$ only 
through its first $k$ coordinates $\theta_1,\dots,\theta_k$. Then there might 
not\footnote{The situation is a bit more complicated: if all we are interested in is interpolation, 
then less than $n$ parameters can be sufficient for certain parametric classes $f(\,\cdot\,;\btheta)$.
However, these solutions are typically 
fragile to noise and hard to compute.} be a 
solution to the $n$ equations $\by=f_n(\btheta)$ for $k<n$ (despite $n\le p$).

One setting in which the conditions for interpolation are better understood is
the  `linear' regime which we discuss next.

\section{The linear regime}
\label{sec:linear_regime}

In certain cases, it can be proved that the weights do not change much during 
training, and it is therefore accurate to approximate $f(\,\cdot\,;\btheta)$
by its first order Taylor expansion in $\btheta_0$.
Justifying this approximation is not the main focus of these lectures which instead
take it as a starting point and derive some of its consequences. 

We will outline nevertheless an explanation, deferring to the literature 
for a rigorous treatment (see \cite{du2018gradient,chizat2019lazy,bartlett2021deep}
and Appendix \ref{app:Bibliography}). We are looking for a solution to 
the interpolating equation $f_n(\btheta)=\by$. By Taylor's theorem,
this yields
\begin{align}
\by-f_n(\btheta_0) &=f_n(\btheta)-f_n(\btheta_0)\\
&= \bD f_n(\btheta_0) (\btheta-\btheta_0)+
\int_{0}^1(\bD f_n(\btheta_t)-\bD f_n(\btheta_0)) (\btheta-\btheta_0)\, \de t\\
& =: \bPhi (\btheta-\btheta_0) + \bfe(\btheta)\, ,
\end{align}
where $\btheta_t:=(1-t)\btheta_0+t\btheta$ and
 we denoted by $\bPhi:=\bD f_n(\btheta_0)\in\reals^{n\times p}$ the Jacobian of the evaluation map at 
the initialization $\btheta_0$. Further
\begin{align}
\|\bfe(\btheta)\|_2 \le&~ L_n\|\btheta-\btheta_0\|_2^2\, ,\label{eq:Bound2ndOrder}\\
L_n := &~ \sup_{\btheta\neq \btheta_0}
\frac{\|\bD f_n(\btheta)-\bD f_n(\btheta_0)\|_{\op}}{\|\btheta-\btheta_0\|_2}\, .
\end{align}
Neglecting the second order contribution in the above equation suggests a solution 
of the form (here $\bPhi^+$ denotes the pseudoinverse)
\begin{align}
\btheta=\btheta_0+\bPhi^{+}(\by-f_n(\btheta_0)+\bdelta) \, ,\label{eq:ThetaExpansion}
\end{align}
with the following equation for $\bdelta$ (writing $\tby:=\by-f_n(\btheta_0)$):
\begin{align}
\bdelta = -\bfe(\btheta_0+\bPhi^{+}(\tby+\bdelta))\,.
\end{align}
Defining $\bF(\bdelta):= -\bfe(\btheta_0+\bPhi^{+}(\tby+\bdelta))$,
by Eq.~\eqref{eq:Bound2ndOrder} we have 
$\|\bF(\bdelta)\|_2\le L_n(\|\bPhi^{\dagger}\tby\|_2+\|\bPhi^{+}\|_{\op}\|\bdelta\|_2)^2$.
In other words, $\bF$ maps the ball of radius $r$ into the ball of radius
$L_n(\|\bPhi^{+}\tby\|_2+\|\bPhi^{+}\|_{\op}r)^2$.
By taking $r=\|\bPhi^+\tby\|_2/\|\bPhi^+\|_{\op}$ we obtain that, for 
\begin{align}
L_n\le \frac{1}{4\|\bPhi^+(\by-f_n(\btheta_0))\|_2 \|\bPhi^+ \|_{\op} }\, ,\label{eq:LipschitzCondition}
\end{align}
the ball of radius $r$ is mapped into itself. Hence, by Brouwer's fixed point 
theorem, there exists a solution (an interpolator) of the form 
\eqref{eq:ThetaExpansion} with $\|\bdelta\|_2<r$. 

Summarizing, under condition \eqref{eq:LipschitzCondition}, there exists an interpolator that
is well approximated by replacing the nonlinear model
$f(\,\cdot\,;\btheta)$ by its linearization 
\begin{align}
f_{\slin}(\bx;\btheta) :=f(\bx;\btheta_0)+
\<\btheta-\btheta_0,\nabla_{\btheta}f(\bx;\btheta_0)\>\, .
\end{align}
Indeed, for $\bdelta=0$, the expression in Eq.~\eqref{eq:ThetaExpansion}
coincides with the solution of the linearized equations
\begin{align}
f_{\slin}(\bx_i;\btheta) =y_i\, ,\;\; \forall i\le n\,,
\end{align}
which minimizes the $\ell_2$ distance from initialization.

This suggests to define the following linearized empirical risk
\begin{align}
\hR_{\slin, n}(\btheta) :=&~ \frac{1}{n}\big\|\by-f_{\slin,n}(\btheta)\big\|_2^2\\
= &~ \frac{1}{n}\big\|\tby-\bPhi(\btheta-\btheta_0)\big\|_2^2\, ,
\end{align}
and the corresponding linearized test error:
\begin{align}
R_{\slin}(\btheta) &:=  \E\big\{\big(y_{\snew}-f_{\slin}(\bx_{\snew};\btheta)\big)^2
\}\, .
\end{align}

Several papers prove that, under conditions analogous to \eqref{eq:LipschitzCondition},
the original problem and the linearized one are close to each other, see
Appendix \ref{app:Bibliography}. Informally, denoting by $\btheta(t)$ the gradient flow for $\hR_n$,
and by $\btheta_{\slin}(t)$ the gradient flow for $\hR_{\slin, n}$, these proofs imply the following:
\begin{enumerate}
\item The empirical risk converges exponentially fast to $0$. Namely, for all $t$, we have
$\hR_n(\btheta(t))\le \hR_n(\btheta_0)\, \exp(-\lambda_0t/2)$, where 
$\lambda_0 := \|\bPhi^+\|_{\op}^2=\sigma_{\min}(\bPhi)^2$.
\item The parameters of one model are tracked by the ones of the other model, i.e. 
$\|\btheta(t)-\btheta_{\slin}(t)\|_2\ll \|\btheta(t)-\btheta_0\|_2$ for all $t$.
\item The linearized model is a good approximation for a fully nonlinear model. Namely, 
for all $t$
\begin{align}
\big\|f_{\slin}(\,\cdot\,;\btheta_{\slin}(t))-f(\,\cdot\,;\btheta(t))\big\|_{L^2}
\ll \big\|f(\,\cdot\,;\btheta(t))\big\|_{L^2}\, .\label{eq:LinearizedModel}
\end{align}
(Here $\|g\|_{L^2}:= \E\{g(\bx)^2\}^{1/2}$.)
\end{enumerate}

From a statistical perspective, the most important point is the last one,
cf. Eq.~\eqref{eq:LinearizedModel}, since it says that the model learnt within 
the linear theory is a good approximation of the original nonlinear model
\emph{at a random test point}, rather that at a training point. In particular,
by triangular inequality
\begin{align}
\big|R(\btheta (t) )^{1/2} - R_{\slin}(\btheta_{\slin}(t) )^{1/2}\big|\ll \E\{f(\bx;\btheta(t))^2\}^{1/2}\,.
\end{align}

From here on, we will focus on such linearized models, and try to understand their
generalization error. Notice that the parameter vector $\btheta_{\slin}(t)$ 
depends on time and hence in principle we would like to study the whole 
function $t\mapsto R_{\slin}(\btheta_{\slin}(t))$. Of particular interest is the limit 
$t\to\infty$ (which corresponds to interpolation). It is an elementary fact that gradient 
flow with a quadratic cost function converges to the minimizer that is closest to the 
initialization in $\|\,\cdot\,\|_{\bS^{-1}}$ norm:
\begin{align}
\lim_{t\to\infty}\btheta_{\slin}(t) 
&=\argmin\Big\{\|\btheta-\btheta_0\|_{\bS^{-1}}:\;
\bD f_n(\btheta_0)(\btheta-\btheta_0) = \by-f_n(\btheta_0)\Big\}\\
&=\btheta_0+\argmin\Big\{\|\bb\|_{\bS^{-1}}:\;
\bD f_n(\btheta_0)\bb = \by-f_n(\btheta_0)\Big\}\, ,
\end{align}
where $\|\bb\|_{\bS^{-1}} = \| \bS^{-1/2} \bb \|_2 = \< \bb, \bS^{-1} \bb \>^{1/2}$.

Hereafter, we will typically drop the subscripts `lin'.
Rather than studying the gradient flow path, we will focus on a different path
that also interpolates between the $\btheta_0$ and the min-$\ell_2$
distance interpolator. Namely, we will consider ridge regression
\begin{align}
\hbb(\lambda) := \argmin_{\bb\in\reals^p}
\Big\{\frac{1}{n}\big\|\tby-\bPhi\bb\big\|_2^2+ \lambda\|\bb\|_{\bS^{-1}}^2\Big\}\, ,
\end{align}
where we recall that $\tby = \by-f_n(\btheta_0)$ and $\bPhi = \bD f_n(\btheta_0)$.

\section{Linearization of two-layer networks}
\label{sec:GeneralLinearization}
As mentioned above, we will focus on the case of two-layer neural networks,
cf. Eq.~\eqref{eq:TwoLayer}. We will assume $\sigma$ to be weakly differentiable with 
weak derivative $\sigma'$. A simple calculation yields
\begin{align}
f_{\slin}(\bx;\btheta_0+\bb) &= f(\bx;\btheta_0)+
\alpha\sum_{i=1}^N b_{1,i}\sigma(\<\bw_{i},\bx\>)+\alpha
\sum_{i=1}^Na_{i}\<\bb_{2,i},\bx\>
\sigma'(\<\bw_{i},\bx\>)\, ,\\
\btheta_0 &:=\big(a_{1},\dots,a_N;\bw_{1},\dots,\bw_{N})\in\reals^{N(d+1)}\, ,\\
\bb &:=\big(b_{1,1},\dots,b_{1,N};\bb_{2,1},\dots,\bb_{2,N})\in\reals^{N(d+1)}\, .
\end{align}

We can rewrite this linear model in terms of the following \emph{featurization
maps}:
\begin{align}
\bphi_{\sRF}(\bx) &= \frac{1}{\sqrt{N}} 
[\sigma(\<\bw_{1},\bx\>);\dots; \sigma(\<\bw_{N},\bx\>)]\, ,\\
\bphi_{\sNT}(\bx) &= \frac{1}{\sqrt{Nd}} 
[\sigma'(\<\bw_{1},\bx\>)\bx^{\sT};\dots; \sigma'(\<\bw_{N},\bx\>)\bx^{\sT}]\, , \label{eq:featureMap_NT}
\end{align}
We have $\bphi_{\sRF}:\reals^d\to\reals^N$ and $\bphi_{\sNT}:\reals^d\to\reals^{Nd}$.
Setting for convenience $\alpha = 1/\sqrt{N}$, we get
\begin{align}\label{eq:linearization_RF_NT}
f_{\slin}(\bx;\btheta_0+\bb) &= f(\bx;\btheta_0)+\<\bb_1,\bphi_{\sRF}(\bx)\>
+\sqrt{d}\<\bb_2,\bphi_{\sNT}(\bx)\>\, .
\end{align}

We define the design matrix 
\begin{align}
\bPhi :=\left[\begin{matrix}
      \mbox{---} \bphi_{\sRF}(\bx_1)   \mbox{---}&     \mbox{---}\bphi_{\sNT}(\bx_1)  \mbox{---}\\
      \mbox{---} \bphi_{\sRF}(\bx_2) \mbox{---}&  \mbox{---}\bphi_{\sNT}(\bx_2)  \mbox{---}\\
      \vdots\\
      \mbox{---} \bphi_{\sRF}(\bx_n)\mbox{---}&\mbox{---}\bphi_{\sNT}(\bx_n)  \mbox{---}\\
      \end{matrix}\right] \, ,
\end{align}
and consider the stepsize scaling matrix
\begin{align}
\bS = {\rm diag}(\underbrace{1,\dots,1}_{N};\underbrace{sd,\dots,sd}_{Nd}) \, .
\end{align}
With suitable redefinition of $\bb = (\bb_1,\bb_2)\in\reals^N\times\reals^{Nd}$,
the ridge regression problem thus reads
\begin{align}
\hbb(\lambda) := \argmin_{\bb\in\reals^p}
\Big\{\frac{1}{n}\big\|\tby-\bPhi\bb\big\|_2^2+ \lambda_{\sRF}\|\bb_1\|_2^2+
\lambda_{\sNT}\|\bb_2\|_2^2\Big\}\, ,\label{eq:Ridge2Layer}
\end{align}    
where $\lambda_{\sRF} := \lambda$, $\lambda_{\sNT} := \lambda/s$.
We conclude this section with two remarks.
\begin{remark}\label{rmk:implicit}
We saw that GD with respect to a quadratic cost function converges to the closest minimizer
to the initialization, where `closest' is measured by a suitably weighted $\ell_2$ distance.
This is the simplest example of a more general phenomenon known as `implicit regularization':
when learning overparametrized models, common optimization algorithms select empirical risk 
minimizers that present special simplicity properties (smallness of certain norms). 
While this is normally achieved by explicitly regularizing the risk to promote those properties,
in overparametrized system it is implicitly induced (to some extent) by the dynamics of the optimization 
algorithm. Examples of this phenomenon include \cite{gunasekar2017implicit,gunasekar2018characterizing,soudry2018implicit,ji2018risk} (see also Appendix \ref{app:Bibliography}).

Note that Eq.~\eqref{eq:Ridge2Layer} also illustrates how the precise form of implicit regularization
depends on the optimization algorithm. We mentioned that gradient flow
converges to the $\lambda=0+$ solution of  Eq.~\eqref{eq:Ridge2Layer}, 
which corresponds to the interpolator that minimizes $\|\bb_1\|_2^2+\|\bb_2\|_2^2/s$.
The precise form of the norm that is implicitly regularized depends on the details of
the optimization algorithm (in this case the ratio of learning rates $s$).
\end{remark}

\begin{remark}\label{rmk:Y0}
The ridge regression problem of Eq.~\eqref{eq:Ridge2Layer} presents another peculiarity.
The responses $\by$ have been replaced by the residuals at initialization
$\tby = \by-f_n(\btheta_0)$. In general, this fact should be taken into account 
when analyzing the linear system. However, it is possible to construct initializations 
close to standard random initializations and yet have $f_n(\btheta_0)=\bfzero$. We will therefore 
neglect the difference between $\by$ and $\tby$. 
\end{remark}

\section{Outline of this tutorial}
 
The rest of this tutorial is organized as follows.
\begin{description}
\item[Chapter 2] studies ridge regression under a simpler model in which the
feature vectors $\bphi(\bx)\in\reals^p$ are completely characterized by their
mean (which we will assume to be zero) and covariance $\bSigma$.
Under suitable concentration assumption on the feature vector, the resulting risk 
is universal and can be precisely characterized.
Despite its simplicity, this model displays many interesting phenomena.

In fact, while the more complex settings in the following chapters do not fit the
required concentration assumptions, their behavior is correctly predicted by this simpler model,
pointing at a remarkable universality phenomenon.
\item[Chapter 3] focuses on the infinite width ($N\to\infty$) limit of the neural tangent
model we derived above. This is described by kernel ridge regression for a rotationally invariant kernel.
We derive the generalization behavior of KRR in high dimension.
\item[Chapter 4] studies `random feature models,' which correspond to setting $\bb_2=0$
in \eqref{eq:Ridge2Layer}, i.e. fitting only the second-layer weights. 
This clarifies what happens when moving from an infinitely wide networks to finite width.
\item[Chapter 5] considers the other limit case in which we set
$\bb_1=0$ and only learn the first layer weights (in the linear approximation).

It is not hard to see that the case in which both $\bb_1,\bb_2$ are fit to data
is very close to the one in which $\bb_1=0$ and $\bb_2$ is fit to data. Therefore
this case yields the correct insights into the generalization behavior of finite width 
neural tangent models.
\item[Chapter 6] finally discusses the limitations of linear theory.
In particular, we discuss some simple examples in which we need to go beyond the neural tangent theory
to capture the behavior of actual neural networks trained via gradient descent.
\end{description}

\chapter{Linear regression under feature concentration assumptions}
\label{ch:linalg}

In this section we consider ridge regression and its limit for vanishing regularization
(minimum $\ell_2$-norm regression), focusing on the overparametrized regime
$p>n$.
Our objective is to understand if and when interpolation or overfitting is compatible with 
good generalization. With this in mind, we start by considering a simple model
in which the feature vectors are completely characterized by their covariance $\bSigma$. 
With little loss of generality, we will assume the vectors to be centered.
Under certain concentration assumptions
on these vectors, a sharp characterization of the prediction risk can be derived. This 
theoretical prediction captures in a precise way numerous interesting phenomena
such as benign overfitting and double descent.

\section{Setting and sharp characterization}
\label{eq:GaussianSetting}

In this chapter we assume to be given i.i.d. samples $\{(y_i,\bz_i)\}_{i\le n}$ where responses 
are given by
\begin{align}
y_i = \<\bbeta_*,\bz_i\>+w_i\, ,\;\;\;\;\;\; \E(w_i|\bz_i)=0 \,,\;\;\; \E(w_i^2|\bz_i)=\tau^2
\, .
\label{eq:GaussianDataModel}
\end{align}
We assume $\E(\bz_i) = \bfzero$, $\E(\bz_i\bz_i^{\sT}) = \bSigma$.
(The case of non-zero $\E(\bz_i)\neq  \bfzero$ could be treated at the cost of some notational burden,
but does not present conceptual novelties.)

While we will state results that hold for a broad class of non-Gaussian vectors, 
for pedagogical reasons we will begin with the simplest example:
\begin{align}
\bz_i\sim\normal(0,\bSigma)\perp 
w_i\sim\normal(0,\tau^2)\,.\label{eq:GaussianDataModel_bis}
\end{align}

The covariance $\bSigma\in\reals^{p\times p}$ and coefficient vector 
$\bbeta_*\in\reals^p$ are unknown, alongside the noise level $\tau$.
We estimate $\bbeta_*$ using ridge regression:
\begin{align}
 \hbbeta(\lambda):= \arg\min_{\bb\in\reals^p}
\Big\{\frac{1}{n}\big\|\by-\bZ\bb\big\|_2^2+ \lambda\|\bb\|_2^2\Big\}\, ,\label{eq:RidgeGaussian}
\end{align}
where $\bZ \in\reals^{n\times p}$ is the matrix whose $i$-th row is $\bz_i$
and $\by\in\reals^n$ is the vector whose $i$-th entry is $y_i$.
\begin{remark}
We emphasize that the above definitions make perfect sense
if $p=\infty$. In this case $\reals^p$ is interpreted to be the Hilbert space of
square summable sequences $\ell_2$ and  $\<\bu,\bv\>=\bu^{\sT}\bv$
is the scalar product in $\ell_2$. In fact, unless specified otherwise, the 
results below cover this infinite-dimensional case.
\end{remark}

We are interested in the excess test error. With a slight abuse of notation, we can
use the notation 
$R_{\sexc}(\hf) = R_{\sexc}(\lambda;\bZ,\bbeta_0,\bw)$
since $\hf(\bx) = \<\hbbeta(\lambda),\bx\>$ is
 a function of $\lambda, \bZ,\bbeta_0,\bw$. We have:
\begin{align}
R_{\sexc}(\lambda;\bZ,\bbeta_0,\bw)&:= 
\E_{\bz_{\snew}}\big\{\big(\<\hbbeta(\lambda),\bz_{\snew}\>-\<\bbeta_*,\bz_{\snew}\>)^2\big\}\\
&= \|\hbbeta(\lambda)-\bbeta_*\|_{\bSigma}^2\, .\label{eq:NoiseRisk}
\end{align}
Recall that in Chapter \ref{ch:intro}, Eq.~\eqref{eq:Excess}, we defined
the excess test error to be the difference between test error and Bayes error.
In this case the Bayes error is equal to $\tau^2$, and therefore the relation 
is particularly simple: $R(f) = R_{\sexc}(f)+\tau^2$ (reverting to using the $f$ to
denote the argument).

We also note that the ridge regression estimator can be written explicitly as
\begin{align}
\hbbeta(\lambda) = \frac{1}{n}\bS_{\lambda}\bZ^{\sT}\by\, ,\;\;\;
\bS_{\lambda} = \Big(\frac{1}{n}\bZ^{\sT}\bZ+\lambda\id_p\Big)^{-1}\, ,
\end{align}
whence we obtain the expression:
\begin{align}
R_{\sexc}(\lambda;\bZ,\bbeta_*,\bw)&:= \Big\|\lambda\bS_{\lambda}
\bbeta_*-\frac{1}{n}\bS_{\lambda}\bZ^{\sT}\bw\Big\|^2_{\bSigma}\, .
\end{align}
Note that the risk \eqref{eq:NoiseRisk} depends on the noise vector $\bw$. 
It is useful to define its expectation with respect to $\bw$,
which can be exactly decomposed in a bias term and a variance term:
\begin{align}
\oR(\lambda;\bZ,\bbeta_*)& := \E_{\bw} R_{\sexc}(\lambda;\bZ,\bbeta_*,\bw)\\
& = \Bias(\lambda;\bZ,\bbeta_*) +\Variance(\lambda;\bZ)\, .
\end{align}
The bias and variance are given by
\begin{align}
\Bias(\lambda;\bZ,\bbeta_*) & := \lambda^2\<\bbeta_*,\bS_{\lambda}\bSigma\bS_{\lambda}\bbeta_*\>\, ,\\
\Variance(\lambda;\bZ) & := \frac{\tau^2}{n}\Trace\Big(\bS_{\lambda}^2\frac{1}{n}\bZ^{\sT}\bZ\bSigma\Big)\, .
\end{align}

Of course these formulas do not provide always simple insights
into the qualitative behavior of the test error. In particular, 
$\Bias(\lambda;\bZ,\bbeta_*)$ and $\Variance(\lambda;\bZ)$ are random 
quantities because of the randomness in $\bZ$. The next theorem shows that these 
quantities concentrate around deterministic predictions that depend uniquely 
on the geometry of $(\bSigma,\bbeta_*)$. 

Before stating this characterization, we introduce some important notions.
\begin{definition}[Effective dimension] \label{asmp:Sigma-bounded-varying}
We say that $\bSigma$ has \emph{effective dimension} $ \constantsig(n)$ 
if, for all $1 \leq k \leq \min\{n, p\}$, 
	\begin{align*}
		\sum_{l=k}^p \sigma_l \leq \constantsig \sigma_k \,.
	\end{align*} 
	Without loss of generality, we will always choose $ \constantsig(n) \geq n$.
\end{definition}
\begin{definition}[Bounded varying spectrum] \label{asmp:Sigma-bounded-varying}
	We say that $\bSigma$ has  \emph{bounded varying spectrum}
	if there exists a monotone decreasing
	function $\psi:(0,1]\to [1,\infty)$ 
	with  $\lim_{\delta \downarrow 0} \psi(\delta) = \infty$, such that
	$\sigma_{\lfloor \delta i \rfloor} / \sigma_i \leq \psi(\delta)$ for all 
	$\delta \in (0, 1]$, $i \in \naturals$ and $\delta i \geq 1$.
\end{definition}

Given $\bSigma$, $\lambda$, let $\lambda_*(\lambda)\ge 0$ be the unique positive solution of
 \begin{align}
 n \Big(1-\frac{\lambda}{\lambda_*}\Big) = \Trace\Big( \bSigma(\bSigma+\lambda_*\id)^{-1}\Big)\, .
 \label{eq:FixedPointProportional}
  \end{align}
  (with $\lambda_*=0$ if $\lambda=0$ and $n\ge p$.)
        Define $\cuB(\bSigma,\bbeta_*)$ and $\cuV(\bSigma)$ by
        \begin{align}
          \cuB(\bSigma,\bbeta_*) &:= \frac{\lambda_*^2\<\bbeta_*,(\bSigma+\lambda_*\id)^{-2}\bSigma\bbeta_*\>}
           {1-n^{-1} \Trace\big(
           \bSigma^2(\bSigma+\lambda_*\id)^{-2}\big)}\label{eq:BiasProportional}\, ,\\
          \cuV(\bSigma) &:=\frac{\tau^2\Trace\big( \bSigma^2(\bSigma+\lambda_*\id)^{-2}\big)}{n-\Trace\big( \bSigma^2(\bSigma+\lambda_*\id)^{-2}\big)}
                          \, . \label{eq:VarianceProportional}
        \end{align}
Let us emphasize that $\cuB(\bSigma,\bbeta_*)$, $\cuV(\bSigma)$ are deterministic quantities.

The next theorem gives sufficient conditions under which  $\cuB(\bSigma,\bbeta_*)$,  $\cuV(\bSigma)$
are accurare multiplicative approximations of the actual bias and variance.  
 \begin{theorem}
	\label{thm:Gaussian}
Assume  $\|\bSigma\|_{\op}=1$, a setting which we can always reduce to by rescaling $\bZ$.
Further assume
$\|\bSigma^{-1/2}\bbeta_*\|<\infty$, and that one of the following scenarios holds:
\begin{description}
\item[1. Proportional regime.] There exists a constant $M<\infty$ 
such that $p/n\in [1/M,M]$, $\sigma_p\ge 1/M$, $\lambda \in [1/M,M]$. 
\item[2. Dimension-free regime.]
 $\bSigma$ has effective dimension   $\constantsig(n)$, bounded-varying spectrum,
and there exist constants $M>0$ and $\gamma\in(0,1/3)$ such that the following hold.
Define 
	  \begin{align}
	  \rho(\lambda):= \frac{\<\bbeta_*,\bSigma(\lambda_*\id+\bSigma)^{-1}\bbeta_*\>}{
	  \|\bbeta_*\|^2\Tr(\bSigma(\lambda_*\id+\bSigma)^{-1})}\, .
	  \end{align}
	  Then we have $\lambda_*(0)\le M$,
	  $\lambda \in  [ \lambda_*(0)/M,
	   \lambda_*(0)M]$ and $\constantsig(n)\le \rho(\lambda)^{1/6}n^{1 + \gamma}$. 
\end{description}

	Then, there exists a constant $C_0$ (depending uniquely on the constants in 
	the assumptions), such that for $n \geq C_0$, the following holds with probability at least $1-n^{-10}$.
	We have
	\begin{align}
\Bias(\lambda;\bZ,\bbeta_*)&= \big(1+\err_B(n)\big)\cdot\cuB(\bSigma,\bbeta_*)\, ,
\label{eq:Bias_kernel_linear_precise}\\
\Variance(\lambda;\bZ)&= \big(1+\err_V(n)\big)\cdot\cuV(\bSigma)\, ,
\label{eq:Var_kernel_linear_precise}
	\end{align}
	where, under the proportional regime (scenario 1 above), 
	we have $|\err_B(n)|\le n^{-0.49}$,  $|\err_V(n)|\le n^{-0.99}$, while, in the dimension-free regime
	(scenario 2 above),
	$|\err_B(n)|\le (\constantsig(n)/n)^3/(\rho(\lambda)^{1/2}n^{0.99})$, 
	 $|\err_V(n)|\le (\constantsig(n)/n)^3/n^{0.99}.$ 
\end{theorem}

\begin{remark}\label{rmk:SolutionFP}
Defining $F_{\bSigma}(x):=n^{-1}\Trace\big( \bSigma(\bSigma+x\id)^{-1}\big)$,
Eq. \eqref{eq:FixedPointProportional} reads:
\begin{align}
1-\frac{\lambda}{\lambda_*} = F_{\bSigma}(\lambda_*)\, .
\end{align}
For $\lambda>0$, this equation has always a unique solution  by monotonicity.

For $\lambda = 0$, the left hand side is constant and equal to $1$. The 
right-hand side is strictly monotone  decreasing with $F_{\bSigma}(0) = p/n$
and $\lim_{x\to\infty}F_{\bSigma}(x) = 0$. Hence for $p/n> 1$ (overparametrized regime), 
the equation has a unique solution $\lambda_*>0$. 

For $p/n\le 1$ (underparametrized regime), we set $\lambda_*=0$ by definition.
\end{remark}

\begin{remark}[Ridgeless limit] By virtue of the previous remark, the predicted
bias and variance $\cuB(\bSigma,\bbeta_*)$ and $\cuV(\bSigma)$, make perfect sense
for the  case $\lambda=0$. Indeed, Theorem \ref{thm:Gaussian} holds for $\lambda=0+$
as well, although this requires an additional argument and possibly larger error terms.
\end{remark}

Theorem \ref{thm:Gaussian} has a simple interpretation in terms of a simpler 
\emph{sequence model,} which we next define.
In the sequence model we want to estimate $\bbeta_*$ from observation $\by^s$ 
given by 
\begin{align}
\by^s = \bSigma^{1/2} \bbeta_* +\frac{\omega}{\sqrt{n}}\beps\, ,\;\;\;\beps \sim\normal(0,\id_p)\, .
\end{align}
We use ridge regression at regularization level $\lambda_*$ as defined in 
Eq.~\eqref{eq:FixedPointProportional}:
\begin{align}
 \hbbeta^s(\lambda_*):= \argmin_{\bb\in\reals^p}
\Big\{\big\|\by^s-\bSigma^{1/2}\bb\big\|_2^2+ \lambda_*\|\bb\|_2^2\Big\}\, .
\end{align}
Then the prediction for the risk of the original model, namely $\cuB(\bSigma,\bbeta_*)+
\cuV(\bSigma)$, coincides with the risk of the sequence model, provided we choose
$\omega$ to be the unique positive solution of
\begin{align}
\omega^2 = \frac{\tau^2}{n}+\E_{\beps}\big\{\|\hbbeta^s(\lambda_*)-\bbeta_*\|_{\bSigma}^2
\big\}\, .
\end{align}
The solution of this equation is easy to express in terms of quantities appearing in
the theorem statement
\begin{align*}
\omega^2 =\frac{\tau^2 +\lambda_*^2\<\bbeta_*,(\bSigma+\lambda_*\id)^{-2}\bSigma\bbeta_*\>}
           {1-n^{-1} \Trace\big(
           \bSigma^2(\bSigma+\lambda_*\id)^{-2}\big)}\, .
\end{align*}
    
To summarize we have the following correspondence:

\phantom{A}

\begin{tabular}{c|cc}
&{\bf Gaussian feature model} & {\bf Sequence model}\\
Design matrix & Random design matrix $\bZ$ & Deterministic design $\bSigma^{1/2}$\\
Ridge penalty & $\lambda$ &  $\lambda_*>\lambda$\\
Noise variance  & $\tau^2$ & $\omega^2>\tau^2$
\end{tabular}

\phantom{A}

Note in particular, as pointed out above (Remark \ref{rmk:SolutionFP}), in the overparametrized
regime we have $\lambda_*>0$ even if $\lambda=0$. We refer to this phenomenon as
\emph{`self-induced regularization'}: the noisy feature vectors in the original 
unregularized problem induce an effective regularization in the equivalent sequence model.

Self-induced regularization is a key mechanism by which an 
interpolating model can generalize well. Roughly speaking, although $\lambda=0+$,
the model behaves as if the sample covariance was replaced by the population one, but the 
regularization was increased from $0$ to $\lambda_*$. The noise in the covariates acts as 
a regularizer.

In order to make this correspondence even more concrete,
we note that 
\begin{align}
\frac{1}{n}\Trace\Big( \bSigma^2(\bSigma+\lambda_*\id)^{-2}\Big)<
\frac{1}{n}\Trace\Big( \bSigma(\bSigma+\lambda_*\id)^{-1}\Big)<1
\end{align}
Let us assume that this inequality holds uniformly. Namely, there exists a constant 
$c_1<\infty$ such that 
\begin{align}
\frac{1}{n}\Trace\Big( \bSigma^2(\bSigma+\lambda_*\id)^{-2}\Big)\le 1-\frac{1}{c_1}
\, .
\end{align}
Substituting in Eqs.~\eqref{eq:VarianceProportional}, \eqref{eq:BiasProportional},
we get
        \begin{align}
          \cuV(\bSigma) &\le \frac{c_1\tau^2}{n}\Trace\big( \bSigma^2(\bSigma+\lambda_*\id)^{-2}\big)  \, , \\                                            
          \cuB(\bSigma,\bbeta_*) &\le c_1\lambda_*^2\<\bbeta_*,(\bSigma+\lambda_*\id)^{-2}\bSigma\bbeta_*\>\,
           .
        \end{align}
        Notice that the expressions on the right-hand side (up to the constant $c_1$)
        are the bias and variance of the sequence model with $\omega^2=\tau^2$.
        In other words, in some cases of interest, simply considering the sequence
        model with $\omega^2=\tau^2$ allows to bound (up to constants) the risk of the 
        original problem.
  
\section{Non-Gaussian covariates}

Theorem  \ref{thm:Gaussian} holds beyond Gaussian covariates, and was proven 
under the following assumptions on the covariates. 

Let $\bz_i :=\bSigma^{-1/2}\bx_i$, so that $\bz_i$ is isotropic, namely
$\E\{\bz_i\} = \bfzero$, $\E\{\bz_i\bz_i^{\sT}\} = \id$.
We then consider the following two models for $\bz_i$, 
depending on a constant $\constantx >0$:
	\begin{enumerate}
		\item[$(a)$] \textbf{Independent sub-Gaussian coordinates}:  $\bz_i$ has independent 
		but not necessarily identically
		 distributed coordinates with uniformly bounded sub-Gaussian norm. Namely: each 
		  coordinate $z_{ij}$ of $\bz_i$ satifies $\E [z_{ij}]= 0$, $\Var (z_{ij}) = 1$ and 
		  $\norm{z_{ij}}_{\psi_2}:= \sup_{p \geq 1} 
		  p^{-\frac 1 2} \E \{|z_{ij}|^p\}^{\frac 1 p} \leq \constantx$.
		\item[$(b)$] \textbf{Convex concentration}: allowing $\bz_i$ to have dependent coordinates, 
		the following holds for any $1-$Lipschitz convex function $\varphi : \reals^p \to \reals$, 
		and for every $t > 0$
		\begin{align*}
			\prob \big\{| \varphi(\bz_i) - \E\varphi(\bz_i)| \geq t\big\} \leq 2 
			\, e^{-t^2/ \constantx^2}\, .
		\end{align*}
	\end{enumerate}

For independent sub-Gaussian coordinates, a version of 
Theorem \ref{thm:Gaussian} was proven in \cite{hastie2022surprises}, although with larger
error terms than stated here. The form stated here, the infinite-dimensional 
case, with regularly varying spectrum, and the case of covariates satisfying convex
concentration were proven in \cite{cheng2022dimension}.

\section{An example: Analysis of a latent space model}
\label{sec:Latent}

\begin{figure}
 \includegraphics[width=0.5\columnwidth,angle=0]{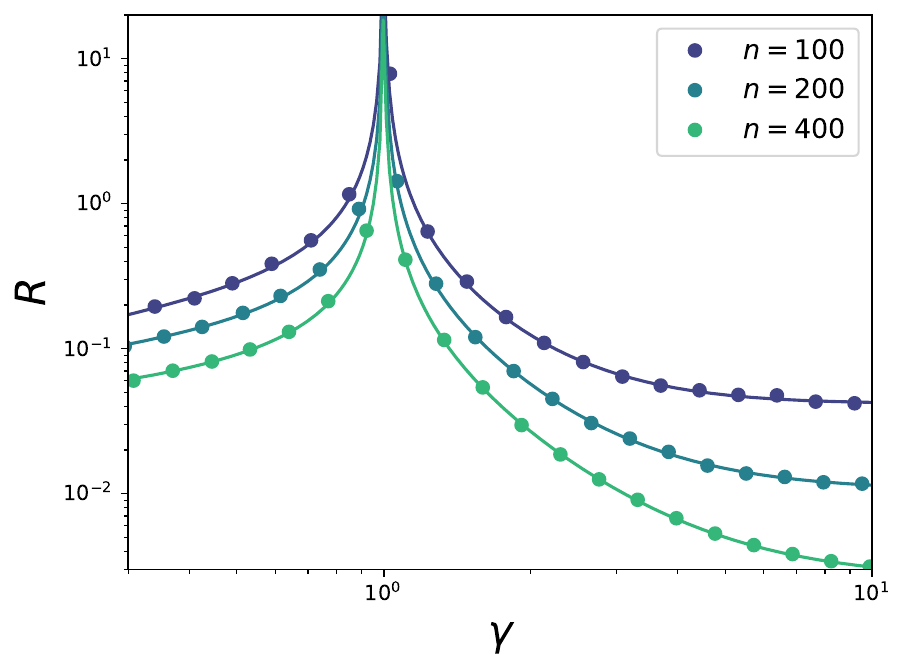}
 \includegraphics[width=0.5\columnwidth,angle=0]{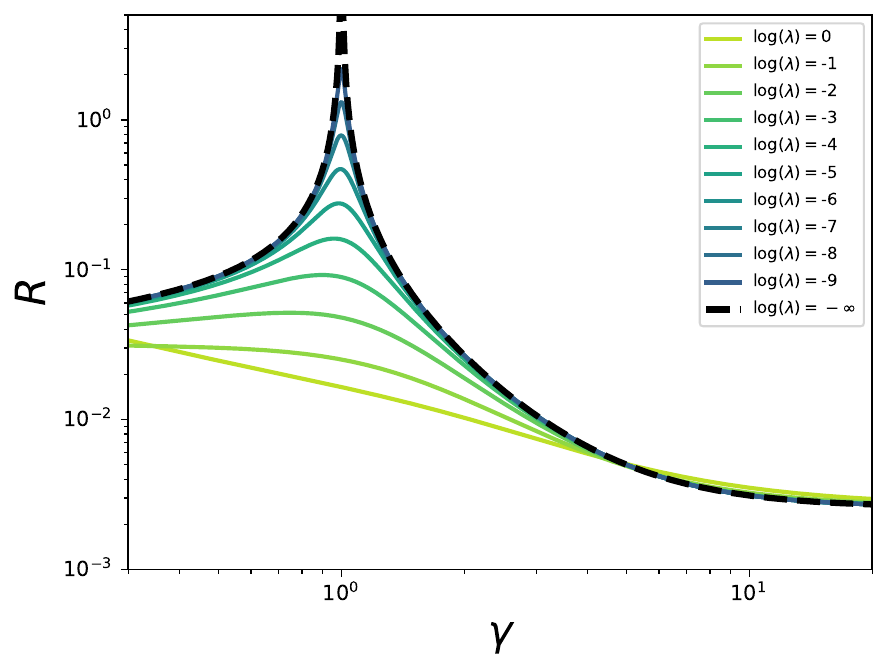}
  \caption{Test error of ridge regression under the latent space model.
   Left: circles are empirical results and curves are theoretical predictions
   within the proportional asymptotics, in the ridgeless limit $\lambda=0+$.  Here 
    $d=20$, $\tau=0$, $r_{\theta}=1$, $\mu=1$.  Different curves correspond
   to different sample sizes. 
   Right: curves correspond to different values of the regularization parameter.
   Here $\gamma\psi=1/20$, $\tau=0$, $r_{\theta}=1$, $\mu=1$. Figure from 
   \cite{hastie2022surprises}. }\label{fig:Ridge_LowRank}
\end{figure}

As an application of the general theory in Section \ref{eq:GaussianSetting},
it is instructive to consider the following \emph{latent space model}.
We assume the  response $y$ to be linear in an underlying covariate vector 
$\bx\in\reals^d$. However, we fit a model in the features $\bz\in\reals^p$,
which are also linear in $\bx$:
\begin{align}
y_i & = \<\btheta_*,\bx_i\>+\xi_i \, ,\;\;\;\;\; \xi_i\sim\normal(0,\tau^2)\, ,\\
\bz_i & = \bW\bx_i +\bu_i\, ,\;\;\;\;\; \bu_i\sim\normal(0,\id_p)\, ,
\end{align}
where the matrix $\bW$ is fixed (at random).
We perform ridge regression of $y_i$ on $\bz_i$ as in Eq.~\eqref{eq:RidgeGaussian}
(with $\bZ$ the matrix whose $i$-th row is given by $\bz_i$).

We assume a simple model for the latent features, namely  
$\bx_i\sim\normal(0,\id_d)$, and $\bW$ proportional to an orthogonal matrix. Namely 
$\bW^{\sT}\bW = (p\mu/d)\id_d$. 
We consider the proportional asymptotics $p,n,d\to\infty$ with 
\begin{align}
\frac{p}{n}\to \gamma\in (0,\infty)\, ,\;\;\;\; \frac{d}{p}\to \psi \in (0,1)\, ,
\end{align}
and $\|\btheta_*\|\to r_{\theta}$.
In particular, $\gamma$ has the interpretation of number of parameters per sample, and $\gamma>1$
corresponds to the overparametrized regime. 

Explicit formulas for the asymptotic risk can be obtained in this limit using Theorem \ref{thm:Gaussian}.
In particular, in the minimum norm limit
$\lambda\to 0$, after a tedious but straightforward calculation,
one obtains
\begin{align}
 \cuB_{\slat}(\psi,\gamma)  & :=\left\{1+\gamma c_0
                                   \frac{  \cuE_1(\psi,\gamma) }{  \cuE_2(\psi,\gamma) }\right\}\cdot\frac{\mu \psi^{-1} r^2_{\theta}}{(1+\mu\psi^{-1})
                                   (1+c_0\gamma(1+\mu\psi^{-1}))^2}\, ,\\
     \cuV_{\slat}(\psi,\gamma)  & :=\sigma^2\gamma c_0
                                   \frac{  \cuE_1(\psi,\gamma) }{  \cuE_2(\psi,\gamma) }\, ,\\
     \cuE_1(\psi,\gamma) & := \frac{1-\psi}{(1+c_0\gamma)^2}+ \frac{\psi(1+\mu\psi^{-1})^2}{(1+c_0(1+\mu\psi^{-1})\gamma)^2}\, ,\\  
\cuE_2(\psi,\gamma) &:= \frac{1-\psi}{(1+c_0\gamma)^2}+ \frac{\psi(1+\mu\psi^{-1})}{(1+c_0(1+\mu\psi^{-1})\gamma)^2}\, .
    \end{align}
    where $\sigma^2=\tau^2+r_{\theta}^2/(1+\mu\psi^{-1})$, and $c_0 =c_0(\psi,\gamma)\ge 0$ is the unique non-negative solution of the following second order equation
    \begin{align}
      1-\frac{1}{\gamma} = \frac{1-\psi}{1+c_0\gamma}+ \frac{\psi}{1+c_0(1+\mu\psi^{-1})\gamma}\, .
      \end{align}

In Figure \ref{fig:Ridge_LowRank}, left frame, we plot theoretical predictions and empirical results 
for the test error of minimum norm interpolation, as a function of the overparametrization ratio.
A few features of this plot are striking:
\begin{enumerate}
\item A large spike is present at the interpolation threshold $\gamma=1$:
the test error grows when the model becomes overparametrized but then descends again at large overparametrization.
This behavior  has been dubbed `double descent' in \cite{belkin2019reconciling} (see Appendix \ref{app:Bibliography} for further discussions).
\item The minimum test error is achieved at large overparametrization $\lambda\gg 1$.
\item In particular overparametrized models behave better than underparametrized ones.
Intuitively, the reason is that the latent space can be identified better when the 
dimension of $\bz_i$ grows. 
\end{enumerate}

In Figure \ref{fig:Ridge_LowRank}, right frame, we plot theoretical predictions
for the test error of ridge regression as a function of the overparametrization ratio
$\gamma$. The number of samples per latent dimension $n/d$ is kept fixed to $(\psi\gamma)^{-1}=20$.
Different curves correspond to different values of $\lambda$.
The test error with optimal lambda is given by the lower envelope of these curves.

Two remarkable phenomena can be observed in these plots:
\begin{enumerate}
\item The spike at $\gamma=1$ is smoothed for positive $\lambda$ and disappears completely for 
the optimal regularizations. In other words, the double descent is not an intrinsic statistical 
phenomenon, and is instead due to under-regularization.
\item Nevertheless, at large overparametrization, the optimal regularization is $\lambda=0+$.
\end{enumerate}

The last point is particularly surprising: at large overparametrization (and large signal-to-noise ratio)
the optimal test error is achieved by an interpolator. The same phenomenon survives at positive 
$\tau$, and can be traced back to the fact that the effective regularization $\lambda_*$
is strictly positive despite $\lambda=0+$.

\section{Bounds and benign overfitting}
\label{sec:bounds_BO}

For a general $\bSigma$, $\bbeta_*$, the characterization given
by Eqs.~\eqref{eq:FixedPointProportional}, \eqref{eq:BiasProportional},
 \eqref{eq:VarianceProportional} may still be too detailed to provide a simple intuition.
 In this section we derive upper bounds on bias and variance that  
expose a particularly interesting phenomenon, which was first established in 
\cite{bartlett2020benign,tsigler2020benign} 
using a technically different approach: benign overfitting.

We  work under the simplifying assumption
\begin{align}
\Tr \big\{\bSigma^2(\bSigma + \lambda_* \id)^{-2}\big\}\le n(1-c_*^{-1})\, ,\label{eq:Cstar}
\end{align}
for a constant $c_*\in (1,\infty)$. Let
 $\bSigma :=\sum_{i\ge 1}\sigma_i\bv_i\bv_i^{\sT}$ be the eigendecomposition
of $\bSigma$, with $\sigma_1\ge \sigma_2\ge \dots \ge 0$.
 Denote by $\bbeta_{*,\le k}:=\sum_{i\le k}\<\bbeta_*,\bv_i\>\bv_i$ 
the orthogonal projection of $\bbeta$
onto the span of the top $k$ eigenvectors $\bv_1,\dots,\bv_k$, and by
 $\bbeta_{*,> k}:=\bbeta_*-\bbeta_{*,\le k}$ its complement.
 Finally, let $k_*:= \max\{k\, :\; \sigma_k \ge \lambda_*\}$. 
 
 The above definitions are quite natural in view of the 
 latent space model discussed in the previous section.
 We want to separate the contribution of ``signal'' directions in covariate space
 (corresponding to the span of $ (\bv_i:i\le k_*)$), from the contribution due to 
 ``junk'' covariates (the projection onto the span of $(\bv_i:i>k_*)$).
 Technically, we bound all traces in 
  Eqs.~\eqref{eq:FixedPointProportional}, \eqref{eq:BiasProportional},
 \eqref{eq:VarianceProportional}  after splitting the contributions of each two subspaces.
 
 Consider  Eq.~\eqref{eq:FixedPointProportional}, which implies
 $n\ge \Tr \{\bSigma(\bSigma + \lambda_* \id)^{-1}\}$.
 We get
 \begin{align*}
 n& \ge \sum_{\ell=1}^{k_*} \frac{\sigma_\ell}{\sigma_\ell + \lambda_*} + \sum_{\ell=k_* +1}^p \frac{\sigma_\ell}{\sigma_\ell + \lambda_*}\\
 & \ge \sum_{\ell=1}^{k_*} \frac{\sigma_\ell }{2\sigma_\ell } + \sum_{\ell=k_* +1}^p \frac{\sigma_\ell}{2\lambda_*}  \\
&\ge 	\frac{k_*}{2}+ \frac{r_1(k_*)}{2b_{k_*}} \, ,
\end{align*}
where we defined $b_{k}:=\sigma_{k}/\sigma_{k+1}$ and 
\begin{align}
r_q(k) := \sum_{\ell>k}\Big(\frac{\sigma_\ell}{\sigma_{k+1}}\Big)^q\,  .\label{eq:Rq}
\end{align}
Also note that the reverse inequality $n\le 2k_*+2 r_1(k_*)$ can be proved along the same line,
as long as $\lambda \le \lambda_*/2$.

In many situations of interest eigenvalues become less spaced as $k\to\infty$,
and therefore $b_k$ is bounded by a constant. For instance, this is the case if 
$\sigma_k\asymp k^{-\alpha+o_k(1)}$.  Further $r_q(k)$ can be regarded as a measure of the 
number of eigenvalues of the same order as $\sigma_{k+1}$, and is therefore 
an `effective rank' at level $k$. 
 
Next we bound $\cuV(\bSigma)$. Recalling that  
$\Tr \prn{\bSigma^2(\bSigma + \lambda_* \id)^{-2}}\le n(1-c_*^{-1})$, we have
\begin{align*}
	\cuV(\bSigma) & \le  \frac{c_* \tau^2}{n} \cdot \prn{ \sum_{\ell=1}^{k_*} \frac{\sigma_\ell^2}{(\sigma_\ell + \lambda_*)^2} + \sum_{\ell=k_* +1}^p \frac{\sigma_\ell^2}{(\sigma_\ell + \lambda_*)^2}} \\
	& \leq \frac{c_* \tau^2}{n} \cdot \prn{k_* + \sum_{l=k_* +1}^d \frac{\sigma_l^2}{\lambda_*^2}}\\
	& \leq c_*\tau^2\Big(\frac{k_*}{n}+\frac{r_2(k_*)}{n}\Big)\\
	& \leq c_*\tau^2\Big(\frac{k_*}{n}+\frac{4 b_{k_*}^2 n }{\overline{r}(k_*)}\Big)  \, ,
\end{align*}
where in the last step we defined
\begin{align}
\overline{r}(k) := \frac{r_1(k)^2}{r_2(k)}\, ,\label{eq:oR}
\end{align}
and used the bound (derived above) $n\ge r_1(k_*)/(2b_{k_*})$.

We proceed similarly for the bias, namely:
\begin{align*}
\cuB(\bSigma,\bbeta_*) & \leq c_* \sum_{\ell=1}^d \frac{\lambda_*^2 \sigma_\ell}{(\sigma_\ell + \lambda_*)^2} \<\bbeta_*,\bv_\ell\>^2 \nonumber \\
	& \leq c_* \prn{\sum_{\ell=1}^{k_*} \lambda_*^2 \sigma_\ell^{-1} \<\bbeta_*,\bv_\ell\>^2 + \sum_{\ell=k_* + 1}^{d} \sigma_\ell  \<\bbeta_*,\bv_\ell\>^2}\\
	& \le c_* 
	\Big(\sigma_{k_*}^2 \|\bbeta_{*,\le k_*}\|_{\bSigma^{-1}}^2 +\|\bbeta_{*,>k_*}\|_{\bSigma}^2\Big)\, .
\end{align*}

We summarize these bounds in the statement below.
\begin{proposition}\label{propo:Benign}
 Define $r_q(k)$ via Eq.~\eqref{eq:Rq},  $\overline{r}(k)$ via Eq.~\eqref{eq:oR},
 $k_*:= \max\{k\, :\; \sigma_k \ge \lambda_*\}$, and  $b_k:=\sigma_k/\sigma_{k+1}$.
 
Under condition \eqref{eq:Cstar}, we have $n \ge k_*/2+ r_1(k_*)/(2b_{k_{*}})$  and
(for $\lambda\le \lambda_*/2$) $n\le 2k_*+2 r_1(k_*)$. Further
\begin{align}
\cuV(\bSigma)  &\le c_*\tau^2\Big(\frac{k_*}{n}+\frac{r_2(k_*)}{n}\Big)
\le c_*\tau^2\Big(\frac{k_*}{n}+\frac{4b^2_{k_{*}}n }{\overline{r}(k_*)}\Big)\, ,\label{eq:CrudeBoundVar}\\
\cuB(\bSigma,\bbeta_*) &\le c_* 
\Big(\sigma_{k_*}^2 \|\bbeta_{*,\le k_*}\|_{\bSigma^{-1}}^2 +\|\bbeta_{*,>k_*}\|_{\bSigma}^2\Big)\, .\label{eq:CrudeBoundBias}
\end{align}
\end{proposition}

As mentioned above, bounds of this form\footnote{There are some technical differences between
the present statement and the results of \cite{bartlett2020benign,tsigler2020benign}.
We refer to \cite{cheng2022dimension} for a discussion of the differences.} were first proven 
in \cite{bartlett2020benign,tsigler2020benign}.
These bounds have an interesting consequence. They allow us to characterize pairs
$\bSigma,\bbeta_*$ (with $p=\infty$) for which min-norm interpolation (the $\lambda=0+$
limit of ridge regression) is `consistent' even if $\tau>0$. 
In other words, two things are happening at the same time:
\begin{enumerate}
\item The fitted model perfectly interpolates the train data (the train error vanishes).
\item The excess test error vanishes as $n\to\infty$. (In statistics language, model is consistent.)
\end{enumerate}
When these two elements occur together, we speak of benign overfitting.

Proposition \ref{propo:Benign} allows to determine sufficient conditions for benign overfitting.
We will assume for simplicity $\sigma_k\to 0$ as $k\to\infty$ and $b_k$ bounded.

Note that the condition  $n\le 2k_*+2 r_1(k_*)$ implies $k_*\to\infty$ as $n\to\infty$.
Hence, for the bias to vanish it is sufficient that $\|\bbeta\|_{\bSigma^{-1}}<\infty$.
Indeed, this implies
$\cuB(\bSigma,\bbeta_*) \le c_* (\sigma_{k_*}^2 \|\bbeta_*\|^2_{\bSigma^{-1}}+
\sigma_{k_*+1}^2\|\bbeta_*\|^2_{\bSigma^{-1}})\to 0$. Summarizing
\begin{align}
\|\bbeta_*\|^2_{\bSigma^{-1}}<\infty\;\; \Rightarrow \;\;\; \cuB(\bSigma,\bbeta_*) \to 0\, .
\end{align}
For instance, if $\<\bbeta_*,\bv_k\>\neq 0$ only for finitely many $k$, then this condition 
is obviously satisfied. More generally, it conveys the intuition that $\bbeta_*$ should
be mostly aligned with the top eigendirections of $\bSigma$. The number of eigendirections one should
take into account diverges with $n$.

Next consider the variance $\cuV(\bSigma)$.
Clearly, in order for the bound in Eq.~\eqref{eq:CrudeBoundVar} to vanish,
the following conditions must be satisfied:
\begin{align}
 \frac{k_*}{n}\to 0\, , \;\;\;\;\;\; \frac{\overline{r}(k_*)}{n}\to\infty\, .\label{eq:BenignVariance}
 \end{align}
In order to get some intuition, let us start by considering the case of 
polynomially decaying eigenvalues $\sigma_k\asymp k^{-\alpha}$. We need to take
$\alpha>1$ to ensure $\|\bx_i\|^2<\infty$ almost surely. 
We then have, for $q\ge 1$,
\begin{align*}
r_q(k) \asymp \sum_{\ell>k} \Big(\frac{k}{\ell}\Big)^{q\alpha}\asymp k\, ,
\end{align*}
whence 
\begin{align*}
k_*(n) \asymp n\, ,\;\;\;\;\; \overline{r}(k_*) \asymp n\, .
\end{align*}
Hence the conditions \eqref{eq:BenignVariance} for the vanishing of
variance do not hold in this case. 

We insist, and consider a more slowly decaying sequence of eigenvalues 
$\sigma_k\asymp k^{-1}(\log k)^{-\beta}$, $\beta>1$. In this case
\begin{align*}
r_q(k) & \asymp k^q (\log k)^{q\beta}\sum_{\ell>k} \frac{1}{\ell^q (\log \ell)^{q\beta}}\\
&\asymp k^q (\log k)^{q\beta}\int_k^{\infty} \frac{1}{x^q (\log x)^{q\beta}}\, \de x\\
& \asymp k^q (\log k)^{q\beta}\int_{\log k}^{\infty} \frac{e^{-(q-1)t}}{t^{q\beta}}\, \de t\, .
\end{align*}
Therefore
\begin{align*}
r_q(k) \asymp \begin{cases}
k \log k &  \mbox{ if $q=1$,}\\
k &  \mbox{ if $q>1$,}
\end{cases}
\end{align*}
whence $\overline{r}(k)\asymp k (\log k)^2$.
Using the bounds for $k_*(n)$ in Proposition \ref{propo:Benign}, we conclude that
\begin{align*}
k_*(n) \asymp \frac{n}{\log n}\, ,\;\;\;\;\; \overline{r}(k_*) \asymp n\log n\, .
\end{align*}
The conditions of Eq.~\eqref{eq:BenignVariance} are therefore satisfied in this case.

\chapter{Kernel ridge regression in high dimension}
\label{ch:random_matrix}

In this chapter we consider linearized neural networks in the infinite-width limit
 $N \to \infty$.
 There are two reasons for beginning our analysis from this limit case:
 \begin{enumerate}
\item In the $N\to\infty$ limit, the ridge regression problem
  \eqref{eq:Ridge2Layer} simplifies  to \textit{kernel ridge regression} 
  (KRR) with respect to an inner-product kernel. On one hand, this is a simpler problem 
  than the original one. On the other KRR is an interesting method for its own sake.
\item As we will see in the next two chapters, mildly overparametrized networks 
in the linear regime behave similarly to their $N=\infty$ limit.
 \end{enumerate}

The specific kernel arising by taking the wide limit of neural networks 
is commonly referred to as  \emph{neural tangent kernel (NTK)} \cite{jacot2018neural}.
For the sake of clarity  we will refer to it as the \emph{infinite width NTK}. 
In the case of fully connected networks (for any constant depth), 
there is hardly anything special about the NTK. As we will see, the analysis can be carried 
out in a unified fashion for any inner product kernel and the behavior is qualitatively 
independent of the specific kernel under some genericity assumptions.

 The goal of this chapter is to obtain a tight characterization of the test error of
  KRR in high dimension.

\section{Infinite-width limit and kernel ridge regression}

Recall from Section \ref{sec:GeneralLinearization} that we are interested in the ridge 
regression estimator
\begin{align}
\hbb(\lambda) = \argmin_{\bb\in\reals^p}
\Big\{\big\|\tby-\bPhi\bb\big\|_2^2+ \lambda \|\bb\|_2^2 \Big\}\, ,\label{eq:RidgeRegProb}
\end{align}    
where $\bPhi = \big[ \bphi ( \bx_1)^\sT , \ldots , \bphi (\bx_n )^\sT \big]^\sT \in 
\reals^{n \times p}$ 
is the matrix containing the feature vectors, and 
$\bphi : \reals^d \to \reals^p$ is a featurization map.
We are particularly interested in the featurization map  obtained 
by linearizing a two-layer neural network, as in Eq.~\eqref{eq:linearization_RF_NT} 
(in this case $p = N(d+1)$). For simplicity, we will assume $f (\bx ; \btheta_0 ) =0$ and 
replace $\tby$ by $\by$ (cf. Remark \ref{rmk:Y0}).

The minimizer of Problem \eqref{eq:RidgeRegProb} can be written explicitly as 
\begin{align}
\hbb (\lambda) = \big( \lambda \id_p + \bPhi^\sT \bPhi \big)^{-1} \bPhi^\sT \by = \bPhi^\sT \big( \lambda \id_n + \bPhi \bPhi^\sT \big)^{-1} \by\, .
\end{align}
The prediction function is given by
\begin{align}
\hat f_\lambda ( \bx) = \< \hbb ( \lambda) , \bphi ( \bx) \> = \bphi (\bx)^\sT \bPhi^\sT \big( \lambda \id_n + \bPhi \bPhi^\sT \big)^{-1} \by \, .
\label{eq:PredictionFunctionRR}
\end{align}

Let us introduce the kernel function $K : \reals^d \times \reals^d \to \reals$ defined by $K ( \bx_1 , \bx_2 ) = \< \phi ( \bx_1 ) , \phi (\bx_2 ) \>$, where $\< \bu, \bv \> = u_1v_1 + \ldots + u_p v_p$ denotes the standard euclidean inner-product on $\reals^p$. We further define $\bK_n = ( K (\bx_i , \bx_j ) )_{1 \leq i,j \leq n} = \bPhi \bPhi^\sT \in \reals^{n \times n}$ the empirical kernel matrix evaluated at the $n$ data points. We can rewrite the prediction function \eqref{eq:PredictionFunctionRR} as
\begin{align}
\hat f_\lambda ( \bx)  = \bK ( \bx , \cdot )^\sT \big( \lambda \id_n + \bK_n \big)^{-1} \by \, ,
\label{eq:PredictionFunctionKRR}
\end{align}
where we denoted $\bK (\bx , \cdot ) = \big( K ( \bx , \bx_1 ) , \ldots , 
K ( \bx , \bx_n ) \big)^\sT \in \reals^n$. Equivalently, the 
prediction function \eqref{eq:PredictionFunctionKRR} corresponds to the
 kernel ridge regression estimator with kernel $K$ and regularization parameter 
 $\lambda$ (see Remark \ref{rmk:KRR} below). In other words, performing ridge regression 
 on a linear model with featurization map $\bphi$ is equivalent to 
 performing kernel ridge regression with kernel $K (\, \cdot\, , \,\cdot\, ) 
 = \< \bphi (\,\cdot\, ) , 
 \bphi (\,\cdot\, ) \>$.

\begin{remark}[Reproducing kernel Hilbert space]\label{rmk:RKHS}
In general, consider a feature map $\bphi : \reals^d \to \cF$ where $\cF$ is a Hilbert space, 
often called the `feature space', with inner product $\<\, \cdot\, , \,\cdot\, \>_{\cF}$ 
and norm $\| \,\cdot\, \|_{\cF} = \<\, \cdot\, , \,\cdot \,\>_{\cF}^{1/2}$. 
Introduce the function space
\begin{align}
\cH := \left\{ h (\cdot ) = \< \btheta , \bphi ( \, \cdot\, ) \>_{\cF} \, \, : \,\, 
\theta \in \cF , \, \| \btheta \|_{\cF} < \infty \right\} \, .
\end{align}
Then $\cH$ is a \emph{reproducing kernel Hilbert space} (RKHS) with reproducing
 kernel given by $K ( \bx_1 , \bx_2 ) = \< \bphi ( \bx_1) , \bphi (\bx_2 ) \>_{\cF}$ and 
 RKHS norm
\begin{align}
\| h \|_{\cH} = \inf \{ \| \btheta \|_{\cF} : \btheta \in \cF , h (\, \cdot\, ) = \< 
\btheta , \bphi ( \, \cdot \, ) \>_{\cF}   \}\, . 
\end{align} 
(Conversely, any RKHS with reproducing kernel $K$ induces 
a featurization map, e.g. taking $\cF = \cH$ and $\bphi ( \bx) = K (\bx, \, \cdot\,  )$.) 

In the simple case of Eq.~\eqref{eq:RidgeRegProb}, $\bphi : \reals^d \to \reals^p$, 
$\cF = \reals^p$ and the RKHS is simply the finite-dimensional set of linear functions 
$h (\bx) = \< \bb , \bphi( \bx) \>$ with $\bb \in \reals^p$ and $\| \bb \|_2 < \infty$. However,
 in general, we can take $\cF$ to be infinite-dimensional.
 The RKHS framework is particularly useful because of this flexibility
 (see next remark). We refer the reader to 
 \cite{berlinet2011reproducing} for a general introduction to the theory of RKHS
  and kernel methods.
\end{remark}

\begin{remark}[Kernel ridge regression]\label{rmk:KRR}  
Kernel ridge regression is a general approach to learning that abstracts the specific
examples of ridge regression that we studied so far. 
In a first step, we map the data $\bx \mapsto \phi (\bx)$ into a feature space 
$(\cF , \<\, \cdot\, , \,\cdot\, \>_{\cF} )$. We then fit a low-norm linear predictor
 with respect to this embedding, i.e. $\hat f_\lambda ( \bx) = \< \hbtheta (\lambda) , 
 \bphi (\bx) \>_{\cF}$  where
\[
\hbtheta (\lambda) := \argmin_{\btheta \in \cF} \Big\{ \sum_{i=1}^n \big( y_i -
 \< \btheta , \bphi (\bx_i) \>_{\cF} \big)^2 + \lambda \| \btheta \|_{\cF}^2 \Big\} \, .
\]
From Remark \ref{rmk:RKHS}, this is equivalent to the following:
\begin{align}\label{eq:def_KRR}
\hat f_\lambda = \argmin_{f \in \cH } \Big\{\sum_{i=1}^n \big( y_i - f ( \bx_i) \big)^2 +
 \lambda \| f \|_{\cH}^2 \Big\}\, ,
\end{align}
where $\cH$ is the RKHS associated to the feature map $\bphi$. By the Representer Theorem, the solution \eqref{eq:def_KRR} is given explicitly by $\hat f_\lambda ( \bx) = \hat a_{1} (\lambda) K (\bx , \bx_1) + \ldots + \hat a_n (\lambda) K (\bx , \bx_n)$ where $K$ is the reproducing kernel of $\cH$ and
\[
\hat \ba (\lambda) = \argmin_{\ba \in \reals^n} \Big\{ \| \by - \bK \ba \|_2^2 + \lambda \ba^\sT \bK \ba \Big\} = (\lambda \id_n + \bK )^{-1} \by \, ,
\]
with $\bK = (K(\bx_i , \bx_j) )_{i,j \in [n]}$ the kernel matrix. This is indeed
 the solution found in Eq.~\eqref{eq:PredictionFunctionKRR}. Note 
 the following important observation: we do not need to evaluate the 
 (potentially infinite-dimensional) feature maps $\phi ( \bx_i)$ but
  only their inner-products $K(\bx_i , \bx_j ) = \< \bphi (\bx_i) , \bphi (\bx_j ) \>_{\cF}$ 
which can often be done efficiently. This is known as the \emph{kernel trick}.
\end{remark}

In our case, the feature map is induced by the linearizion of a two-layer neural network, 
$\phi ( \bx) = \big[ \bphi_{\sRF} ( \bx) , \bphi_{\sNT} (\bx ) \big] \in \reals^{N(d+1)}$, 
where we recall that  
\begin{align}
\bphi_{\sRF}(\bx) &= \frac{1}{\sqrt{N}} 
[\sigma(\<\bw_{1},\bx\>);\dots; \sigma(\<\bw_{N},\bx\>)]\, ,\\
\bphi_{\sNT}(\bx) &= \frac{1}{\sqrt{Nd}} 
[\sigma'(\<\bw_{1},\bx\>)\bx^{\sT};\dots; \sigma'(\<\bw_{N},\bx\>)\bx^{\sT}]\, .
\end{align}
The associated kernel is given by 
\begin{align}
K_N ( \bx_1 , \bx_2 ) = \< \bphi ( \bx_1 ) , \bphi ( \bx_2 ) \>  =   
K_N^{\sRF} ( \bx_1 , \bx_2 ) + K_N^{\sNT} ( \bx_1 , \bx_2 )\, , 
\end{align}
where
\begin{align}
K_N^{\sRF} ( \bx_1 , \bx_2 ) : =&~ \< \bphi_{\sRF} ( \bx_1 ) , \bphi_{\sRF} (\bx_2) \> = \frac{1}{N} \sum_{i = 1}^N \sigma ( \< \bw_i , \bx_1 \>) \sigma ( \< \bw_i , \bx_2 \> ) \, , \\
K_N^{\sNT} ( \bx_1 , \bx_2 ) : =&~ \< \bphi_{\sNT} ( \bx_1 ) , \bphi_{\sNT} (\bx_2) \> = \frac{1}{Nd} \sum_{i = 1}^N \< \bx_1 , \bx_2 \> \sigma ' ( \< \bw_i , \bx_1 \>) \sigma  ' ( \< \bw_i , \bx_2 \> )\, .
\end{align}
These kernels are random because of the random weights $\bw_i$, and are finite-dimensional, with rank at most $N$ and $Nd$ respectively.

 We draw $\bw_1 , \ldots , \bw_N$ i.i.d.~from a common distribution $\nu$ on $\reals^d$. 
 As the number of neurons goes to infinity, both kernels converge pointwise to their 
 expectations by law of large numbers
 \begin{align}
 K_N^{\sRF} ( \bx_1 , \bx_2 )  \, \, \to \,\, K^{\sRF} ( \bx_1 , \bx_2 ) \, , \qquad K_N^{\sNT} ( \bx_1 , \bx_2 )  \, \, \to \,\, K^{\sNT} ( \bx_1 , \bx_2 ) \, ,
 \label{eq:Pointwise}
 \end{align}
 where (with $\bw \sim \nu$)
 \begin{align}
K^{\sRF} (  \bx_1 , \bx_2 ) =&~ \E_{\bw} \left\{ \sigma \big( \< \bw , \bx_1 \> \big) \sigma \big( \< \bw , \bx_2 \> \big) \right\} \, , \\
 K^{\sNT} ( \bx_1 , \bx_2 ) =&~ \frac{1}{d} \< \bx_1 , \bx_2 \>  \E_{\bw} \left\{ \sigma ' \big( \< \bw , \bx_1 \> \big) \sigma ' \big( \< \bw , \bx_2 \> \big) \right\} \, .
 \end{align}
 Let us emphasize that the pointwise convergence of Eqs.~\eqref{eq:Pointwise}
 does not provide any quantitative control on how large $N$ has to be for 
  finite-$N$ ridge regression to behave similarly to $N=\infty$ ridge regression.
  This question will be addressed in the next two chapters.

We will consider either $\bw \sim \Unif ( \S^{d-1})$ (with $\S^{d-1} = \{ \bw \in \reals^d : \| \bw \|_2 = 1 \}$ the unit sphere in $d$ dimensions) or $\bw \sim \normal ( 0 , \id_d / d)$.  These correspond to standard initializations used in neural networks and both are very close to each other for $d \gg 1$. Since the distribution of $\bw$ is invariant under rotations in $\reals^d$, the kernels $K^{\sRF}$ and $K^{\sNT}$ are also invariant under rotations and can therefore be written as functions of $\| \bx_1 \|_2$, $\| \bx_2 \|_2$ and $\< \bx_1 , \bx_2 \>$. 

We will assume hereafter that the data is normalized with $\| \bx \|_2 = \sqrt{d}$. Then we can write
 \begin{align}\label{eq:InnerProductKernels}
 K_N^{\sRF} ( \bx_1 , \bx_2 ) = h^{(d)}_{\sRF} \left( \frac{\< \bx_1 , \bx_2 \>}{d} \right) \, , \qquad K_N^{\sNT} ( \bx_1 , \bx_2 )  = h^{(d)}_{\sNT} \left( \frac{\< \bx_1 , \bx_2 \>}{d} \right) \, .
 \end{align}
 Kernels that only depend on the inner product of their inputs are also called \emph{inner-product} or \emph{dot-product} kernels.

\begin{remark}
Let us comment on the normalization choice in these notes. First,
 for $\| \bx \|_2 = \sqrt{d}$ and $\bw_i \sim \Unif ( \S^{d-1})$ or
  $\normal ( 0 , \id_d / d)$, the input $\< \bw_i , \bx \>$ to the non-linearity $\sigma$
   has variance of order $1$. This is the correct behavior for the activation function to 
   behave in a nontrivial manner. 
    Second, we allow the kernels $h^{(d)}_{\sRF}$ and 
   $h^{(d)}_{\sNT}$ to depend on the dimension, and hence the scaling of the argument
    $\< \bx_1 , \bx_2 \> / d$ is somewhat arbitrary. However, the specific choice 
   of Eq.~\eqref{eq:InnerProductKernels} is motivated by the fact that, with this scaling
   the functions $h^{(d)}_{\sRF}$, $h^{(d)}_{\sNT}$ converge to a well defined limit as 
   $d\to\infty$.
   More precisely:
\begin{itemize}
\item[$1)$] For $\bw \sim \normal ( 0 , \id_d / d )$, $h^{(d)}_{\sRF}$ and $h^{(d)}_{\sNT}$ 
are independent of $d$ and are given by (with $(G_1 , G_2 ) \sim \normal ( 0 , \id_2)$)
\begin{align}
h_{\sRF} (t) =&~ \E_{G_1, G_2} \left\{  \sigma \big(G_1\big)  \sigma \big( tG_1 + \sqrt{1 - t^2} G_2 \big)\right\}\, ,\\
h_{\sNT} (t) =&~ t \, \E_{G_1, G_2} \left\{  \sigma ' \big(G_1\big)  \sigma ' \big( tG_1 + \sqrt{1 - t^2} G_2 \big)\right\}\, ,
\end{align}
 where the expectation is taken with respect to $(G_1 , G_2 ) \sim \normal ( 0 , \id_2)$.
 
\item[$2)$] For $\bw \sim \Unif ( \S^{d-1} )$, $h^{(d)}_{\sRF}$ and $h^{(d)}_{\sNT}$  
are nearly independent of $d$. Namely, because of the concentration of the norm of Gaussian random vectors, we have $h^{(d)}_{\sRF} \to h_{\sRF}$ and $h^{(d)}_{\sNT} \to h_{\sNT}$ as $d \to \infty$, the same kernels as for isotropic Gaussian weights.
\end{itemize}
\end{remark}

\begin{remark}\label{rmk:multilayer_NTK}
More generally, consider a multilayer fully-connected neural network defined as 
\begin{align}
f ( \bx ; \btheta ) : = \bW_L  \sigma \big(  \bW_{L-1} \sigma \big( \bW_{L-2} \cdots \sigma (\bW_1 \bx ) \cdots \big) \big) \, ,
\end{align}
where the parameters are $\btheta = ( \bW_1 , \ldots , \bW_L )$ with $\bW_{L} \in \reals^{1 \times d_{L}}$, $\bW_\ell \in \reals^{d_{\ell+1} \times d_\ell}$ and $d_1 = d$. If $(\bW_\ell)_{ij} \sim_{iid} \normal ( 0 , \tau_\ell^2 )$, then the expected kernel 
\begin{align}
K^{\sNT,\stL} ( \bx_1 , \bx_2 ) := \E_{\btheta} \left\{ \< \nabla_\btheta f ( \bx_1 ; \btheta ) , \nabla_\btheta f ( \bx_2 ; \btheta ) \> \right\} \, ,
\end{align}
is rotationally invariant. In particular, if the inputs are normalized
 $\| \bx_i \|_2 = \sqrt{d}$, then the kernel must take the form
\begin{align}
K^{\sNT,\stL} ( \bx_1 , \bx_2 ) = h^{(d)}_{\sNT,\stL} \left( \frac{\< \bx_1 , \bx_2 \>}{d} \right) \, .
\end{align}
Similarly to the two-layer case, if we scale the parameters $(\bW_\ell)_{ij} \sim_{iid} \normal ( 0 , \tau_\ell^2 / d_{\ell} )$ and take $\min_{\ell =1 , \ldots, L} d_{\ell} \to \infty$, we have pointwise convergence 
\begin{align}
K^{\sNT,\stL}_N ( \bx_1 , \bx_2 ) := \< \nabla_\btheta f ( \bx_1 ; \btheta ) , \nabla_\btheta f ( \bx_2 ; \btheta ) \>  \to K^{\sNT,\stL} ( \bx_1 , \bx_2 )\, .
\end{align}
This limiting kernel is referred to as the \emph{neural tangent kernel}  \cite{jacot2018neural}.
\end{remark}

\section{Test error of KRR in the polynomial high-dimensional regime}
\label{sec:KRR_test_error}

In the previous section, we saw that under suitable distribution of the weights, ridge regression 
with large enough network size can be approximated by kernel ridge regression with an 
inner-product kernel. In the following, we characterize the risk of KRR for general 
inner-product kernels (in particular the neural tangent kernel with any number of 
layers by Remark \ref{rmk:multilayer_NTK}). We 
 defer to Chapters \ref{ch:random_features} and \ref{ch:neural_tangent} the important
  question of how many neurons is needed for the finite-width networks to have similar 
  risk as their infinite-width limits.

Throughout this section we consider an isotropic model for the distribution of the covariates $\bx_i \in \reals^d$. We assume to be given i.i.d. samples $\{ ( y_i , \bx_i ) \}_{i \leq n}$ with 
\begin{align}
y_i  = f_\star ( \bx_i) + \eps_i\, , \qquad \bx_i \sim \Unif \big( \S^{d-1} (\sqrt{d} )\big)\, ,
\label{eq:DistKRRsetting}
\end{align}
where $f_\star \in L^2 := L^2 ( \S^{d-1} (\sqrt{d} ) )$ is 
a general square-integrable function on the sphere of radius $\sqrt{d}$, i.e. $\E \{ f_\star ( \bx )^2 \} < \infty$, and the noise $\eps_i$ is independent of $\bx_i$, with $\E \{ \eps_i \} =0$, $\E \{ \eps_i^2 \} = \tau^2$.

We will consider a general rotationally invariant kernel $K ( \bx_1 , \bx_2 ) = h ( \< \bx_1 , \bx_2 \> / d)$, where we take $h : [ -1 , +1 ] \to \reals$ independent of the dimension. We further assume $h$ to be `generic', meaning that in the basis of Hermite polynomials $\{ \He_k \}_{k \geq 0}$,
\begin{align}\label{eq:genericity_KRR}
h \big(t /\sqrt{d} \big) = \sum_{k = 0}^\infty c_{d,k} \He_k (t) \, ,\qquad c_{d,k} := \frac{1}{k!}\E_{G \sim \normal (0,1)} [ \He_k(G) h (G/\sqrt{d} ) ] \, ,
\end{align}
has all its Hermite coefficients\footnote{The normalization is not important,
but to be definite, we choose the standard one $\E[\He_j(G) \He_k(G)]=k!\mathbbm{1}_{j=k}$
($G\sim\normal(0,1)$).} satisfying $d^{k/2} c_{d,k}  \to c_k >0$ as $d \to \infty$ 
for all $k \geq 0$. This corresponds to a universality condition:
 if $c_k = 0$, then the KRR estimator will not fit degree-$k$ spherical components of
  the target function, no matter the number of samples.
  
  Several of the assumptions above have been partially relaxed in the literature, and 
  we will provide pointers below. However, for clarity of exposition, we
   consider the simplest possible setting.

We are interested in the KRR estimator $\hat f_\lambda$ with kernel $h$ as described above. Recall that $\hat f_\lambda$ is given by (see Eq.~\eqref{eq:PredictionFunctionKRR})
\begin{align}
\hat f_\lambda ( \bx)  = \bK ( \bx , \cdot )^\sT \big( \lambda \id_n + \bK_n \big)^{-1} \by \, .
\end{align}
We are interested in the excess test error under square loss which we denote
\begin{align}
R_{\KRR} ( f_\star ; \bX , \by , \lambda ) := \E_{\bx} \left\{ \big( f_\star ( \bx) - \hat f_{\lambda} ( \bx) \big)^2 \right\} \, .
\label{eq:TestErrorKRR}
\end{align}

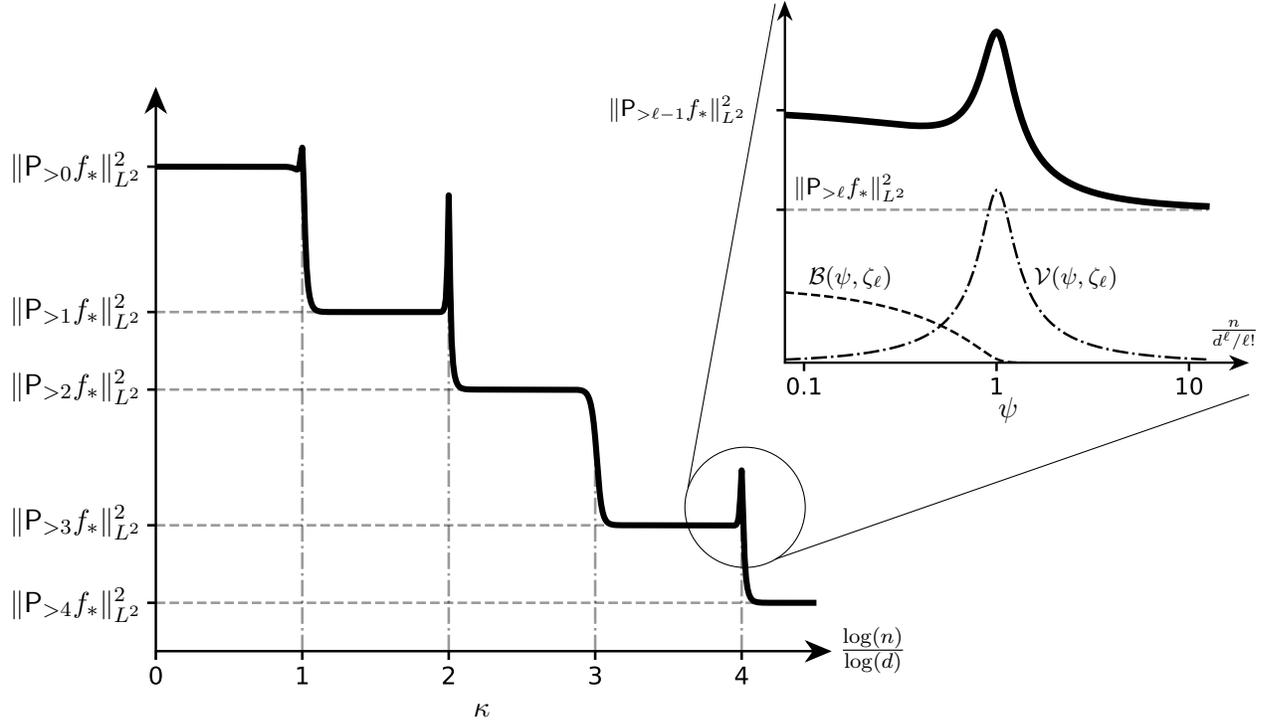
\begin{figure}[t]
\begin{tikzpicture}

\node[inner sep=0pt] (russell) at (0,0)
    {\includegraphics[width=.7\textwidth]{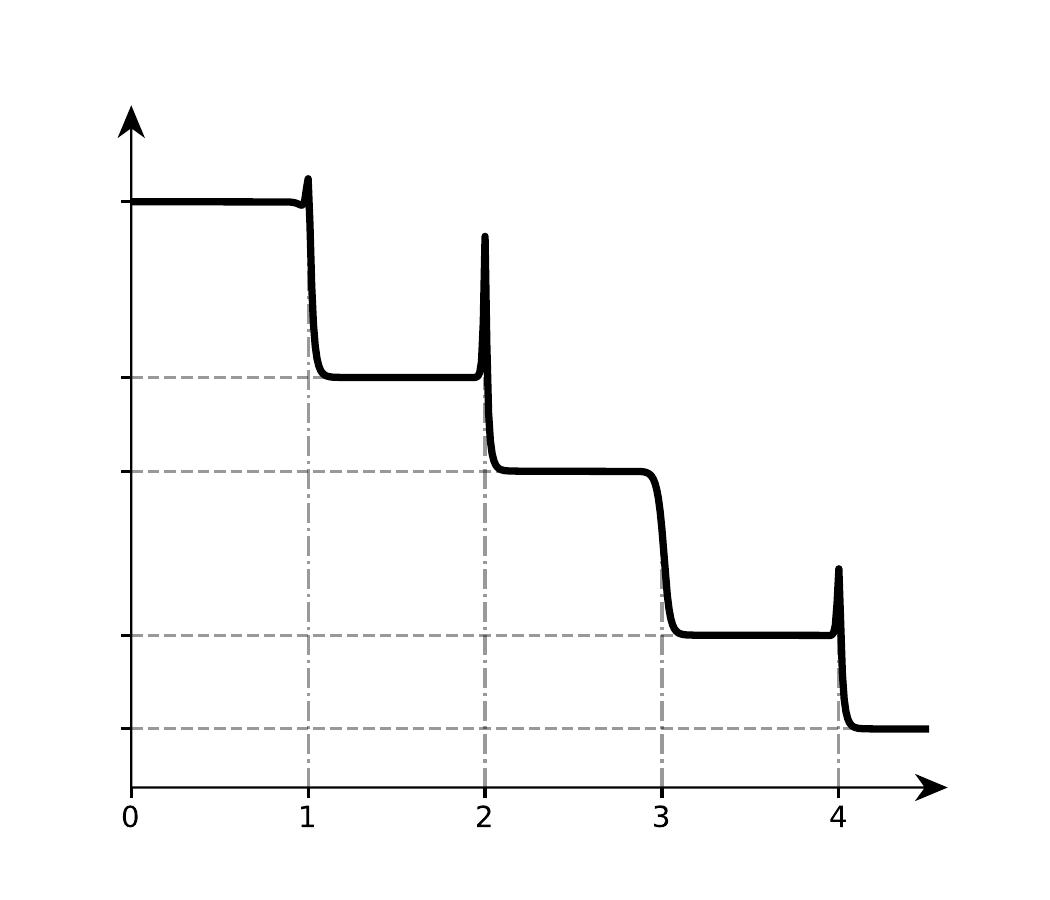}};
\node[inner sep=0pt] (whitehead) at (7,2.5)
    {\includegraphics[width=.48\textwidth]{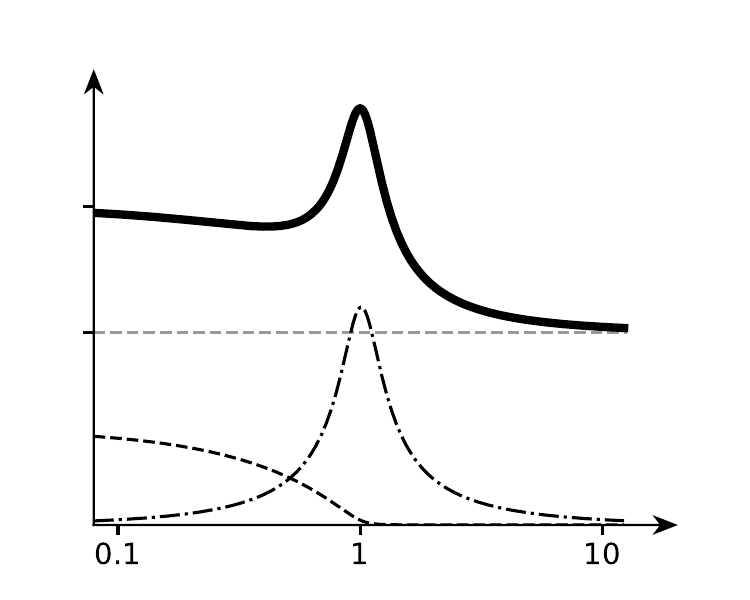}};

 \node[rectangle] (a) at (-5.4,2.7) {$\| \proj_{>0} f_* \|_{L^2}^2$};
 \node[rectangle] (a) at (-5.4,0.8) {$\| \proj_{>1} f_* \|_{L^2}^2$};
 \node[rectangle] (a) at (-5.4,-0.2) {$\| \proj_{>2} f_* \|_{L^2}^2$};
 \node[rectangle] (a) at (-5.4,-2) {$\| \proj_{>3} f_* \|_{L^2}^2$};
 \node[rectangle] (a) at (-5.4,-3.1) {$\| \proj_{>4} f_* \|_{L^2}^2$};
 
 \node[rectangle] (a) at (5.2,-3.7) {$\frac{\log(n)}{\log(d)}$};
 \node[rectangle] (a) at (0,-4.5) {$\kappa$};
 
  \node[rectangle] (a) at (7,-0.5) {$\psi$};
  
\node[rectangle] (a) at (10,0.5) {\footnotesize $\frac{n}{d^\ell / \ell! }$};

\node[rectangle] (a) at (2.6,3.5) {\footnotesize $ \| \proj_{>\ell-1} f_* \|_{L^2}^2$};

\node[rectangle] (a) at (4.9,2.45) {\footnotesize $ \| \proj_{>\ell} f_* \|_{L^2}^2$};

\node[rectangle] (a) at (4.9,1.25) {\footnotesize $ \cB(\psi,\zeta_\ell)$};

\node[rectangle] (a) at (7.9,1.25) {\footnotesize $ \cV(\psi,\zeta_\ell)$};
 
\draw (3.5, -1.8) circle (0.8);
\draw (2.74, -1.54) -- (3.9, 4.9);
\draw (3.9, -2.49) -- (10.2, -0.3);

\end{tikzpicture}
\vspace{-20pt}
\caption{A cartoon illustration of the test error of KRR in the polynomial
 high-dimensional scaling $n/d^\kappa \to \psi$, as $n,d \to \infty$, for any 
 $\kappa,\psi \in \reals_{>0}$. The test error follows a staircase, with peaks that
  can occur at each $\kappa =\ell \in \naturals$, depending on the effective 
  regularization $\zeta_\ell$ and effective signal-to-noise ratio $\text{SNR}_\ell$ 
  at that scale. \label{fig:illustation_KRR}}

\end{figure}

The next theorem characterizes the risk of KRR up to a vanishing constant in the
 high-dimensional polynomial regime. For $\ell \in \naturals$, we denote by 
 $\proj_{\leq \ell} : L^2 \to L^2$ the orthogonal projector onto the subspace 
 of polynomials of degree at most $\ell$, $\proj_{>\ell} := \id - \proj_{\leq \ell}$
  and $\proj_{\ell} := \proj_{\leq \ell}  \proj_{> \ell - 1}$. Further, $o_{d,\prob}$ 
  will denote the standard little-o in probability: $h_1 (d) = o_{d,\prob}(h_2(d))$ if 
  $h_1 (d) / h_2 (d)$ converges to zero in probability.
\begin{theorem}\label{thm:RiskKRR}
Assume the data $\{( y_i , \bx_i ) \}_{i \leq n}$ are distributed according to model \eqref{eq:DistKRRsetting} and the kernel function $h$ satisfies the genericity condition \eqref{eq:genericity_KRR}. 

\begin{itemize}
\item[(a)] If $d^{\ell +\delta} \leq n \leq d^{\ell +1 - \delta}$ for some integer $\ell$ and constant $\delta >0$,  then there exists a constant $\lambda_\star = \Theta_d (1)$ such that for any regularization parameter $\lambda \in [ 0 , \lambda_\star ]$, we have (cf. Eq.~\eqref{eq:TestErrorKRR})
\begin{align}
R_{\KRR} ( f_\star ; \bX , \by , \lambda ) = \| \proj_{>\ell} f_\star \|_{L^2}^2 + o_{d,\prob} (1) \cdot \big( \| f_\star \|_{L^2}^2 + \tau^2 \big) \, .
\label{eq:boundKRR}
\end{align}
Furthermore, no kernel method with dot-product kernel can do better (i.e. have smaller risk).

\item[(b)] If $n/ (d^\ell / \ell! ) \to \psi$ for some integer $\ell$ and constant $\psi >0$, then denoting $\zeta_\ell = (\lambda + c_{>\ell})/c_{\ell}$ the effective regularization at level $\ell$ with $c_{> \ell} = \sum_{k >\ell} c_k$, we have
\begin{equation}
\begin{aligned}
R_{\KRR} ( f_\star ; \bX , \by , \lambda ) =&~ \|\proj_\ell f_\star \|_{L^2}^2 \cdot \cB (\psi , \zeta_\ell ) + \big( \| \proj_{>\ell} f_\star \|_{L^2}^2 + \tau^2 \big) \cdot \cV (\psi , \zeta_\ell )  + \| \proj_{>\ell} f_\star \|_{L^2}^2 \\ 
&~  + o_{d,\prob} (1) \cdot \big( \| f_\star \|_{L^2}^2 + \tau^2 \big) \, ,
\end{aligned}
\label{eq:boundKRR_integer}
\end{equation}
where the definitions of $\cB (\psi , \zeta_\ell )$ and $\cV (\psi , \zeta_\ell )$ can be found in \cite{misiakiewicz2022spectrum}. 
\end{itemize}

\end{theorem}
\begin{remark}
Part $(a)$ of the above theorem was proven in \cite{ghorbani2021linearized} and was later generalized
to other RKHS in \cite{mei2022generalization} under a `spectral gap' assumption. Part $(b)$ was proven in \cite{misiakiewicz2022spectrum,hu2022sharp,xiao2022precise}.
\end{remark}

In words, Eq.~\eqref{eq:boundKRR} implies that $\hat f_\lambda$ only fits the projection 
of $f_\star$ onto low degree polynomials: if $d^{\ell} \ll n \ll d^{\ell +1}$, KRR learns 
the best degree-$\ell$ polynomial approximation to $f_\star$ and nothing else. Equivalently,
we can decompose 
\begin{align}
\label{eq:SolKRRbenignOver}
\hat f_\lambda = \proj_{\leq \ell} f_\star + \Delta \, ,
\end{align}
where 
$\| \Delta \|_{L^2} = (\| f_\star \|_{L^2} + \tau^2 ) \cdot o_{d,\prob} (1)$ does not contribute to the test error.
For example, this theorem implies that if $n \leq d^{1.99}$, KRR can only fit the linear component of $f_\star$. Each time $\log (n) / \log(d)$ crosses an integer value, KRR learns one more degree polynomial approximation: the risk presents a \emph{staircase decay} when $n$ increases, with peaks that can occur at $n = d^\ell / \ell!$. This phenomena is illustrated with a cartoon in Figure \ref{fig:illustation_KRR}.

As mentioned above, the conclusions of Theorem \ref{thm:RiskKRR} apply to any 
rotationally invariant kernel (under the genericity assumption). Notably,
they apply to the neural tangent kernel of 
fully-connected networks of any depth (as explained in Remark \ref{rmk:multilayer_NTK}). 
In the linear regime and in high dimension, depth does not appear to
play an important role for fully-connected neural networks.

\begin{figure}
\begin{center}
\includegraphics[width=.59\textwidth]{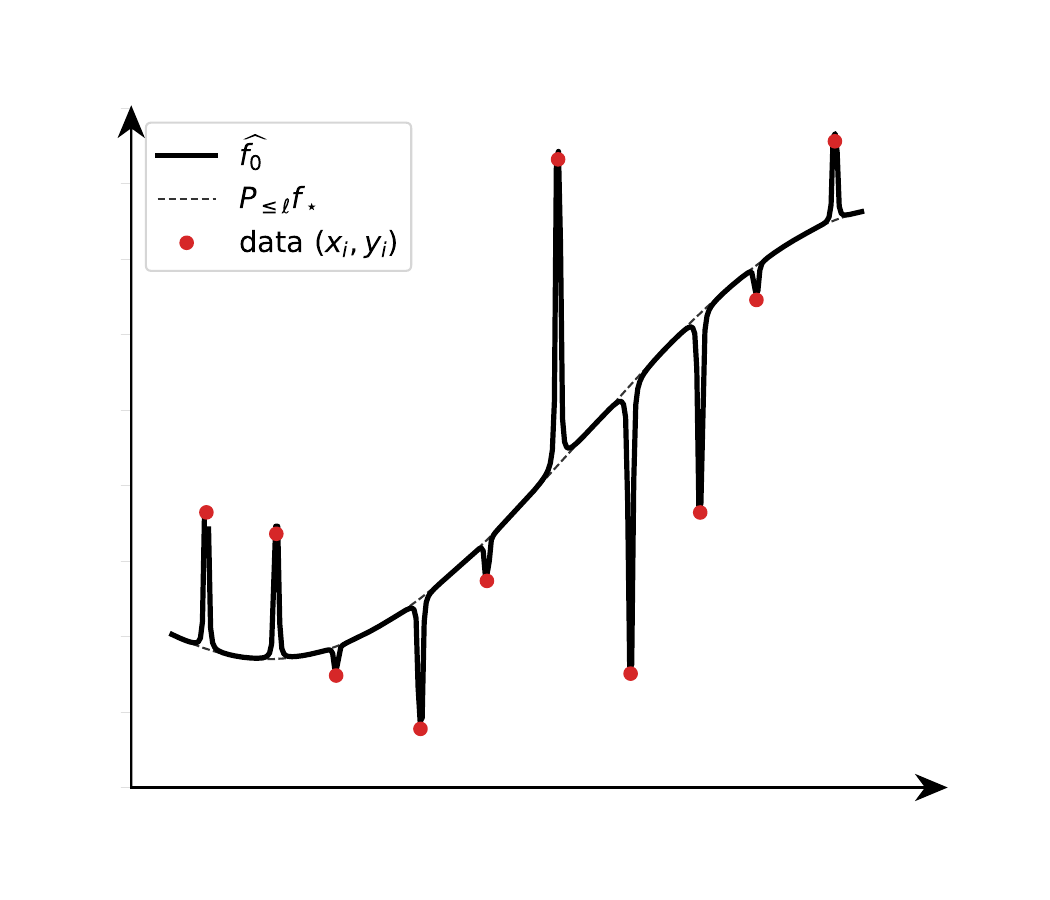}
\put(-16,27){$\bx$}
\put(-260,115){$y$}
\end{center}
\vspace{-20pt}
\caption{A cartoon illustration of the benign overfitting phenomenon in kernel
 ridgeless regression. This cartoon is supposed to capture the qualitative behavior 
 of kernel min-norm interpolators in high dimension. The interpolator decomposes in the sum of
 a smooth part, that captures the best low-degree approximation of the target, and a spiky 
 part that interpolates the noisy data. \label{fig:BO_lectures}}
\end{figure}

%
%
%

\begin{remark}[Benign Overfitting] As discussed in Section \ref{sec:linear_regime},
 one particularly interesting solution is the case $\lambda = 0+$:
\begin{equation}\label{eq:kernelridgeless}
\hat f_0 := \argmin_{f \in \cH} \Big\{ \| f \|_{\cH} \; : \; f (\bx_i) = y_i\, , \; \forall i\in [n ] \Big\} \, ,
\end{equation}
which corresponds to the \emph{minimum-norm interpolating solution}. 
 In this case, the KRR solution perfectly interpolates the noisy data
   $y_i = f_\star ( \bx_i) + \eps_i$. Problem \eqref{eq:kernelridgeless} is sometimes referred to as kernel `ridgeless' regression following \cite{liang2020just}.
   
Theorem
 \ref{thm:RiskKRR} establishes that KRR with $\lambda =  0+$ is optimal among all kernel
  methods when $d^\ell \ll n \ll d^{\ell+1}$, in the sense that it achieves the best 
  possible test error. This is another example of \emph{benign overfitting} where 
  interpolation does not harm generalization, as described in Section \ref{sec:bounds_BO}.
   Recalling Eq.~\eqref{eq:SolKRRbenignOver}, we can decompose $\hat f_{0} = \proj_{\leq \ell} f_\star + \Delta$ where  
\begin{itemize}
\item[$1)$] $\proj_{\leq \ell} f_\star$ is a smooth component good for prediction;
\item[$2)$] $\Delta$ is a spiky component useful for interpolation, 
$\Delta (\bx_i) = \proj_{>\ell} f_\star (\bx_i ) + \eps_i$, but which does not 
contribute to the test error, since $\| \Delta \|_{L^2} \ll 1$.
\end{itemize}
We illustrate benign overfitting in kernel regression with a cartoon in Figure 
\ref{fig:BO_lectures}. 
\end{remark}

\begin{remark}[Multiple peaks in the risk curve] 
For $n \asymp d^\ell$, KRR transitions from not fitting the degree-$\ell$ polynomial 
components at all for $n \ll d^\ell$ to fitting all degree-$\ell$ polynomials when 
$n \gg d^\ell$. The behavior between these two regimes, for  $n \asymp d^\ell$,
is more complex. Because of the degeneracy of the kernel operator eigenvalues at that scale, 
the spectrum of the kernel matrix follows a shifted and rescaled Marchenko-Pastur distribution.
 This can lead to a peak in the risk curve, whenever the effective regularization 
 $\zeta_\ell$ or the effective signal-to-noise ratio
  $\text{SNR}_\ell := \| \proj_\ell f_* \|_{L^2}^2 / ( \| \proj_{>\ell} f_* \|_{L^2}^2 + \tau^2)$ 
  are small enough.  We refer the interested reader to \cite{misiakiewicz2022spectrum} 
  for more details.
\end{remark}

\section{Diagonalization of inner-product kernels on the sphere}
\label{sec:diag_inner_prod_kernel}

In the next section we will present some ideas of the proof of Theorem \ref{thm:RiskKRR}.
Before doing that, it is useful to provide some background on the functional space
 $ L^2 ( \S^{d-1} (\sqrt{d}))$ of square-integrable functions on the sphere.
 For a more in-depth presentation, we refer to \cite{szeg1939orthogonal,chihara2011introduction,efthimiou2014spherical}.
 
 We use the orthogonal decomposition 
\begin{align}
 L^2 ( \S^{d-1} (\sqrt{d})) = \bigoplus_{k=0}^\infty \cV_{d,k} \, ,
\end{align}
where $\cV_{d,\ell}$ is the space of degree-$\ell$ polynomials orthogonal to $ \bigoplus_{k=0}^{\ell-1}\cV_{d,k}$
 (orthogonality is with respect to the $L^2$ inner-product $\< f , g \>_{L^2} = \E \{ f (\bx) g (\bx) \}$). 
 Equivalently, $\cV_{d,k}$ is the space of all spherical harmonics of degree $k$, and has dimension 
\begin{align}
\dim ( V_{d,k} ) = B(d,k) := \frac{2 k + d - 2}{k} {{k + d - 3}\choose{k - 1}} \, .
\end{align}
Note that $B(d,k) = (d^k/k !) (1 + o_d (1))$. Consider 
$\{ Y_{ks} \}_{s \leq B(d,k)}$ an orthonormal basis of $\cV_{d,k}$. In particular, we have
 $\< Y_{k u} , Y_{\ell v} \>_{L^2} = \delta_{k \ell} \delta_{u v}$, and the set
  $\{ Y_{ks} \}_{k \geq 0 , s \leq B(d,k)}$ forms a complete orthonormal 
 basis of $ L^2 ( \S^{d-1} (\sqrt{d}))$. 

It will be convenient to introduce the following notations: $\cV_{d,\leq \ell} = 
\bigoplus_{k=0}^\ell \cV_{d,k}$ the space of all polynomials of degree at most $\ell$, and 
its orthogonal complement $\cV_{d, >\ell} = \bigoplus_{k=\ell+1}^\infty \cV_{d,k}$. Further we 
denote by $\proj_{k}$ the orthogonal projection on $\cV_{d,k}$, 
by $\proj_{\leq \ell} = \proj_0 + \ldots + \proj_{\ell}$ 
the orthogonal projection on $\cV_{d,\leq \ell}$, and $\proj_{>\ell} = \id - \proj_{\leq \ell}$
 the orthogonal projection on $\cV_{d,>\ell}$.

\vspace{+15pt}

Consider now a rotationally invariant kernel defined on $\S^{d-1} (\sqrt{d})$, i.e. a 
positive semidefinite kernel\footnote{This means a measurable function such that 
$\sum_{i,j\le m} K(\bx_i , \bx_j) \alpha_i\alpha_j\ge 0$ for any $m$ and any collection of points 
$(\bx_i)_{i\le m}$, and weights $(\alpha_i)_{i\le m}$.}
 $K : \S^{d-1} ( \sqrt{d} ) \times \S^{d-1} (\sqrt{d}) \to \reals$ such that 
 $K(\bx_1 , \bx_2) = h( \< \bx_1 , \bx_2 \> /d)$ for some 
 measurable function $h: [ -1, 1 ] \to \reals$. To the kernel function 
 $K$, we associate its integral operator $\Kop : L^2 ( \S^{d-1} (\sqrt{d})) \to L^2 ( \S^{d-1} (\sqrt{d}))$ defined by
\begin{align}
\Kop f ( \bx_1 ) = \E_{\bx_2 \sim \Unif ( \S^{d-1} (\sqrt{d})) } \big\{ h (\< \bx_1 , \bx_2 \> / d ) f ( \bx_2 ) \big\} \, .
\end{align}
The proof of Theorem \ref{thm:RiskKRR} relies on the eigendecomposition of inner-product 
kernels on the sphere. We know that by the spectral theorem for compact operators, 
the kernel function $K$ can be diagonalized as
\begin{align}
K (\bx_1 , \bx_2 ) = \sum_{j = 1}^\infty \lambda_j^2 \psi_j (\bx_1) \psi_j ( \bx_2)\, ,
\end{align}
where $\{ \psi_j \}_{j \geq 1}$ is an orthonormal basis of $ L^2 ( \S^{d-1} (\sqrt{d}))$, and the $\{ \lambda_j^2 \}_{j \geq 1}$ are the eigenvalues in nonincreasing order $\lambda_1^2 \geq \lambda_2^2 \geq \lambda_3^2 \geq \cdots \geq 0$.

By rotational invariance, the subspaces $V_{d,k}$ are eigenspaces of the kernel operator $\Kop$, i.e. we can write
\begin{align}
\Kop = \sum_{k = 0}^\infty \xi_{d,k}^2 \proj_{k} \, ,\label{eq:Eigendecomposition}
\end{align}
where we now denote $\xi_{d,k}^2$ the eigenvalue associated to the eigenspace $\cV_{d,k}$. 

Since  $\{ Y_{ks}:s\le B(d,k) \}$ form an orthonormal basis of $\cV_{d,k}$, 
we have
\begin{align}
\proj_k f = \sum_{s = 1}^{B(d,k)} Y_{ks}  \<Y_{ks} ,f\>_{L^2}\, .
\end{align}
Equivalently, the projector is represented as an integral operator
\begin{align}
\proj_k f (\bx_1) &= B(d,k)\cdot\E_{\bx_2} \{ Q^{(d)}_k (\<\bx_1 , \bx_2 \>) f (\bx_2 ) \}, ,\\
Q^{(d)}_k (\< \bx_1 , \bx_2 \>) &:= \frac{1}{B(d,k)}\sum_{s = 1}^{B(d,k)} Y_{ks} (\bx_1) Y_{ks} ( \bx_2) \, ,
\end{align}
Note that $Q^{(d)}_k (\,\cdot\,)$ must be a function of the inner product $\< \bx_1 , \bx_2 \>$
by rotational invariance, and must be a polynomial of degree $k$ because so are the $\{ Y_{ks} \}$.
Indeed, this is a special function known as a Gegenbauer polynomial.
The polynomials $\{ Q^{(d)}_k \}_{k\geq 0}$ form an orthogonal basis of $L^2 ([-d,d], \nu_d)$
 where $\nu_d$ is the distribution of $\sqrt{d} x_1$ with $\bx \sim \Unif ( \S^{d-1} (\sqrt{d}))$. 
 
 The following properties of Gegenbauer polynomials follow from the above representation:
 \begin{enumerate}
 \item $Q^{(d)}_k(d) = \E_{\bx}\{Q^{(d)}_k (\< \bx , \bx \>) \}
 =B(d,k)^{-1}\sum_{s = 1}^{B(d,k)} \E\{Y_{ks} (\bx_1)^2\}$. Therefore
 \begin{align}
  Q^{(d)}_k(d) = 1\, .
  \end{align}
  \item Since projectors are idempotent ($\proj_k^2= \proj_k$), it follows that
 \begin{align}
  \E_{\bx}\{Q^{(d)}_k(\<\bx_1,\bx\>)Q^{(d)}_k(\<\bx,\bx_2\>)\big\} = \frac{1}{B(d,k)}\,
  Q^{(d)}_k(\<\bx_1,\bx_2\>)\, .
 \end{align}
  \item In particular 
  \begin{align}
  \<Q^{(d)}_j,Q^{(d)}_{k}\>_{L^2(\tau_d)} =  \frac{1}{B(d,k)}\, \mathbbm{1}_{j=k}\, .
  \end{align}
  \end{enumerate}

Representing the projection operator in Eq.~\eqref{eq:Eigendecomposition},
we get the following diagonalization of inner-product kernels,
\begin{align}\label{eq:eigenIPKernel}
K (\bx_1 , \bx_2 ) = \sum_{k = 0}^\infty \xi_{d,k}^2 \sum_{s = 1}^{B(d,k)} Y_{ks} (\bx_1) 
Y_{ks} ( \bx_2)\, .
\end{align}
In particular, this means that
\begin{align}
\xi_{d,k}^2 = \E_{\bx} \left\{ Q^{d}_k (\sqrt{d} x_1) h( x_1 / \sqrt{d} ) \right\} \,.
\end{align}

When $d \to \infty$,  the measure $\nu_d$ converges weakly to $\normal (0,1)$ and 
therefore $B(d,k)^{1/2} Q^{(d)}_k$ converges to the degree-$k$ normalized Hermite polynomial 
$\He_k/\sqrt{k!}$. The `genericity condition' \eqref{eq:genericity_KRR} 
can be restated as $\xi_{d,k}^2 B(d,k) \to c_k > 0$ as $d \to \infty$.

\section{Proof sketch}
\label{sec:proof_sketch_KRR}

In this section, we outline the proof of Eq.~\eqref{eq:boundKRR} in Theorem 
\ref{thm:RiskKRR}.(a).  The proof crucially depends on controlling the empirical kernel matrix 
$\bK_n = ( K (\bx_i , \bx_j ) )_{1 \leq ij \leq n}$. While this is a non-linear (in the data)
 random matrix, which is usually hard to study, $\bK_n$ simplifies in our polynomial 
 high-dimensional regime. We explain below how $\bK_n$ can be approximately decomposed into a
  low-rank matrix (coming from the low-degree polynomials) plus a matrix proportional to the 
  identity 
  (coming from the high-degree non-linear part of the kernel).  

Using the diagonalization of inner-product kernels in Eq.~\eqref{eq:eigenIPKernel}, we can 
decompose the kernel matrix into a low-frequency and a high-frequency component,
\begin{align}
\bK_n = \sum_{k = 0}^\ell \xi_{d,k}^2 \bY_{k} \bY_{k}^\sT + \sum_{k \geq \ell+1} \xi_{d,k}^2\bY_{k} \bY_{k}^\sT =: \bK_n^{\leq \ell} + \bK_n^{>\ell} \, ,
\end{align}
where $\bY_{k} = ( Y_{ks} ( \bx_i)  )_{i \leq n , s \leq B(d,k)} \in \reals^{n \times B(d,k)}$ is
 the matrix of degree-$k$ spherical harmonics evaluated at the training data points.  We claim that 
 the following properties hold:
\begin{description}
\item[Low-frequency part:] $\bK_n^{\leq \ell}$ has rank at most $B(d,0) + \ldots + B(d, \ell ) = \Theta_d (d^\ell)$ much lower than $n$ (recall that we assumed $n \geq d^{\ell +\delta}$). Furthermore, one can show that for any $k \leq \ell$,
\begin{align}
\big\| n^{-1} \bY_{k}^\sT \bY_k  - \id_{B(d,k)} \big\|_{\op} =  o_{d,\prob} (1)\, .
\end{align} 
Therefore, we have 
\begin{align}
\sigma_{\min} \big( \bK_n^{\leq \ell} \big) \geq   \min_{k \leq \ell} \left\{ \xi_{d,\ell}^2 \sigma_{\min} \left(\bY_{k} \bY_{k}^\sT  \right) \right\} = \Theta_{d,\prob} (n d^{-\ell}) = \Theta_{d,\prob} (d^\delta) \, ,
\end{align}
where we used $n \geq d^{\ell + \delta}$. Hence $\bK_n^{\leq \ell}$ corresponds to a low rank 
matrix with diverging eigenvalues along the low-frequency component (degree at most $\ell$ polynomials) of $\bK_n$.
\item[High-frequency part:]  $\bK_n^{> \ell}$ is approximately proportional to the
identity with $\| \bK_n^{>\ell} - \gamma \id_n \|_{\op} = o_{d,\prob}(1)$, where 
\begin{align}
\gamma = \sum_{k \geq \ell+1} \xi_{d,k}^2 B(d,k) = \sum_{k \geq \ell+1} c_k + o_d(1)\, .
\end{align}
\end{description}

We are now ready to sketch the proof of Theorem \ref{thm:RiskKRR}.(a). Let us decompose the KRR solution into a low-frequency (the projection on degree-$\ell$ polynomials) and a high-frequency component:
\begin{align}
\hat f_\lambda ( \bx) \approx \bK(\bx , \cdot ) \left( \bK_n^{\leq \ell} + (\lambda + \gamma) \id_n \right)^{-1} \by = \hat f_{\lambda, \leq \ell} ( \bx)  + \hat f_{\lambda, > \ell} ( \bx) \, ,
\end{align}
where 
\begin{align}\label{eq:KRR_sol_self_induced}
\hat f_{\lambda, \leq \ell} ( \bx) :=&~ \proj_{\leq \ell} \hat f_\lambda ( \bx)  = \bK_{\leq \ell} (\bx , \cdot ) \left( \bK_n^{\leq \ell} + (\lambda + \gamma) \id_n \right)^{-1} \by \, ,\\
\hat f_{\lambda, > \ell} ( \bx) :=&~ \proj_{> \ell} \hat f_\lambda ( \bx)  = \bK_{> \ell} (\bx , \cdot ) \left( \bK_n^{\leq \ell} + (\lambda + \gamma) \id_n \right)^{-1} \by \, .
\end{align}
The risk can be written as the sum of contributions along $\proj_{\leq \ell}$ and $\proj_{>\ell}$:
\begin{align}\label{eq:decompoRiskKRR}
\E \left\{ \big( f_{\star} (\bx) - \hat f_{\lambda} (\bx) \big)^2 \right\} = \underbrace{\E \left\{ \big( \proj_{\leq \ell} f_{\star} (\bx) - \hat f_{\lambda, \leq \ell} (\bx) \big)^2 \right\}}_{({\rm I})} + \underbrace{\E \left\{ \big( \proj_{> \ell} f_{\star} (\bx) - \hat f_{\lambda, > \ell} (\bx) \big)^2 \right\}}_{({\rm II})} 
\end{align}
We can bound these two terms separately:
\begin{description}
\item[Term $({\rm I})$:] This term is equivalent to the test error of doing kernel ridge regression with kernel $K_{\leq \ell}$, regularization parameter $\lambda+\gamma$, target function $\proj_{\leq \ell} f_{\star} (\bx)$ and data $y_i = \proj_{\leq \ell} f_{\star} (\bx_i) + \Tilde \eps_i$ where $\Tilde \eps_i =  \proj_{> \ell} f_{\star} (\bx_i) + \eps_i$ (note that the noise $\Tilde \eps$ is not independent of $\proj_{\leq \ell} f_{\star} (\bx_i)$ anymore but is still uncorrelated). The dimension of the target space (i.e. $\cV_{d, \leq \ell}$) and the rank of the kernel are now $\Theta_d ( d^\ell)$ much smaller than the number of samples $n$. Hence, term $({\rm I})$ corresponds to the test error of KRR in low-dimension. We can therefore expect (and show) that this error is vanishing:
\begin{align}
({\rm I}) = \| \proj_{\leq \ell} f_\star \|_{L^2}^2 \cdot o_{d,\prob} (1) + \Var (\Tilde \eps_i ) \cdot o_{d,\prob} (1) = \big( \|  f_\star \|_{L^2}^2 + \tau^2 \big) \cdot o_{d,\prob} (1)\, .
\end{align}

\item[Term $({\rm II})$:] Similarly to $\bK_{>\ell}$, we can show that 
\begin{align}
\Big\| n \E_{\bx} \left\{ \bK_{>\ell}  (\bx , \cdot)^\sT \bK_{>\ell}  ( \bx , \cdot ) \right\} - \kappa \cdot \id_n \Big\|_{\op} = o_{d,\prob}(1) \, ,
\end{align}
with $\kappa =  \sum_{k \geq \ell+1} n \xi_{d,k}^4 B(d,k) = \Theta_d ( n d^{-\ell - 1} ) =o_d (1)$ where we used that $n \leq d^{\ell+1 - \delta}$. We can therefore bound
\begin{equation}\label{eq:KRRboundI}
\begin{aligned}
\| \hat f_{\lambda, > \ell}  \|_{L^2}^2 \leq&~ \Big\| \E_{\bx} \left\{ \bK_{>\ell}  (\bx , \cdot)^\sT \bK_{>\ell}  ( \bx , \cdot ) \right\}\Big\|_{\op}  \Big\|  \left( \bK_n^{\leq \ell} + (\lambda + \gamma) \id_n \right)^{-1} \by \Big\|_2^2  \\
\leq &~ o_{d,\prob}(1) \cdot \frac{1}{(\lambda + \gamma)^2} \cdot \frac{\| \by \|_2^2}{n}  = o_{d,\prob} (1) \cdot \big( \| f_{\star} \|_{L^2}^2 + \tau^2 \big) \, ,
\end{aligned}
\end{equation}
where we used by Markov's inequality that 
\begin{align}
n^{-1} \| \by \|_2^2 = O_{d,\prob}(1) \cdot \E \{ n^{-1} \| \by \|_2^2 \} = O_{d,\prob}(1) \cdot \big( \| f_{\star} \|_{L^2}^2 + \tau^2 \big)\, .
\end{align}
Therefore the contribution of the second term is 
\begin{align}\label{eq:KRRboundII}
({\rm II}) = \| \proj_{>\ell} f_{\star} \|_{L^2}^2 +  \big( \|  f_\star \|_{L^2}^2 + \tau^2 \big) \cdot o_{d,\prob} (1)\, .
\end{align}
\end{description}
Combining the bounds \eqref{eq:KRRboundI} and \eqref{eq:KRRboundII} in Eq.~\eqref{eq:decompoRiskKRR} yields part (a) of Theorem \ref{thm:RiskKRR}.

\begin{remark}[Self-induced regularization] \label{rmk:self_induced_KRR}
From the proof of Theorem \ref{thm:RiskKRR}, we see that the high-frequency component $\bK^{>\ell}_n$ 
of the kernel matrix concentrates on identity and plays the role of an additive 
\emph{self-induced ridge regularization} $\gamma > 0$. Hence the effective 
regularization parameter of the KRR solution is $\lambda + \gamma$ 
(see for example Eq.~\eqref{eq:KRR_sol_self_induced}), which is bounded away 
from zero even when taking $\lambda = 0+$.  This explains why kernel ridgeless regression generalizes well even when interpolating noisy training data. 
\end{remark}

\chapter{Random features}
\label{ch:random_features}
The random feature (RF) model is a two-layer neural network in 
which the first layer weights are not trained, and kept equal to their random initialization.
It was initially introduced in \cite{rahimi2007random} as a randomized approximation for
kernel methods. Here we regard it as a particularly simple limit case of linearized neural networks,
cf. Section \ref{sec:GeneralLinearization}.

 We will use the RF model as a toy model to explore two important questions about  neural networks trained 
  in the linear regime:
  \begin{itemize}
\item[$(i)$] How large should
   $N$ be for the generalization error of RF to match the error associated with its
    infinite-width kernel limit ($N \to \infty$) obtained in Chapter \ref{ch:random_matrix}? 
\item[$(ii)$]  What is the impact on the performance of RF when $N$ is not chosen sufficiently large?
\end{itemize}
While the RF model presents a  simpler structure than the (finite width) neural tangent model, 
we expect them to display similar behavior, provided we match the number of parameters 
in the two models. Chapter \ref{ch:neural_tangent} partly confirms this picture.

\section{The random feature model} 

The random feature model is given by
\begin{align}\label{eq:def_RF_model}
f_{\sRF} ( \bx ; \ba ) = \frac{1}{\sqrt{N}} \sum_{i=1}^N a_i \sigma ( \< \bw_i , \bx \> )\, ,
\end{align}
where $\ba = ( a_1 , \ldots , a_N ) \in \reals^N$ are the free parameters that 
are trained, while the weights $(\bw_i)_{i \leq N}$ are fixed random $\bw_i \sim_{iid} \nu$. 
Equivalently, the RF model can be seen as training the second layer weights of a two-layer neural network, while keeping the first-layer weights fixed. 

Recall the definition of the featurization map $\phi_{\sRF} : \reals^d \to \reals^N$ given by
\begin{align}\label{eq:featureMap_RF}
\bphi_{\sRF}(\bx) = \frac{1}{\sqrt{N}}  [\sigma(\<\bw_{1},\bx\>);\dots; \sigma(\<\bw_{N},\bx\>)] \, ,
\end{align}
so that  $f_{\sRF} ( \bx ; \ba) = \< \ba , \phi_{\sRF} (\bx) \>$. We are interested in the ridge regression solution
\begin{align}
\hat \ba (\lambda) =&~ \argmin_{\ba \in \reals^N} \left\{ \sum_{i=1}^n \big( y_i - f_{\sRF} ( \bx_i , \ba )  \big)^2 + \lambda \| \ba \|_2^2 \right\} \\
=&~ \argmin_{\ba \in\reals^N}
\Big\{\big\|\by-\bPhi\ba\big\|_2^2+ \lambda \|\ba\|_2^2 \Big\}\, ,\label{eq:RFRidgeRegProb}
\end{align}  
where the design matrix $\bPhi = \big[ \bphi_{\sRF}(\bx_1)^\sT , \ldots , \bphi_{\sRF}(\bx_n )^\sT \big]^\sT \in \reals^{n \times N}$ has entries $\Phi_{ij} = \frac{1}{\sqrt{N}} \sigma ( \< \bx_i , \bw_j \>)$. Throughout this chapter, we will consider sampling $\bw_i \sim_{iid} \Unif ( \S^{d-1} )$. 

Let us emphasize here the reason why the RF model is much simpler to study than the neural 
tangent model. The entries of the featurization map \eqref{eq:featureMap_RF} (and the columns of $\bPhi$) are iid with respect to the randomness in $\bw_i$. On the other hand, in the case of the neural tangent model, the entries of the featurization map $\bphi_{\sNT}(\bx) = (Nd)^{-1/2}
[\sigma'(\<\bw_{1},\bx\>)\bx^{\sT};\dots; \sigma'(\<\bw_{N},\bx\>)\bx^{\sT}]$ are only block
 independent with respect to the randomness in $\bw_i$. Tracking the dependency structure 
 in the design matrix makes the analysis of the neural tangent model harder. 
 See Chapter \ref{ch:neural_tangent} for an example of such an analysis. 

\section{Test error in the polynomial high-dimensional regime}

In Chapter \ref{ch:random_matrix}, we saw that as $N \to \infty$, the ridge regression solution converges to the kernel ridge regression solution with inner-product kernel $K_{\sRF}$. What is the impact of finite $N$ on the generalization error of the RF model?

We consider the same setting as in Section \ref{sec:KRR_test_error}. Assume we are given i.i.d.~samples $\{ ( y_i , \bx_i ) \}_{i \leq n}$ with 
\begin{align}
y_i  = f_\star ( \bx_i) + \eps_i\, , \qquad \bx_i \sim \Unif \big( \S^{d-1} (\sqrt{d} )\big)\, ,
\label{eq:DistKRRsetting}
\end{align}
where $f_\star \in L^2 := L^2 ( \S^{d-1} (\sqrt{d} ) )$ and the noise $\eps_i$ is independent of $\bx_i$, with $\E \{ \eps_i \} =0$, $\E \{ \eps_i^2 \} = \tau^2$. 

We consider the RF model as in Eq.~\eqref{eq:def_RF_model}. We will assume the following `genericity' condition on the activation function $\sigma$:  \emph{(i)} $|\sigma (x ) | \leq c_0 \exp ( c_1 |x| )$ for some constants $c_0, c_1 >0$, and \emph{(ii)} for any $k \geq 0$, $\sigma$ has non-zero Hermite coefficient $\mu_k (\sigma ) := \E_{G} \{ \sigma (G) \He_k (G) \} \neq 0$ (where $G \sim \normal (0,1)$).

We are interested in the random feature ridge regression (RFRR) solution 
\begin{align}
\hat f_{\sRF} ( \bx ; \hat \ba (\lambda ) ) = \phi_{\sRF} ( \bx)^\sT \bPhi^\sT \big( \lambda \id_n + \bPhi \bPhi^\sT \big)^{-1} \by \, ,
\end{align}
and consider the excess test error with square loss which we denote by
\begin{align}
R_{\sRF} ( f_\star ; \bX , \by , \bW, \lambda ) := \E_{\bx} \left\{ \big( f_\star ( \bx) - f_{\sRF} ( \bx ; \hat \ba (\lambda) ) \big)^2 \right\} \, .
\label{eq:TestErrorRFRR}
\end{align}

\begin{theorem}\label{thm:RiskRFRR}
Assume $d^{\ell_1 +\delta} \leq n \leq d^{\ell_1 +1 - \delta}$, $d^{\ell_2 +\delta} \leq N \leq d^{\ell_2 +1 - \delta}$, $\max ( N / n , n / N)\geq d^\delta$ for some integers $\ell_1, \ell_2$ and constant $\delta >0$. Denote $\ell = \min (\ell_1 , \ell_2)$. Further assume that $\sigma$ verifies the genericity conditions. Then there exists a constant $\lambda_\star = \Theta_d ( (N/n) \wedge 1 )$ such that for any regularization parameter $\lambda \in [ 0 , \lambda_\star ]$ and all $\eta >0$, we have (cf. Eq.~\eqref{eq:TestErrorRFRR})
\begin{align}
R_{\sRF} ( f_\star ; \bX , \by , \bW, \lambda ) = \| \proj_{>\ell} f_\star \|_{L^2}^2 + o_{d,\prob} (1) \cdot \big( \| f_\star \|_{L^{2+\eta}}^2 + \tau^2 \big) \, .
\label{eq:boundRFRR}
\end{align}
\end{theorem}

In this theorem, the number of neurons $N$ and the number of samples $n$ play symmetric roles. 
The test error of RFRR is determined by the minimum of $N$ and $n$, as long as $N$, $n$ and
 $d^\ell$ for integers $\ell$ are well separated. 
 
 We can distinguish two regimes:
\begin{description}
\item[Underparametrized regime:] For $N \ll n$ (less parameters than number of samples), the test error is limited by the number of random features and the approximation error dominates, i.e.,
\begin{equation}\label{eq:def_app_error}
\begin{aligned}
R_{\sRF} ( f_\star ; \bX , \by , \bW, \lambda )  =&~ R_{\sApp} (f_\star ; \bW )+ o_{d,\prob}(1) \, , \\ 
R_{\sApp} (f_\star ; \bW ) :=&~\inf_{\ba \in \reals^N} \E_\bx \left\{ \big( f_\star (\bx) - \hat f_{\sRF} (\bx ; \ba ) \big)^2 \right\} \, ,
\end{aligned}
\end{equation}
which corresponds to the best fit of $f_\star$ by a linear combination of $N$ random features.
 If $d^\ell \ll N \ll d^{\ell +1}$, we have $\hat f_{\sRF} \approx \proj_{\leq \ell} f_\star$ and 
 the model fits the best degree-$\ell$ polynomial approximation to $f_\star$ and nothing else.
  
  To get an intuitive picture of this phenomenon, notice that the model $\hat f_{\sRF}$ is contained in the subspace 
  ${\rm span} \{ \sigma ( \< \bw_i , \,\cdot\, \>): i\leq N \} \subset L^2 ( \S^{d-1} (\sqrt{d}))$ of dimension $N$. By parameter-counting, in order to approximate the space of degree-$\ell$ polynomials of dimension $\Theta_d (d^\ell)$, we need the number of parameters to be $\Omega_d ( d^\ell)$, which matches the requirement in our theorem.

\item[Overparametrized regime:] For $n \ll N$ (more parameters than number of samples), the test error is limited by the number of samples and the statistical error dominates, i.e.,
\begin{equation}\label{eq:def_stat_error}
R_{\sRF} ( f_\star ; \bX , \by , \bW, \lambda )  = R_{\KRR} ( f_\star ; \bX , \by , \lambda ) + o_{d,\prob}(1) \, ,
\end{equation}
where $R_{\KRR}$ is the risk of KRR with kernel $K_{\sRF}$ (corresponding to the infinite-width limit $N\to \infty$) as obtained in Theorem \ref{thm:RiskKRR}.
 If $d^\ell \ll n \ll d^{\ell+1}$, we have $\hat f_{\sRF} \approx \proj_{\leq \ell} f_\star$. Again, this matches the parameter-counting heuristic: the information-theoretic lower bound to learn the space of degree-$\ell$ polynomials is $\Omega_d (d^\ell)$ samples.

\end{description}

\begin{figure}[t]

\centering
 \includegraphics[width=.99\textwidth]{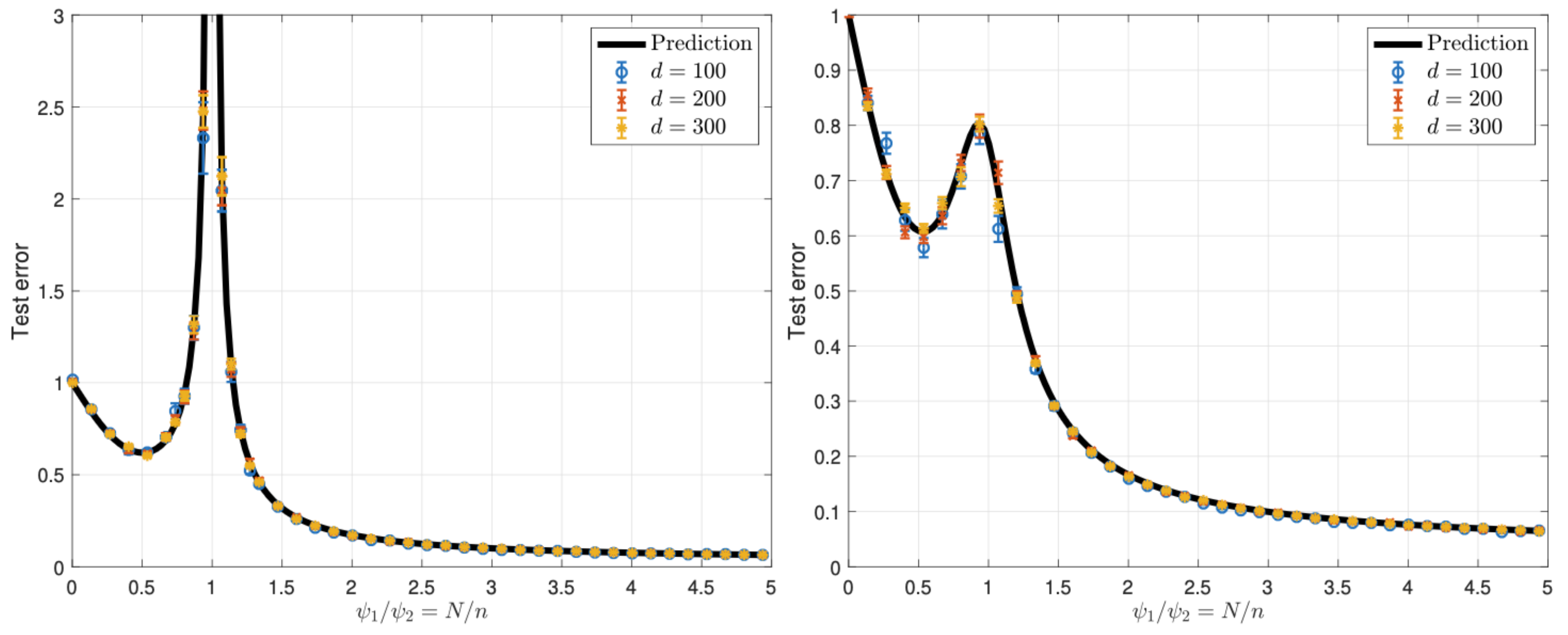}

  \caption{Double descent in the test error of random feature 
  ridge regression with ReLu activation ($\sigma = \max \{ 0, x\}$). The data are generated 
  using $y_i = \< \bbeta , \bx_i \>$ with $\| \bbeta \|_2 = 1$ and $\psi_2 = n/d = 3$. 
  The regularization parameter is chosen to be $\lambda = 10^{-8}$ (left frame) and 
  $\lambda = 10^{-3}$ (right frame). The continuous black lines correspond to 
  theoretical predictions. The colored 
  symbols are numerical simulations averaged over 20 instances and for several dimensions 
  $d$. (Figure from \cite{mei2019generalization}.)}\label{fig:DD}
\end{figure}

To summarize, denoting $R_{\sRF} ( n , N )$ the test error achieved by the RF model with $n$ samples and $N$ neurons, as long as $n$, $N$ and $d^\ell$ for integers $\ell$ are well separated, we have
\begin{align}
R_{\sRF} ( n , N ) = \max \{ R_{\sRF} ( n , \infty ), R_{\sRF} (\infty , N)  \} + o_{d,\prob}(1)\, ,
\end{align}
where $R_{\sRF} (\infty , N)$ corresponds to the approximation error \eqref{eq:def_app_error} and $R_{\sRF} (n , \infty)$ to the statistical error \eqref{eq:def_stat_error}.

Before sketching the proof of Theorem \ref{thm:RiskRFRR}, let us make three further comments:
\begin{description}
\item[Optimal overparametrization.] In the RF model, we are free to choose the number of neurons 
$N$ and it is interesting to ask the following question: given a sample size $n$, how large
 should we choose $N$? Theorem \ref{thm:RiskRFRR} shows that the test error decreases until
  $N \approx n$, and then stays roughly the same as long as $N \geq n^{1 +\delta}$ (for some $\delta >0$,
   although this specific condition is mainly due to our proof technique) and achieves the same 
   test error as the limiting kernel method $N = \infty$ (Theorem \ref{thm:RiskKRR}).
    Further overparametrization does not improve 
    the risk.

\item[Benign overfitting.] The flipside of the last remark is that we can keep
increasing $N$ above $n^{1+\delta}$ (hence overparametrize the model by an arbitrary amount),
without deteriorating the generalization error. 

\item[Optimality of interpolation.] In the overparametrized regime, the 
minimum-norm interpolating solution $\lambda = 0+$ achieves the best test error. 
Hence, similarly to KRR as discussed in Chapter \ref{ch:random_matrix}, the random
 feature model displays the \emph{benign overfitting} property. The mechanism for 
 RF is very similar to the KRR case (Remark \ref{rmk:self_induced_KRR}): the high-degree 
 non-linear part of the activation function $\sigma$ acts as noise in the features and
  behaves as an effective \emph{self-induced regularization} (see next section).
\end{description}

\section{Double descent and proportional asymptotics}

In Theorem \ref{thm:RiskRFRR}, we assumed that the values of $n$ and $N$ are sufficiently
 well separated. Indeed, when $N $ is roughly equal to $n$, the design matrix $\bPhi$ becomes 
 nearly square and its condition number becomes large. This leads to a peak in the
  test error at the interpolation threshold\footnote{The name `interpolation threshold' 
  comes from the fact that as long as $N\geq n$, the RF model has enough parameters to 
  interpolate the training data. If $\lambda =0+$, then the training error becomes 
  $0$ at $N=n$.} $N=n$. This has been called the `double descent' phenomenon, and we already encountered it in Section \ref{sec:Latent}.
  
   In the present context, the interpretation of this phenomenon is even clearer.
  As the number of parameters increases, the test error first follows the classical 
  U-shaped curve, with error initially decreasing and then increasing as it approaches the interpolation threshold.
    This is in line with the classical picture of bias-variance tradeoff, where
     increasing the model complexity reduces model misspecification (i.e. the model 
     is too simple to fit the data) but increases sensitivity to statistical fluctuations
      (i.e. the model may overfit to noise in the training data). However, unlike 
      the traditional U-shaped curve, the test error descends again beyond 
      the interpolation threshold, and the optimal test error is achieved when the number of 
      parameters is significantly larger than the number of samples.   

The double descent phenomenon in the random feature model has been studied in the 
proportional asymptotic regime (when $N,n,d\to \infty$ with $N/ d \to \psi_1$ and $n/d \to \psi_2$) 
in \cite{mei2019generalization}. Figure \ref{fig:DD} reports the test error of random 
feature ridge regression obtained numerically 
for a fixed value of $\psi_2 = 3$ and two different regularization parameters $\lambda = 10^{-8}$
 (left) and $\lambda = 10^{-3}$ (right). 
 
 Continuous lines correspond to the theoretical predictions from
 \cite{mei2019generalization}, that take the form sketched below.
 \begin{theorem}\label{thm:main_theorem_propottional}
Let the activation function $\sigma$ satisfy the genericity and growth assumptions stated above.
Define
\begin{align}
\ob_0 := \E\{ \sigma(G)\}, ~~~~ \ob_1:=\E\{G\sigma(G)\},~~~~
\ob_\star^2 := \E\{\sigma(G)^2\}- \ob_0^2- \ob_1^2 \, ,
\end{align}
where expectation is with respect to $G\sim\normal(0,1)$. 
Assuming $0 < \ob_0^2, \ob_1^2, \ob_{\star}^2 < \infty$, define  $\ratio$ by
\begin{align}
\ratio := \frac{\ob_1}{ \ob_\star}\, . 
\end{align}
 and consider 
proportional asymptotics $N/d\to\psi_1$, $n/d\to\psi_2$. Finally, let the regression function  
$\{f_d \}_{d \ge 1}$ be such that $f_d(\bx) = \beta_{d, 0} +\<\bbeta_{d, 1},\bx\> + f_d^{\sNL}(\bx)$, where
$f_d^{\sNL}(\bx)$ is a centered rotationally invariant
Gaussian process indexed by $\bx \in \S^{d-1}(\sqrt d)$, satisfying $\E_{\bx}\{f_d^{\sNL}(\bx)\bx \}=\bfzero$.
Assume 
 $\E\{ f_d^{\sNL}(\bx)^2\} \to \normf_\star^2$, $\|\bbeta_{d, 1}\|_2^2\to F_1^2$.

Then for any value of the regularization parameter $\lambda > 0$, the asymptotic prediction error 
of random feature ridge regression satisfies
\begin{align*}
\E_{\bX, \bW, \beps, f_d^{\sNL}} 
\Big\vert
 R_{\sRF}(f_d; \bX, \bW, \lambda) - \Big[  \normf_1^2 \cuB(\ratio, \psi_1, \psi_2, \lambda/\ob_{\star}^2)+ 
(\tau^2 + \normf_\star^2) \cuV(\ratio, \psi_1, \psi_2, \lambda/\ob_{\star}^2) + 
\normf_\star^2  \Big] \Big\vert = o_{d}(1)\, ,
\end{align*}
where $\E_{\bX, \bW, \beps, f_d^{\sNL}}$ denotes expectation with respect to data covariates $\bX$, 
first layer weigth vectors $\bW$, data noise $\beps$, and $f_d^{\sNL}$ the nonlinear part of the true regression function (as a Gaussian process).
The functions $\cuB, \cuV$ are explicitly given. 
\end{theorem}

Overall, the behavior of random feature ridge regression (RFRR) in high dimension 
can be summarized as follows: as $n$ increases while $N$ is kept fixed, the test 
error of RFRR initially concentrates on the statistical error (test error of kernel ridge regression,
 with $N = \infty$) for $n \ll N$, then peaks at the interpolation threshold $n=N$ 
 (the double descent phenomenon), and finally saturates on the approximation
  error (test error with $n =\infty$) for $n \gg N$.

\section{Polynomial asymptotics: Proof sketch}

In this section we outline the proof of Theorem \ref{thm:RiskRFRR}.

Recall that the random feature ridge regression solution is given by
\begin{align}
\hat f_{\sRF} ( \bx ; \hat \ba (\lambda ) ) = \phi_{\sRF} ( \bx)^\sT \bPhi^\sT \big( \lambda \id_n + \bPhi \bPhi^\sT \big)^{-1} \by \, .
\end{align}
To prove Theorem \ref{thm:RiskRFRR}, we need to control the behavior of the feature matrix $\bPhi \in \reals^{n \times N}$:
\begin{equation}
\bPhi = \begin{bmatrix}
\sigma ( \< \bx_1 , \bw_1 \>) & \sigma ( \< \bx_1 , \bw_2 \>)  & \cdots & \sigma ( \< \bx_1 , \bw_N  \>) \\
\sigma ( \< \bx_2 , \bw_1 \>) & \sigma ( \< \bx_2 , \bw_2 \>)  & \cdots & \sigma ( \< \bx_2 , \bw_N  \>) \\
\vdots & \vdots & & \vdots \\
\sigma ( \< \bx_n , \bw_1 \>) & \sigma ( \< \bx_n , \bw_2 \>)  & \cdots & \sigma ( \< \bx_n , \bw_N  \>) 
\end{bmatrix}\, .
\end{equation}
Similarly to the kernel matrix in Section \ref{sec:proof_sketch_KRR}, we show that in 
the polynomial regime $\bPhi$ can be approximately decomposed into a low-rank matrix 
(coming from the low-degree polynomials up to degree $\ell = \min (\ell_1 , \ell_2)$) plus a 
matrix that is proportional to an orthogonal matrix, whose span is approximately orthogonal to the 
low rank part. This decomposition again rely on an operator diagonalization.

Consider the activation function $\sigma$, seen as an integral operator from $L^2 ( \S^{d-1} (\sqrt{d}))$ to $L^2 ( \S^{d-1} (\sqrt{d}))$ (for convenience, we rescale the argument to have $\bw \sim \Unif ( \S^{d-1} (\sqrt{d}))$):
\begin{align*}
g ( \bw) \to \int \sigma ( \< \bx , \bw \>/\sqrt{d} )  g (\bw) \tau_d (\de \bw) \, ,
\end{align*} 
where $\tau_d$ denotes the uniform measure on $\S^{d-1} (\sqrt{d})$. From assumption \textit{(i)}, we have $\sigma ( \< \cdot, \cdot \>/\sqrt{d} ) \in L^2 (  \S^{d-1} (\sqrt{d})^{\otimes  2} )$ and from the spectral theorem for compact operators, the function $\sigma$ can be diagonalized as 
\begin{align}\label{eq:sigma_diag_general}
\sigma ( \< \bx, \bw \>/\sqrt{d} ) = \sum_{j = 1}^\infty \lambda_j \psi_j ( \bx ) \phi_j (\bw) \, ,
\end{align}
where $(\psi_j)_{j \geq 1}$ and $(\phi_j)_{j \geq 1}$ are two orthonormal bases of $L^2 (\S^{d-1} (\sqrt{d}))$, and the $\lambda_j$ are the eigenvalues with nonincreasing absolute values $|\lambda_1| \geq |\lambda_2| \geq \cdots \geq 0$. By rotational invariance of the operator (see Section \ref{sec:diag_inner_prod_kernel}), $\sigma$ is diagonalizable with respect to the spherical harmonics orthogonal basis
\begin{align}
\sigma ( \< \bx, \bw \>/\sqrt{d} ) = \sum_{k=0}^\infty \xi_{d,k} \sum_{s = 1}^{B(d,k)} Y_{ks} (\bx) Y_{ks} (\bw) \, .
\end{align}
Assumption \textit{(ii)} on the activation $\sigma$ implies that $| \xi_{d,k} | \asymp d^{-k/2}$. 

Using the notations in Eq.~\eqref{eq:sigma_diag_general}, let $\bpsi_k = (\psi_k  (\bx_1) , \ldots , \psi_k ( \bx_n ) )^\sT$ be the evaluation of the $k$-th left eigenfunction at the $n$ data points, and let $\bphi_k = (\phi_k (\bw_1 ) , \ldots , \phi_k (\bw_N ) )^\sT$ be the evaluation of the $k$-th right eigenfunction at the $N$ random first-layer weights. Further, let $k(\ell) := \sum_{k \leq \ell} B(d,k)$. Similarly to Section \ref{sec:proof_sketch_KRR}, we expand $\bPhi$ into a low-frequency and a high-frequency component:
\begin{align}
&\bPhi =  \sum_{j = 1}^\infty \lambda_j \bpsi_j \bphi_j^\sT =: \bPhi_{\leq \ell} + \bPhi_{>\ell} \, , \\
& \bPhi_{\leq \ell}  = \sum_{ j =1}^{k(\ell)} \lambda_j \bpsi_j \bphi_j^\sT =: \bpsi_{\leq \ell} \bLambda_{\leq \ell} \bphi_{\leq \ell}^\sT \, , \qquad \bPhi_{>\ell}  = \sum_{ j =k(\ell)+1}^{\infty} \lambda_j \bpsi_j \bphi_j^\sT\, ,
\end{align}
where $\bLambda_{\leq \ell} = \diag ( \lambda_1 , \ldots, \lambda_{k(\ell)})$, $\bpsi_{\leq \ell} \in \reals^{n \times k (\ell)}$ is the matrix whose $j$-th column is $\bpsi_j$, and $\bphi_{\leq \ell} \in \reals^{N \times k (\ell)}$ is the matrix whose $j$-th column is $\bphi_j$. Recalling $| \xi_{d,k} | \asymp d^{-k/2}$, for $d$ sufficiently large, $\bLambda_{\leq \ell}$ contains exactly all the singular values associated to the spherical harmonics of degree at most $\ell$.

Notice that the matrix $\bPhi$ has the same distribution under the mapping $\bPhi \leftrightarrow\bPhi^\sT$, $n \leftrightarrow N$, $\bx_i \leftrightarrow \bw_j$. Hence, without loss of generality let us consider the overparametrized case $N \geq n^{1+\delta}$ with $d^{\ell+\delta} \leq n \leq d^{\ell +1 - \delta}$. Then we have the following properties:
\begin{description}
\item[Low-frequency part:] $\bPhi_{\leq \ell}/\sqrt{N}$ has rank $k(\ell) = \Theta_d (d^\ell)$ much smaller than $\min(n,N) = n \geq d^{\ell +\delta}$. Furthermore, 
\begin{align}
\big\| \bpsi_{\leq \ell}^\sT \bpsi_{\leq \ell} / n - \id_{k(\ell)} \big\|_{\op} = o_{d,\prob}(1) \, , \qquad \big\| \bphi_{\leq \ell}^\sT \bphi_{\leq \ell} / N - \id_{k(\ell)} \big\|_{\op} = o_{d,\prob}(1) \, ,
\end{align}
and $\sigma_{\min} \big( \bPhi_{\leq \ell} / \sqrt{N} \big)  = \Omega_{d,\prob} ( \min_{k \leq \ell} n | \xi_{d,k} |) = \Omega_{d,\prob} (d^\delta)$. Hence, $\bPhi_{\leq \ell}/\sqrt{N}$ corresponds to a low-rank spiked matrix along the low-frequency components (degree-$\ell$ polynomials).

\item[High-frequency part:] The singular values of $\bPhi_{>\ell}/ \sqrt{N}$ concentrates, i.e., $\big\| \bPhi_{>\ell} \bPhi_{>\ell}^\sT / N - \gamma \id_n \big\|_{\op} = o_{d,\prob} (1)$ where
\begin{align}
\gamma = \sum_{ k \geq \ell+1} \xi_{d,k}^2 B(d,k) \, .
\end{align}
Furthermore, $\| \bPhi_{>\ell} \bphi_{\leq\ell}^\sT / N \|_{\op} = o_{d,\prob} (\gamma^{1/2})$. Hence the high-frequency component $\bPhi_{>\ell}$ behaves similarly to a random noise matrix, independent of $\bphi_{\leq\ell}$.

\end{description}

Based on the above description of the structure of the feature matrix, 
ridge regression with respect to the random features $\sigma ( \< \bw_j , \cdot \>)$ is
 essentially equivalent to doing a regression against a polynomial kernel of degree $\ell$, 
 with $\ell$ depending only on the smaller of the sample size and the network size. The same
  mechanism as in kernel regression appears: the high-frequency part of the activation
   $\sigma$ effectively behaves as noise in the features and can be replaced by an effective ridge regularization $\gamma$ (another example of \textit{self-induced regularization}).

\begin{remark}
Working with the uniform measure over the sphere is particularly convenient 
because one can exploit known properties of spherical harmonics. 
However, the analysis in the polynomial regime can be extended to other distributions
of the covariate vectors $\bx_i$ and of the weights $\bw_{j}$ under appropriate
abstract conditions \cite{mei2022generalization}.
\end{remark}

\chapter{Neural tangent features}
\label{ch:neural_tangent}
In this section we consider the neural tangent model associated to 
a finite-width fully connected two-layer neural network. This model was already introduced
in Section \ref{sec:GeneralLinearization} and captures the behavior of two-layer
networks initialized at random, and trained in the linear/lazy regime.
As throughout these notes, we focus on square loss. 

As discussed in Chapter \ref{ch:random_matrix}, the infinite-width limit of
linearized neural networks corresponds to kernel ridge regression (KRR) with respect
to a certain inner product kernel. Here we want to address the following key question:
\begin{quote}
 How large the width of a two-layer network 
should be for the test error to be well approximated by its infinite-width limit?
\end{quote}

It turns out that the answer to this question is a natural generalization of the one we obtained for the 
random feature model  in the last chapter. Roughly speaking,
if the number of parameters is large compared to the sample size (hence the model is overparametrized), 
then the test error is well approximated by the infinite width test error. Notice
that the total number of parameters in this case is $Nd$, and therefore this condition translates
into
\begin{align}
Nd\gg n\, ,
\end{align}
whereby for the random feature model, the overparametrization condition amounted to $N\gg d$.
%
%
\section{Finite-width neural tangent model}

Recall that the neural tangent model  is defined by 
\begin{align}\label{eq:def_NT _model}
f_{\sNT} ( \bx ; \bb ) = \frac{1}{\sqrt{Nd}} \sum_{i=1}^N \<\bb_i,\bx\> \sigma( \< \bw_i , \bx \> )\, .
\end{align}
Here  $\bb= (\bb_i,\dots,\bb_N)\in\reals^{Nd}$ is a vector of parameters that are learnt from data,
while $(\bw_i)_{i\le N}$ are i.i.d. first-layer weights at initialization, with common distribution 
$\nu$. 
In our analysis, we will take $\nu=\Unif(\S^{d-1})$ to be the uniform measure over the sphere. 
This model can be equivalently described in terms of the  NT featurization map 
\begin{align}
\bphi_{\sNT}(\bx) &= \frac{1}{\sqrt{Nd}} 
[\sigma'(\<\bw_{1},\bx\>)\bx^{\sT};\dots; \sigma'(\<\bw_{N},\bx\>)\bx^{\sT}]\, . 
\end{align}
We then have $f_{\sNT} ( \bx ; \bb) = \< \bb , \phi_{\sNT} (\bx) \>$.

We  are given i.i.d.~samples $\{ ( y_i , \bx_i ) \}_{i \leq n}$
and  are interested in the ridge regression solution
\begin{align}
\hat \bb (\lambda) 
=&~ \argmin_{\bb \in\reals^{Nd}}
\Big\{\big\|\by-\bPhi\bb\big\|_2^2+ \lambda \|\bb\|_2^2 \Big\}\, ,\label{eq:NTRidgeRegProb}
\end{align}  
where we introduced the design matrix $\bPhi\in \reals^{n\times Nd}$ by
\begin{align}
\bPhi :=\left[\begin{matrix}
    \mbox{---}\bphi_{\sNT}(\bx_1)  \mbox{---}\\
      \mbox{---}\bphi_{\sNT}(\bx_2)  \mbox{---}\\
      \vdots\\
      \mbox{---}\bphi_{\sNT}(\bx_n)  \mbox{---}\\
      \end{matrix}\right] \, .
\end{align}

We assume the same model as in the previous chapter, namely 
\begin{align}
y_i  = f_\star ( \bx_i) + \eps_i\, , \qquad \bx_i \sim \Unif \big( \S^{d-1} (\sqrt{d} )\big)\, ,
\end{align}
where $\eps_i$ is noise independent of $\bx_i$, with $\E\{\eps_i \} = 0$, $\E \{\eps_i^2 \} = \tau^2$. 
The corresponding test error is
\begin{align}
R_{\sNT} ( f_\star ; \bX , \by , \bW, \lambda ) :=
 \E_{\bx} \left\{ \big( f_\star ( \bx) - \hat f_{\sNT} ( \bx ; \hat \bb (\lambda) ) \big)^2 \right\} \, .
\end{align}

\begin{remark}[RF+NT]
Recall (cf. Section \ref{sec:GeneralLinearization}) that
the full model obtained by linearizing two-layer neural networks
with respect to both first layer and second layer weights takes 
the form
\begin{align}
f_{\slin}(\bx;\ba,\bb) &= f_0(\bx)+ f_{\sRF}(\bx;\ba)+f_{\sNT}(\bx;\bb)\, ,
\end{align}
where $f_0(\bx)$ is the neural network at initialization.
 
We introduced two simplifications here. First, we dropped the initialization term $f_0(\bx)$.
As already mentioned in Section \ref{sec:GeneralLinearization}, we can indeed endorse
$f_0(\bx)=0$ at the price of doubling the number of neurons and choosing the initialization
 $(\bw_j)_{j\le 2N}$ so that $\bw_{N+j}=\bw_j$ and $a_{N+j} = - a_j$. The analysis of this initialization is 
 the same as the one pursued here.
 
 Second, we drop the term $f_{\sRF}(\bx;\ba)$. This can be shown to have a negligible effect
 in high dimension (for a generic activation function). Informally,  it amounts to reducing the
  number of parameters from $Nd+N$ to $Nd$, which is a negligible change.  
\end{remark}

\begin{remark}[Complexity at prediction time]
So far, our presentation mirrored the one of the RF model in the last 
chapter. However, there is an important practical difference in the complexity of
evaluating the functions $f_{\sRF}(\,\cdot\, ;\ba)$ and $f_{\sNT}(\,\cdot\, ;\bb)$
which is the task we need to perform at prediction (a.k.a. inference) time.

Consider first the RF model. Organizing the weights $\bw_i$ as rows of a matrix
$\bW\in\reals^{N\times d}$, the function $f_{\sRF}(\,\cdot\, ;\ba)$ can be evaluated at a point 
$\bx$ using $O(Nd)$ operations. First, we compute $\bz = \bW\bx$ (which can be done with $O(Nd)$
operations), and then we output $\sum_{i=1}^N a_i\sigma(z_i)$ which can be done in $O(N)$
operations (if evaluating the one-dimensional function $\sigma$ takes $O(1)$ operations).

Next consider the NT model. We arrange the parameters $\bb_i$ into  $\bB\in\reals^{N\times d}$.
Next we evaluate $\bu = \bB\bx$, $\bz = \bW\bx$ and finally output $\sum_{i=1}^N u_i\sigma'(z_i)$.
Under similar assumptions, this results in $O(Nd)$ operations.

At first sight, the two models behave similarly. However, as we 
saw in the last section, the `approximation power' of the RF model is 
controlled by the number of parameters $p=N$. Similarly, in the NT model we will see that 
it is controlled by $p=Nd$. Hence the complexity of prediction 
at constant number of parameters is $O(pd)$ for the RF model compared to $O(p)$
for the NT model.
In other words, the NT model is much simpler to evaluate in high dimension. 
\end{remark}

\begin{remark}[Complexity of SGD training]
Consider next training either the RF or the NT model using stochastic gradient descent
(SGD) under quadratic loss. Of course, in this case the cost function is quadratic and 
we can use faster algorithms than SGD. However, these algorithms are not available, or of common use
for actual non-linear networks, and we hope to gain some insights into those cases.
We assume a batch of size $B$, which we denote by $\{(y_1,\bx_1), \dots (y_B,\bx_B)\}$,
and will just compute the complexity of a single SGD step.

For the RF model the gradient takes the form  (here $\hR_B(\ba)$ is the empirical risk with respect to
the batch)
\begin{align}
\nabla_{\ba}\hR_B(\ba) = -\sigma(\bW\bX^{\sT})\br\, ,
\end{align}
where $\bW\in\reals^{N\times d}$ is the matrix whose $i$-th row is $\bw_i$,
$\bX\in \reals^{B\times d}$ is the matrix whose $j$-th row is $\bx_i$, and $\br\in\reals^B$,
with $r_i = y_i-f_{\sRF}(\bx_i;\ba)$. Assuming $\sigma$ can be evaluated in $O(1)$ time, the above 
gradient can be evaluated in time $O(NdB) = O(pd B)$.

For the NT model, we view the parameters as a matrix $\bb\in\reals^{N \times d}$ whose $i$-th row
corresponds to neuron $i$. We then have
\begin{align}
\nabla_{\bb}\hR_B(\bb) = -\sum_{i=1}^B r_i\left[\begin{matrix}
\mbox{---} \sigma(\bw_1^{\sT}\bx_i)\cdot \bx_i^{\sT}\mbox{---}\\
\mbox{---} \sigma(\bw_2^{\sT}\bx_i)\cdot \bx_i^{\sT}\mbox{---}\\
\vdots\\
\mbox{---} \sigma(\bw_N^{\sT}\bx_i)\cdot \bx_i^{\sT}\mbox{---}\\
\end{matrix}\right]
 \, .
\end{align}
Again, this can be evaluated in time $O(NdB)=O(pB)$. The complexity is significantly smaller than for RF.
\end{remark}

\section{Approximation by infinite-width KRR: Kernel matrix}

The empirical kernel matrix plays a crucial role in ridge regression,
and hence it is useful to begin by analyzing its structure.
We denote the kernel matrix associated to the finite-width NT model
by $\bK_N\in\reals^{Nd\times Nd}$. By definition, its $i,j$-th entry
is $(\bK_N)_{ij} =\<\bphi_{\sNT}(\bx_i),\bphi_{\sNT}(\bx_j)\>$. Explicitly:
\begin{equation}
(\bK_N)_{ij} = \frac{1}{Nd}\sum_{k=1}^N \sigma'(\<\bw_k,\bx_i\> ) \sigma'(\<\bw_k , \bx_j\> ) \< \bx_i, \bx_j \>\, .
\label{eq:KernelDef}
\end{equation}
The corresponding infinite-width kernel matrix is given by the expectation of the above
\begin{equation}
\label{eq:InfiniteWidthKernelDef}
(\bK)_{ij} =  \frac{\langle \bx_i, \bx_j \rangle}{d}\cdot\E_\bw \big\{ \sigma'(\< \bw,\bx_i\>) \sigma'(\<  \bw ,\bx_j\>) \big\}\, .
\end{equation}
Here and below, we denote by $\E_{\bw}$, $\prob_{\bw}$ expectation and probability
with respect to $\bw$ at fixed $(\bx_i:\, i\le n)$. 

By the law of large numbers
(and under very minimal conditions on $\sigma$, e.g. $\E_{\bw}[\sigma ' (\<\bw,\bx_i\>)^2]<
\infty$), we have $\lim_{N\to\infty}(\bK_N)_{ij}= (\bK)_{ij}$ for fixed $i,j$.
 However, we are interested in $N, n, d$ all diverging simultaneously,
 and the convergence of the matrix entries does not provide strong information on the 
 overall matrix (e.g. its eigenvalues), which is our main focus.

The activation function will enter our estimates through the following
parameters
\begin{align}
v_{\ell}(\sigma) := \sum_{k\ge\ell} \frac{1}{k!}\<\He_k,\sigma'\>_{L^2(\normal(0,1))}^2\, .
\end{align}
Here $\<f,g\>_{L^2(\normal(0,1))}= \E\{f(G)g(G)\}$ (where $G\sim\normal(0,1)$) 
is the usual $L^2$ inner product with respect to the Gaussian measure, and $\He_k$
is the $k$-th Hermite polynomial, with the standard normalization 
$\<\He_j,\He_k\>_{L^2(\normal(0,1))}= k!\delta_{jk}$.
We further make the following assumption on $\sigma$. 
\begin{assumption}[Polynomial growth]\label{ass:Sigma}
We assume that $\sigma$ is weakly differentiable with weak derivative $\sigma'$ satisfying 
$|\sigma'(x)|\le B(1+|x|)^B$ for some finite constant $B>0$, and that
 $v_{\ell}(\sigma') > 0$. (The latter is equivalent to $\sigma$ not being a maximum
 degree-$\ell$ polynomial.)
\end{assumption}

We next state a kernel concentration result.
\begin{theorem}[Kernel concentration]\label{thm:invert2}
Let $\gamma= (1-\eps_0)v_{\ell}(\sigma)$ for some constant $\eps_0\in (0,1)$.
 Further, let $\beta>0$ be arbitrary.
Then, there exist constants 
$C', C_0>0$ such that the following hold.
 
Define the event:
\begin{equation}
\cA_\gamma := \big\{ \bK \succeq \gamma \bI_n, ~ \| \bX \|_\op \le 2(\sqrt{n}+\sqrt{d}) \big\}.
\end{equation}
For any $\bX\in\cA_{\gamma}$, we have
\begin{equation}
  \P_\bw \left( \big\| \bK^{-1/2} \bK_N \bK^{-1/2} - \id_n \big\|_\op > 
\sqrt{\frac{(n+d)( \log (nNd))^{C'}}{Nd}} + \frac{(n+d)( \log (nNd))^{C'}}{Nd} \right) \le
d^{-\beta}.\label{eq:MainKernelBound}
\end{equation}
Further if $n\le d^{\ell+1}/(\log d)^{C_0}$, then 
 $\prob(\cA_\gamma)\ge 1- n^{-\beta}$ for all $n$ large enough.
\end{theorem}
\begin{remark}
Note that the event $\cA_{\gamma}$ only depends on the matrix of covariates $\bX$.
Hence, we view $\cA_{\gamma}$ as a set in $\reals^{n\times d}$ and write $\bX\in \cA_{\gamma}$
if $\bX$ satisfies the stated conditions. 
Equation \eqref{eq:MainKernelBound} holds for any such $\bX$, not necessarily random ones
(the probability being only with respect to $\bW$).  

Since $\prob(\cA_\gamma)\ge 1- n^{-\beta}$ under the stated conditions, we also have
that, if $n\le d^{\ell+1}/(\log d)^{C_0}$, then with probability at least
$1-d^{-\beta}$ (both with respect to $\bX$ and $\bW$) 
\begin{equation}\label{ineq:invert2}
\big\| \bK^{-1/2} \bK_N \bK^{-1/2} - \id_n \big\|_\op  \le  \sqrt{\frac{(n+d)( \log (nNd))^{C'}}{Nd}} + 
\frac{(n+d)( \log (nNd))^{C'}}{Nd}.
\end{equation}
\end{remark}

\begin{remark}
A first attempt at proving concentration would be to try to control $\|\bK_N-\bK\|_{\op}$. 
However, this results in suboptimal bounds because $\bK$ has eigenvalues
on several well separated scales. Indeed, we saw in Chapter
\ref{ch:random_matrix} that $\bK$ has one eigenvalue of order $n$, $d$ eigenvalues
of order $n/d$,  approximately $d^k/k!$ eigenvalues of order $n/d^k$ for any 
$k\le \ell$, and $n-O(d^{\ell})$ eigenvalues of order $1$.

In Theorem \ref{thm:invert2} we bound $\| \bK^{-1/2} \bK_N \bK^{-1/2} - \id_n \|_\op$,
hence effectively looking at each group of eigenvalues at different scales.
\end{remark}

Assume for simplicity $n\ge c_1d$ for some constant $c_1$. There is little loss of 
generality in doing so. Indeed,  even if we know that the target function is linear 
$f_*(\bx) = \<\bbeta_*,\bx\>$, it is information theoretically impossible to 
achieve prediction error less that (say) half the error of the trivial predictor $\hf(\bx) =0$,
unless $n\ge c_1d$  for some constant $c_1$.

In this setting the norm bound in Theorem \ref{thm:invert2}
(equivalently, the right-hand side of Eq.~\eqref{ineq:invert2}) is small as soon 
as
\begin{align}
\frac{Nd}{(\log Nd)^C}\gg n \, .\label{eq:OverparametrizationNT}
\end{align}
Modulo the log factors, this is the sharp overparametrization condition $Nd\gg n$.
This is also our first piece of evidence that the relevant control parameter
is the ratio of number of parameters to sample size, rather than number of 
neurons to sample size. 

%
%

\section{Approximation by infinite-width KRR: Test error}

By itself, controlling the kernel matrix as we did in Theorem 
\ref{thm:invert2} does not allow to bound the test error, which involves evaluating
the regression function at a fresh test point. Explicitly, we have 
\begin{align}
\hf(\bx;\lambda) &= K_N(\bx,\bX)\big(\bK_N+\lambda\id_N\big)^{-1}\by\\
& =  K_N(\bx,\,\cdot\,)^{\sT}\big(\bK_N+\lambda\id_N\big)^{-1}\bff_{*} + K_N(\bx,\bX)\big(\bK_N+\lambda\id_N
\big)^{-1}\beps\, ,
\end{align}
where we used the notations  $K_N(\bx,\,\cdot\,):= (K_N(\bx,\bx_i):\; i\le n)\in\reals^n$
and $\bff_{*} := (f_*(\bx_i):\; i\le n)\in\reals^n$.
Consider, to be concrete, the bias contribution to the test error. We get
\begin{align*}
\Bias(\lambda) &= \E_{\bx}\big\{\big[f_*(\bx)- K_N(\bx,\,\cdot\,)^\sT \big(\bK_N+\lambda\id_N\big)^{-1}\bff_{*}\big]^2\big\}\\
&=\E_{\bx}\big\{f_*(\bx)^2\big\}-2
\E_{\bx}\big\{f_*(\bx)K_N(\bx,\,\cdot\,)\big\}^{\sT}\big(\bK_N+\lambda\id_N\big)^{-1}\bff_{*}\\
&\phantom{AAA}+\bff_*\big(\bK_N+\lambda\id_N\big)^{-1}\bK^{(2)}_N\big(\bK_N+\lambda\id_N\big)^{-1}\bff_{*}\, ,
\end{align*}
where $\bK^{(2)}_N\in\reals^{n\times n}$ is the matrix with entries 
$(\bK^{(2)}_N)_{ij} = \E_{\bx}\{K_N(\bx_i,\bx)K_N(\bx,\bx_j)\}$.

Clearly the bound on the kernel matrix afforded by Theorem \ref{thm:invert2}
is not sufficient to control this expression. Indeed, proving the result below
is significantly more challenging.
\begin{theorem}\label{thm:gen}
Let  $B,c_0>0$, $\ell\in\naturals$  be fixed. 
 Then, there exist constants $C_0, C,C'>0$ such that the following holds.  

If $\sigma$ satisfies Assumption  \ref{ass:Sigma},  and further 
\begin{align}
&  c_0d \le n \le \frac{d^2}{(\log d)^{C_0} }, & \text{if}~\ell = 1, \label{eq:AssLge1}\\
& d^{\ell} (\log d)^{C_0} \le n \le \frac{d^{\ell + 1}}{(\log d)^{C_0} }, & \text{if}~\ell > 1.
\label{eq:AssLge2}
\end{align}

If $Nd/(\log (Nd))^C\ge n$, then for any $\lambda\ge 0$,
\begin{align*}
R_{\NT}(f_*;\lambda) &= R_{\KRR}(f_*;\lambda) + 
O_{d,\prob}\Big(\tau_+^2\sqrt{\frac{n(\log (Nd))^{C'}}{Nd}}\Big)
\end{align*}
where $\tau_+^2 := \|f_*\|^2_{L^2}+\tau^2$.
\end{theorem}

\begin{figure}
 \includegraphics[width=0.5\columnwidth,angle=0]{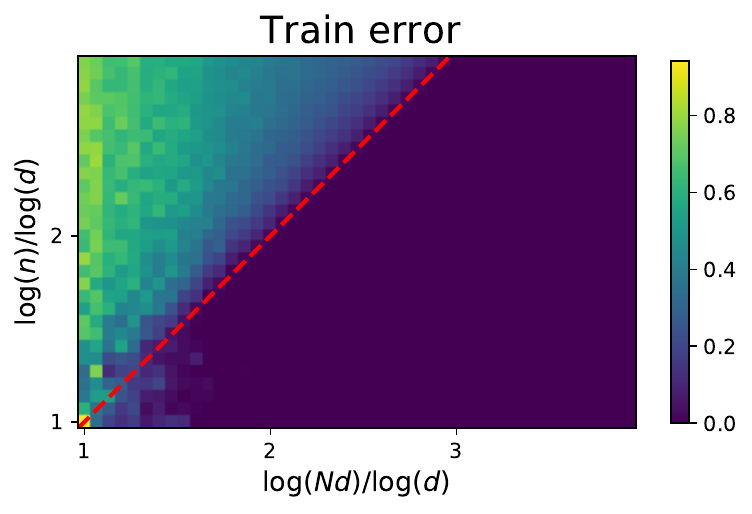}
   \includegraphics[width=0.5\columnwidth,angle=0]{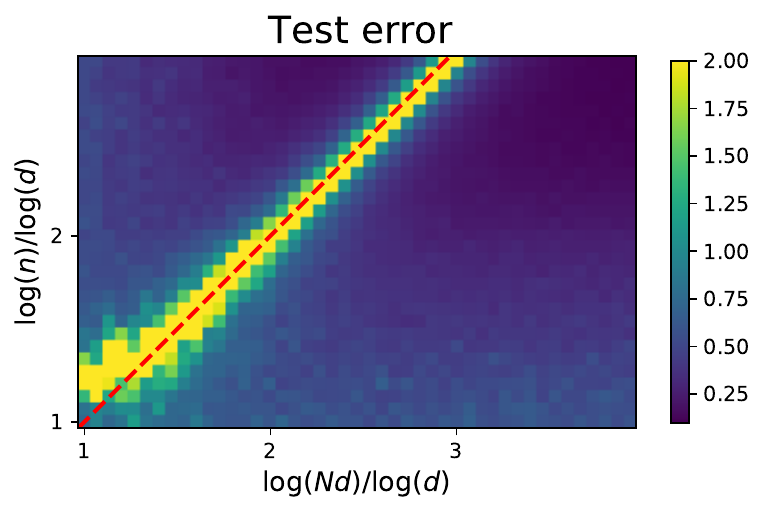}
   \caption{Neural Tangent min-norm regression ($\lambda=0+$) in $d=20$ dimensions
   (see text for details).
   Left: train error as a function of the sample size $n$ and number of parameters
   $Nd$. Right: test error. Results are averaged over $10$ realizations.
   (From \cite{montanari2022interpolation}.)}\label{fig:HeatMaps_NTK}
 \end{figure}

\begin{figure}
\begin{center}
   \includegraphics[width=0.57\columnwidth,angle=0]{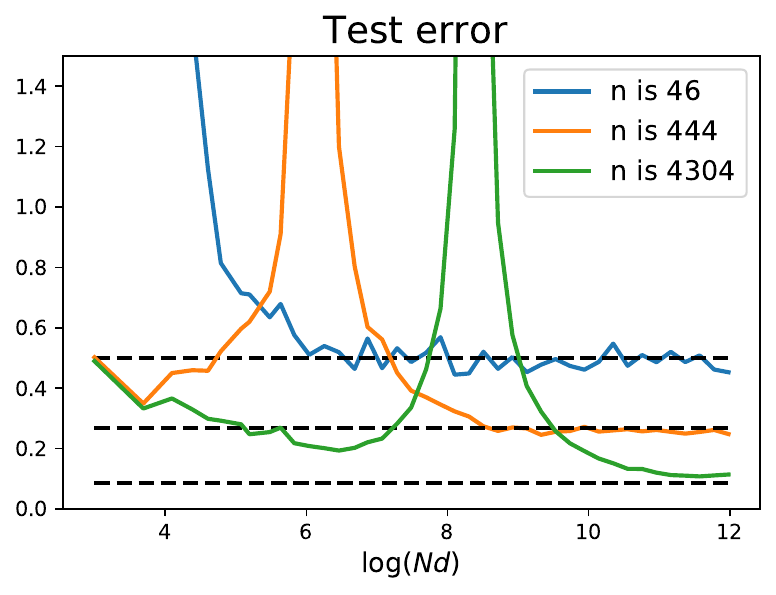}
\end{center}
   \caption{Test error Neural Tangent min-norm regression ($\lambda=0+$) in $d=20$ dimensions
   (see text for details).
   Color curves refer to $n=46, 444$ and $4304$, and dashed lines report the corresponding
   KRR limit. (From \cite{montanari2022interpolation}.)}\label{fig:Double_NTK}
 \end{figure}

As anticipated, this theorem establishes that the infinite width kernel is a good approximation
to the finite width kernel under the overparametrization condition
\eqref{eq:OverparametrizationNT} which is only a polylogarithmic factor above
the optimal condition $Nd\gg n$.

Roughly speaking, the condition of Eqs.~\eqref{eq:AssLge1} and \eqref{eq:AssLge2} 
rules out cases in which $n$ is comparable to $d^{\ell}$ for an integer $\ell$. As we saw in 
Chapter \ref{ch:random_matrix}, when  $n\asymp d^{\ell}$, the structure of the 
infinite width kernel matrix $\bK$ is more complex, and the KRR with respect to inner 
product kernels `partially' fits the degree $\ell$ component of the target function.
We expect this condition to be a proof artifact.

Figures \ref{fig:HeatMaps_NTK} and \ref{fig:Double_NTK} 
(from \cite{montanari2022interpolation}) illustrate the content of 
Theorem \ref{thm:gen} via numerical simulations. In these simulations we fix
$d=20$,  $\tau=0.5$ and consider the target function
\begin{align}
f_*(\bx) = \sqrt{\frac{4}{10}} \,\He_1(\<\bbeta_*,\bx\>)+
\sqrt{\frac{2}{10}} \, \He_2(\<\bbeta_*,\bx\>)+\sqrt{\frac{1}{120}} \,\He_4(\<\bbeta_*,\bx\>)\, .
\end{align}
We can clearly observe the main phenomena captured by Theorem \ref{thm:gen}.
The train error vanishes for $Nd\ge n$, while the test error converges to a limit
for large $Nd/n$. This limit is accurately reproduced by (infinite-width) KRR.

\begin{remark}
As we saw in Chapter \ref{ch:random_matrix}, under the conditions
of Eqs.~\eqref{eq:AssLge1}, \eqref{eq:AssLge2}, KRR with respect to an inner product kernel approximates
the projection of the target function onto degree $\ell$ polynomials.
Putting this result together with Theorem \ref{thm:gen} (under the conditions in the theorem's
statement) yields 
\begin{align}
R_{\NT}(f_*;\lambda) &=R_{\PRR}(f_*;\lambda + v_{\ell}(\sigma)) + 
O_{d,\prob}\Big(\tau_+^2\sqrt{\frac{n(\log (Nd))^{C'}}{Nd}} \, + \tau_+^2 \sqrt{\frac{n (\log n)^C}{d^{\ell+1}}} \Big)\, .
\label{eq:PRR}
\end{align}
Here $R_{\PRR}(f_*;\lambda')$ is the risk of polynomial ridge regression (PRR), i.e.
ridge regression with respect to polynomials of maximum degree $\ell$, at regularization
 $\lambda'$. Note that Eq.~\eqref{eq:PRR} is somewhat more quantitative than the statement
  Chapter \ref{ch:random_matrix}. Namely, the difference between KRR and PRR 
  is upper bounded by $(n/d^{\ell+1})^{1/2}$, up to a polylogarithmic factor.
 
 We refer to \cite{montanari2022interpolation} for a more complete discussion,
 but  point out that Eq.~\eqref{eq:PRR} displays one more example of self-induced
 regularization.  Namely, the ridge regularization parameter in PRR is
 equal $\lambda + v_{\ell}(\sigma)$. The original parameter $\lambda$ gets inflated
 by an additive term $v_{\ell}(\sigma)$ that comes from the high frequency part of the activations.
\end{remark}
%
%
\section{Proof sketch: Kernel concentration}

The proof of Theorem \ref{thm:invert2} is simple and instructive, and 
hence we present a brief sketch.  We will work under the simplifying assumption that 
$\sup_t|\sigma'(t)|\le M<\infty$, and $n\ge d$.

The proof uses the following classical matrix concentration inequality
\cite{vershynin2018high}[Thm.~5.4.1].
(Below, for a random variable $U$, $\esssup(U):=\inf\{t\in\reals: \prob(U\le t)=1\}$.)
\begin{theorem}[Matrix Bernstein inequality]
Let $\{\bZ_k\}_{k\le N}$ be a sequence of symmetric and independent random matrices
$\bZ_k\in\reals^{n\times n}$.
Define
\begin{align}
v:= \Big\|\sum_{k=1}^N\E[\bZ_k^2] \Big\|_{\op}\, ,\;\;\;\;\;
L:= \max_{k\le N}{\rm esssup} \|\bZ_k\|_{\op}\, .
\end{align}
Then, for any $t\ge 0$. 
\begin{align}\label{ineq:bern}
\P\Big( \Big\|\frac{1}{N}\sum_{k=1}^N  \bZ_k   \Big\|_{\op} \ge t \Big) \le 
2n \cdot \exp \left( - \frac{N^2t^2/2}{v + NLt/3} \right),
\end{align}
\end{theorem}

In other words, matrix Bernstein states that, with probability at least
$1-n^{-\beta}$
\begin{align}
\Big\|\frac{1}{N}\sum_{k=1}^N  \bZ_k   \Big\|_{\op}\le
\frac{C(\beta)}{N}\max\Big(\sqrt{v\log n}, L\log n\Big)\, .\label{eq:BasicBernstein}
\end{align}
We define
\begin{align}
\bD_k := \diag\big(\sigma'(\<\bw_k,\bx_1\>),\dots,\sigma'(\<\bw_k,\bx_n\>)\big)\,.
\end{align}
Letting $\bS_N:= \bK^{-1/2}\bK_N\bK^{-1/2}-\id$ denote the matrix
that we want to bound, we have
\begin{align}
\bS_N &= \frac{1}{N}\sum_{k=1}^N \bZ_k\, ,\\
\bZ_k &:= \frac{1}{d}\bK^{-1/2}\bD_k\bX\bX^{\sT}\bD_k\bK^{-1/2}-\id\, .
\end{align}

We begin by bounding the coefficient $L$ in Bernstein inequality. By triangle inequality
\begin{align}
\|\bZ_k\|_{\op}&\le \frac{1}{d}\lambda_{\min}(\bK)^{-1} M^2\|\bX\bX^{\sT}\|_{\op}+1\, .
\end{align}
Therefore, on the event $\cA_{\gamma}$
\begin{align*}
L & \le \frac{M^2}{\gamma d} 4(\sqrt{n}+\sqrt{d})^2+1\\
& \le C \frac{n}{d}\, ,
\end{align*}
where the constant $C$ depends on $M$, $\gamma$.

Next consider the coefficient $v$.
We have of course $\E[\bZ_k^2]\succeq \bfzero$. On the other hand,
\begin{align*}
\E[\bZ_k^2] & 
= \frac{1}{d^2}\E\big\{\bK^{-1/2}\bD_k\bX\bX^{\sT}\bD_k\bK^{-1}\bD_k\bX\bX^{\sT}\bD_k\bK^{-1/2}\big\}-\id\\
&\preceq \frac{M^2}{\gamma d^2}
\E\big\{\bK^{-1/2}\bD_k\bX\bX^{\sT}\bX\bX^{\sT}\bD_k\bK^{-1/2}\big\}\\
&\preceq \frac{M^2}{\gamma d^2}4(\sqrt{n}+\sqrt{d})^2\E\big\{\bK^{-1/2}\bD_k\bX\bX^{\sT}\bD_k\bK^{-1/2}\big\}\\
& = \frac{M^2}{\gamma d}4(\sqrt{n}+\sqrt{d})^2\id \, .
\end{align*}
Summing over $k$, we get $v\le CNn/d$.

Summarizing, there exists a constant $C$ (depending only on $M$ and $\gamma$)
such that 
\begin{align*}
L\le \frac{Cn}{d}\, ,\;\;\; v\le \frac{CNn}{d}\, .
\end{align*}
Applying Bernstein inequality \eqref{eq:BasicBernstein}, we get
\begin{align*}
\Big\|\bK^{-1/2}\bK_N\bK^{-1/2}-\id  \Big\|_{\op}&\le
\frac{C}{N}\max\left(\sqrt{\frac{Nn}{d}\log n}, \frac{n}{d}\log n \right)\\
&\le C\left(\sqrt{\frac{n}{Nd}\log n}+ \frac{n}{Nd}\log n \right)\, .
\end{align*}
This proves Theorem \ref{thm:invert2} in the case of bounded $\sigma'$.
The general case requires to handle large values of $\sigma'(\<\bw_k,\bx_i\>)$
via a truncation argument. This results in the extra logarithmic factors, which are likely
to be a proof artifact.

\chapter{Why stop being lazy (and how)}
\label{ch:drawbacks}
These lecture notes were focused so far on learning within a training regime
in which the model $f(\bx;\btheta)$ is well approximated by its first order
Taylor expansion  (in the model parameters) around its initialization. This style of analysis is
 useful in elucidating
some puzzling properties of modern machine learning:
\begin{enumerate}
\item Overparametrization helps optimization.
\item Gradient-based algorithms select a specific model among all the ones that minimize
the empirical risk.
\item The specific models can overfit the training data (vanishing training error)
and yet generalize to unseen data.
\end{enumerate}
The linear regime provides a simple and analytically tractable setting in which 
these phenomena can be understood in detail.

Finally, there is experimental evidence that, in some settings, the linear theory captures
the behavior of real SGD-trained neural networks \cite{lee2019wide,geiger2020disentangling} (see also Appendix \ref{app:Bibliography}).

On the other hand, it is important to emphasize that linearized neural networks
are significantly less powerful than fully trained neural networks. 
In other words, while neural networks can be trained in a regime in which they are 
well approximated by their first order Taylor expansion, this is the byproduct of
specific choices on the parametrization. Other choices are  possible, leading to
different and potentially better models.

Following \cite{chizat2019lazy} we will use the somewhat irreverent term `lazy training'  
to refer to training schemes that produce networks in the linear regime.
In this chapter we will discuss limitations of this approach, and how other 
training schemes overcome them. In particular, we will clarify that lazy training is not
a necessary consequence of the infinite width limit. Other non-lazy infinite-width limits exist, and can outperform lazy ones.

\section{Lazy training fails on ridge functions}
\label{sec:RidgeLazy}

One of the simplest problems on which lazy training is suboptimal is provided 
by `ridge functions', i.e. functions that depend on a one-dimensional projection of the 
covariate vector.

As in the previous chapters, assume to be given data $\{(\bx_i,y_i)\}_{i\le n}$,
where $\bx_i\sim\Unif(\S^{d-1}(\sqrt{d}))$, $y_i = f_*(\bx_i) +\eps_i$ and
$\eps_i$ is noise with $\E \{ \eps_i \}=0$,  $\E \{ \eps_i^2 \}=\tau^2$. A ridge function
is a target function of the form: 
\begin{align}
f_*(\bx)=\varphi(\<\bw_*,\bx\>)\, ,\label{eq:RidgeTarget}
\end{align}
where\footnote{In fact, the term ridge function  refers in applied mathematics and statistics 
to the broader class of targets of the form $f_*(\bx)=\psi(\bU_*^{\sT}\bx)$
where $\psi:\reals^k\to\reals$ and $\bU_*^{\sT}\bU_*=\id_{k}$ \cite{logan1975optimal}.
Unfortunately, this terminology can be confusing because we are applying ridge
regression to learn ridge functions, and the two uses of ``ridge" are not related!}
 $\varphi:\reals\to\reals$ and $\|\bw_*\|_2=1$.

We saw in the previous chapters that the learning under isotropic covariates in
the linear regime is controlled by the decomposition of $f_*$ into polynomials.
Let us compute the mass of $f_*$ in the subspace $\cV_{d,k}$ of polynomials of degree $k$ that 
are orthogonal to polynomials of maximum degree $k-1$. 
Using the theory of Section \ref{sec:diag_inner_prod_kernel}, we obtain
\begin{align*}
\lim_{d\to\infty}\|\proj_{k} f_*\|^2_{L^2} &= 
\lim_{d\to\infty} \, B(d,k)\cdot \E_{\bx}\big\{\varphi(\<\bw_*,\bx\>) Q^{(d)}_k(\<\bw_*,\bx\>)\big\}^2\\
&= \frac{1}{k!}\E\big\{\varphi(G) \He_k(G)\big\}^2\, .
\end{align*}
In particular\footnote{Inverting follows from dominated convergence if, for instance 
$|\varphi(t)|\le C\exp(C| t| )$ for a constant $C >0$ and all $t \in \reals$.}
\begin{align*}
\lim_{d\to\infty}\|\proj_{>\ell} f_*\|^2_{L^2} = \sum_{k=\ell+1}^{\infty}
\frac{1}{k!}\E\big\{\varphi(G) \He_k(G)\big\}^2 =:b_{\ell}\, .
\end{align*}
Unless $\varphi$ is a polynomial, we have $b_{\ell}>0$ strictly for all $\ell$.
This simple remark has an immediate consequence. We saw in previous chapters that
if the sample size is $n\ll d^{\ell+1}$, then linearized neural network only learn the 
best approximation of the target by degree-$\ell$ polynomials. Hence, their excess risk 
is bounded away from zero. This consequence is stated more precisely  below.
\begin{corollary}\label{coro:RidgeKRR}
Fix $\delta>0$.
Let $R_{{\sf X}}(f_*)$, ${\sf X}\in \{\KRR, \RF, \NT\}$ be (respectively) the excess test error of 
 kernel ridge regression (with inner product kernel), random feature ridge regression, 
 or neural tangent regression. For ${\sf X} = \RF$, assume $N\ge n^{1+\delta}$, and for
  ${\sf X} = \NT$, assume $Nd\ge n^{1+\delta}$. Under the data model above
  (in particular, $y_i=\varphi(\<\bw_*,\bx_i\>)+\eps_i$), 
  if $n\le d^{\ell+1-\delta}$ then  the following lower
  bound holds in probability:
\begin{align}
\liminf_{n,d\to\infty} R_{{\sf X}}(f_*) \ge b_{\ell}\, .
\end{align}
Further unless $\varphi$ is a polynomial, we have $b_{\ell}>0$ strictly for all $\ell$.
\end{corollary}

This is somewhat disappointing. After all, the function \eqref{eq:RidgeTarget}
looks extremely simple, and only has $d$ free parameters (plus, eventually $O(1)$ parameters
to approximate the one-dimensional function $\varphi$). 
One extreme case is given by $\varphi=\sigma$. Note that Corollary \ref{coro:RidgeKRR} does not 
change in this case: linearized neural networks cannot learn a target function that coincides with
 a single neuron from any polynomial number of samples.

It is natural to
wonder whether learning ridge functions might be hard for some hidden reason.
It turns out that ridge functions can often be learnt efficiently, even if we limit ourselves to 
gradient-based methods.
For instance,  consider  using a one-neuron network $f(\bx;\bw):=\sigma(\<\bw,\bx\>)$
in the case $\varphi=\sigma$. We perform gradient descent with respect to the empirical
risk:
\begin{align}
\hR_n(\bw) = \sum_{i=1}^n\big(y_i-\sigma(\<\bw,\bx_i\>)\big)^2\, .
\end{align}
The corresponding excess test error is $R(\bw) = \E\{(\sigma(\<\bw_*,\bx\>)-\sigma(\<\bw,\bx\>))^2\}$.
The following result from \cite{mei2018landscape} implies that  gradient descent 
succeeds in learning this target.
\begin{proposition}\label{propo:OneNeuron}
Under the above data distribution (namely, $\bx_i\sim\Unif(\S^{d-1}(\sqrt{d}))$ and
$y_i = \sigma(\<\bw_*,\bx_i\>)+\eps_i$) assume $\sigma$ is bounded and three times differentiable 
with bounded derivatives. Further assume $\sigma'(t)>0$ for all $t$. 
Then there exists a constant $C = C_{\sigma,\tau}$ such that, if $n\ge C\, d \log d$, 
the following happens:
\begin{enumerate}
\item  The empirical risk $\hR_n(\bw)$ has a unique (local, hence global) minimizer 
$\hbw_n\in\reals^d$.
\item Gradient descent converges globally to $\hbw_n$. 
\item The excess test error at this minimizer is bounded as
\begin{align}
R(\hbw_n) \le C\sqrt{\frac{d\log n}{n}}\, .
\end{align}
\end{enumerate}
\end{proposition}

Considering ---to be definite--- the setting of this proposition,
and learning using two layer networks.
We find ourselves in the following peculiar situation:
\begin{enumerate}
\item A one-neuron network can learn the target from $n=O(d\log d)$ samples,
which roughly matches the information theoretic lower bound.
\item As the number of neurons gets large, we learn from 
\cite{jacot2018neural} that a two-layer neural network is well approximated by 
the corresponding  NT model.
\item The NT model cannot learn the same target from any polynomial (in $d$)
number of samples.
\end{enumerate}

In other words, the learning ability of two-layer networks seems to deteriorate
as the number of neurons increases. Needless to say, this is a very counterintuitive behavior.

What is `going wrong'?

As we will see in the next section, point 2 above is what is going wrong.
Wide neural networks are well approximated by their linear counterpart only under
a specific scaling of the network weights. Other behavior (and other infinite-width limits)
are obtained if different scalings are chosen. Under these scalings, the obstructions and
counterintuitive behaviors above disappear.

\section{Non-lazy infinite-width networks (a.k.a. mean field)}

Let us restart from scratch, and write once more the parametric form of a two-layer 
neural network, which we first wrote down in Eq.~\eqref{eq:TwoLayer}:
\begin{align}
\hf(\bx;\bTheta) = \alpha_N\sum_{i=1}^Na_i\, \sigma(\<\bw_i,\bx\>)\, ,\;\;\;\;\;
\btheta_i = (a_i,\bw_i)\in\reals^{d+1}\, .\label{eq:TwoLayer_again}
\end{align}
Here we denoted collectively by $\bTheta: = \big((a_i,\bw_i):\; i\le N\big)$
the network parameters. 
We mentioned briefly in Chapter \ref{ch:intro}
that in certain regimes, such a network is well approximated by its first order Taylor expansion around
the initialization. Roughly speaking, this is the case when $N$ is large, but only if we 
scale $\alpha_N$ in a suitable way, i.e. $\alpha_N\asymp N^{-1/2}$. 

We want to revisit the infinite-width limit, under a different scaling of the model, namely
\begin{align}
\hf(\bx;\bTheta) = \frac{1}{N}\sum_{i=1}^Na_i\, \sigma(\<\bw_i,\bx\>)\, .
\end{align}
This is sometimes referred to as the `mean field' scaling, after \cite{mei2018mean}, which 
we will follow here. 

It is useful to denote by $\btheta_i=(a_i,\bw_i)\in\reals^{d+1}$ the vector of parameters
for neuron $i$, and define $\sigma_*(\bx;(a,\bw)) := a\sigma(\<\bw,\bx\>)$.
Consider SGD training with respect to square loss. We will fix a stepsize $\eta$
and index iterations by $t\in\{0,\eta,2\eta,\dots\}$. The SGD iteration reads:
\begin{align}\label{eq:SGD}
\btheta_i^{t+\eta} = \btheta^t_i +\eta(y_{I(t)} - \hf(\bx_{I(t)};\bTheta^t)) 
\nabla_{\btheta } \sigma_* ( \bx_{I(t)} ; \btheta^t_i )\, ,
\end{align}
where $I(t)$ is the index of the sample used at step $t$ of SGD. 
We will have in mind two settings: $(i)$~Online SGD, whereby at each step a fresh 
sample from the population is used (in this case we are optimizing the population error); 
$(ii)$~Standard SGD with batch size $1$, whereby $I(t)\sim \Unif([n])$. 

When the stepsize $\eta$ becomes small, SGD effectively averages over a large number 
of samples in a fixed time interval $[t, t+\Delta]$, and it is
natural to expect that the above dynamics is well approximated by the following
gradient flow
\begin{align}\label{eq:GF}
\dot\btheta_i^{t} = \E_{y,\bx}\big\{(y- \hf(\bx;\bTheta^t)) 
\nabla_{\btheta } \sigma_* ( \bx ; \btheta^t_i )\}\, .
\end{align}
Here $\E_{y,\bx}$ is the expectation over the population distribution for online SGD,
or over the sample for standard SGD. 
We can conveniently rewrite the above flow by introducing the functions
\begin{align}
V(\btheta) &= - \E_{y, \bx}[y \sigma_*(\bx;\btheta)]\\
& =- \E_{y, \bx}[y a\sigma(\<\bw,\bx\>)], \nonumber\\
U(\btheta_1, \btheta_2) &= \E_{\bx}[\sigma_*(\bx;\btheta_1) \, \sigma_*(\bx;\btheta_2)]\\
&= \E_{\bx}[a_1\sigma(\<\bw_1,\bx\>) \, a_2\sigma(\<\bw_2,\bx\>)]\, . \nonumber
\end{align}
We can then rewrite Eq.~\eqref{eq:GF} as 
\begin{align}\label{eq:GF-2}
\dot\btheta_i^{t} = -\nabla_{\btheta}V(\btheta^t_i)-\frac{1}{N}
\sum_{j=1}^N\nabla U(\btheta^t_i,\btheta^t_j)\, .
\end{align}
This expression makes it clear that the right-hand side depends only on $\btheta_i^t$
and on the empirical distribution of the neurons
\begin{align}
\hrho_{N,t} := \frac{1}{N}\sum_{i=1}^N\delta_{\btheta^t_i} \, . 
\end{align}
Indeed, we can further rewrite Eq.~\eqref{eq:GF-2}
\begin{align}\label{eq:GF-3}
\dot\btheta_i^{t} = -\nabla_{\btheta}V(\btheta^t_i)-\int
\nabla V(\btheta^t_i,\btheta') \, \hrho_{N,t}(\de\btheta')\, .
\end{align}

We can put on our physicist's hat and interpret Eq.~\eqref{eq:GF-3} as describing the evolution
of $N$ particles, where the velocity of each particle is a function of the density
at time $t$. Hence the density must satisfy a continuity partial differential equation (PDE):
\begin{align}
    \partial_t \hrho_{N,t} & =
 \nabla_{\btheta} \cdot \big(\hrho_{N,t}(\btheta) \nabla_\btheta \Psi(\btheta; \hrho_{N,t})\big) \, ,
 \label{eq:PDE-N}
\end{align}
where
\begin{align}
  \Psi(\btheta; \rho) & = V(\btheta)  + \int U(\btheta, \btheta') \, \rho(\de \btheta') \, .
\end{align}
Let us point out that Eq.~\eqref{eq:PDE-N} is a completely equivalent description
of the gradient flow \eqref{eq:GF}, and holds for any finite $N$, provided one is careful
about defining gradients of delta functions\footnote{More precisely,
Eq.~\eqref{eq:PDE-N} has to hold in weak sense, i.e. integrated against a suitable
test function.} and similar technicalities.

The formulation of Eq.~\eqref{eq:PDE-N} has two interesting properties:
\begin{itemize}
\item The evolution takes place in the same space for different values of $N$,
this is the space of probability measures $\hrho_t$ in $d+1$ dimensions.
While we initially introduced Eq.~\eqref{eq:PDE-N} for probability measures that are sums
of point masses, the same PDE makes sense more generally.
\item Equation~\eqref{eq:PDE-N} `factors out' the invariance of 
Eq.~\eqref{eq:GF} under permutations of the neurons $\{1,\dots,N\}$.
\end{itemize}

Because of these reasons, it is very easy (at least formally)
to take the $N\to\infty$ limit in Eq.~\eqref{eq:PDE-N}.  Assume that 
the initialization satisfies $\hrho_{N,0}\Rightarrow\rho_{\sinit}$
as $N\to\infty$, where $\Rightarrow$ denotes the weak convergence of measures. For instance this is the case if
 \begin{align}
 \{(a^0_i,\bw^0_i)\}_{i\le N}\sim_{iid} \rho_{\sinit}\, .
 \end{align}
 Then (under suitable technical conditions) it can be proven that
 $\hrho_{N,t}\Rightarrow \rho_t$ for any $t$, where $\rho_t$ satisfies
 \begin{align}
    \partial_t \rho_t & =
 \nabla_{\btheta} \cdot \big(\rho_t(\btheta) \nabla_\btheta
  \Psi(\btheta; \rho_t)\big) \, ,\label{eq:PDE-lim} \\
  \rho_0& = \rho_{\sinit}\, .\nonumber
\end{align}
We also note that the potential $\Psi(\btheta; \rho)$ can be interpreted 
as the first order derivative of the population error with respect to
a change of density of neuron at $\btheta$. Using physics notations:
\begin{align}
\Psi(\btheta;\rho) = \frac{\delta R(\rho)}{\delta \rho(\btheta)}\, ,
\end{align} 
where we defined
\begin{align}
R(\rho) &:= \frac{1}{2}\E\Big\{\Big(f_*(\bx)-\int \sigma_*(\bx;\btheta)\, \rho(\de\btheta)\Big)^2\Big\}\label{eq:PopulationRiskRho}\\
& = \frac{1}{2}\E(f_*(\bx)^2) +  \int V(\btheta) \rho(\de \btheta) + \frac{1}{2}\int U(\btheta, \btheta') \rho(\de \btheta) \rho(\de \btheta')\, .
\nonumber
\end{align}

In conclusion, by scaling the function $f(\bx;\bTheta)$, we obtained a
different limit description of infinitely wide networks. Rather than a linearly parametrized model,
which is fitted via min-norm regression, we obtained a fully nonlinear model.

\begin{remark}
Nonlinear continuity equations of the form \eqref{eq:PDE-lim} have been studied for a long
 time in mathematics, starting from the seminal work of Kac \cite{kac1956foundations}
 and McKean \cite{mckean1966class}. Within this literature, this is known as the McKean-Vlasov equation.
We refer to Appendix \ref{app:Bibliography} for further pointers to this line of work.
\end{remark}

\section{Learning ridge functions in mean field: basics}

Summarizing, the two-layer neural network \eqref{eq:TwoLayer_again} behaves in different 
ways depending on how we scale the coefficient $\alpha_N$. In particular:
\begin{itemize}
\item If $\alpha_N \asymp 1/\sqrt{N}$, for large enough $N$, the neural network
is well approximated by the neural tangent model. In particular, for $n\le d^{1+\delta}$,
the network only learns the projection of the target function onto linear functions.
\item If  $\alpha_N \asymp 1/\sqrt{N}$, the SGD trained network converges as $N\to\infty$
to the network $f(\bx;\rho_t) := \int a\, \sigma(\<\bw,\bx\>)\, \rho_t(\de a,\de \bw)$,
where $\rho_t$ solves the PDE \eqref{eq:PDE-lim}.
\end{itemize}
What is the behavior of the PDE \eqref{eq:PDE-lim} under the data distribution introduced in 
Section \ref{sec:RidgeLazy}, namely $\bx_i\sim\normal(0,\id_d)$, $y_i = \varphi(\<\bw_*,\bx_i\>)+\eps_i$?
A fully rigorous treatment of this question is somewhat intricate and goes
beyond the scope of this overview. We will limit ourselves to some simple
considerations, and overview recent progress on this topic in Appendix \ref{app:Bibliography}.
We will assume that the conditions of Proposition \ref{propo:OneNeuron} hold.

Consider an uninformative initialization. In the present case, this means the first layer weights 
$\bw_i$ have spherically symmetric distribution. For the sake of concreteness, we can assume 
them to be Gaussian and take 
\begin{align}
\rho_{\sinit}(\de a,\de\bw)=\rP_A\otimes \normal(\bfzero,\gamma^2\id_d/d)\, ,\label{eq:RhoInit}
\end{align}
where the distribution $\rP_A$ of the second-layer weights can be arbitrary. 
(The scaling of the covariance above with $d$ is chosen so that $\|\bw_i\|=\Theta(1)$
for large $d$.)

Now, the data distribution is invariant under rotations that leave $\bw_*$ unchanged.
As a consequence, it is not hard to show that the evolution \eqref{eq:PDE-lim}
is equivariant under the same group. Namely, let $\bR\in\reals^{d\times d}$ be 
a rotation such that $\bR\bw_* = \bw_*$. For a probability distribution
$\rho(\de a,\de\bw)$ on $\reals\times\reals^d$, we let $\bR_{\#}\rho$ be the push forward
under this rotation\footnote{$\bR_{\#}\rho$ is the law of $(a,\bR\bw)$
when $(a,\bw)\sim \rho$.} (acting trivially on the first factor $\reals$).
Then if $\rho_0$ and $\rho_0' = \bR_{\#}\rho_0$ are two initializations for the PDE,
if $\rho_t$, $\rho_t'$ are the corresponding solutions to Eq.~\eqref{eq:PDE-lim},
then we necessarily have $\rho_t'=\bR_{\#}\rho_t$. 

Since the initialization \eqref{eq:RhoInit} is invariant under such rotations,
$\rho_t$ will be invariant for all $t$. In other words, under $\rho_t$, $(a,\bw)$
is uniformly random, given $(a,\<\bw_*,\bw\>,\|\proj_{\perp}\bw\|_2)$,
where $\proj_{\perp}$ is the projection orthogonal to $\bw_*$.
We can therefore write a PDE for the joint distribution of these three
quantities which we denote by $\orho_t(\de a,\de s,\de r)$. 
Writing $\bz = (a,s,r)$, $\orho_t$ is obtained by solving the PDE
\begin{align}
\partial_t \orho_t = \partial_a(\orho_t\partial_a\Psi(\bz;\orho_t))+
 \partial_s(\orho_t\partial_s\Psi(\bz;\orho_t))+
\frac{1}{r} \partial_r(r\orho_t\partial_r\Psi(\bz;\orho_t))\, .\label{eq:ReducedPDE}
\end{align}
The initialization \eqref{eq:RhoInit} translates into
\begin{align}
\orho_{\sinit}= \rP_A \otimes \normal( 0 ,\gamma^2/d)\otimes \sQ_{d-1,\gamma}\, ,
\end{align}
where $\sQ_{k,\gamma}$ is the law of 
the square root of a chi-squared with $k$ degrees of freedom, rescaled by $\gamma/\sqrt{k+1}$.

\begin{figure}
\begin{center}
 \includegraphics[width=0.6\columnwidth,angle=0]{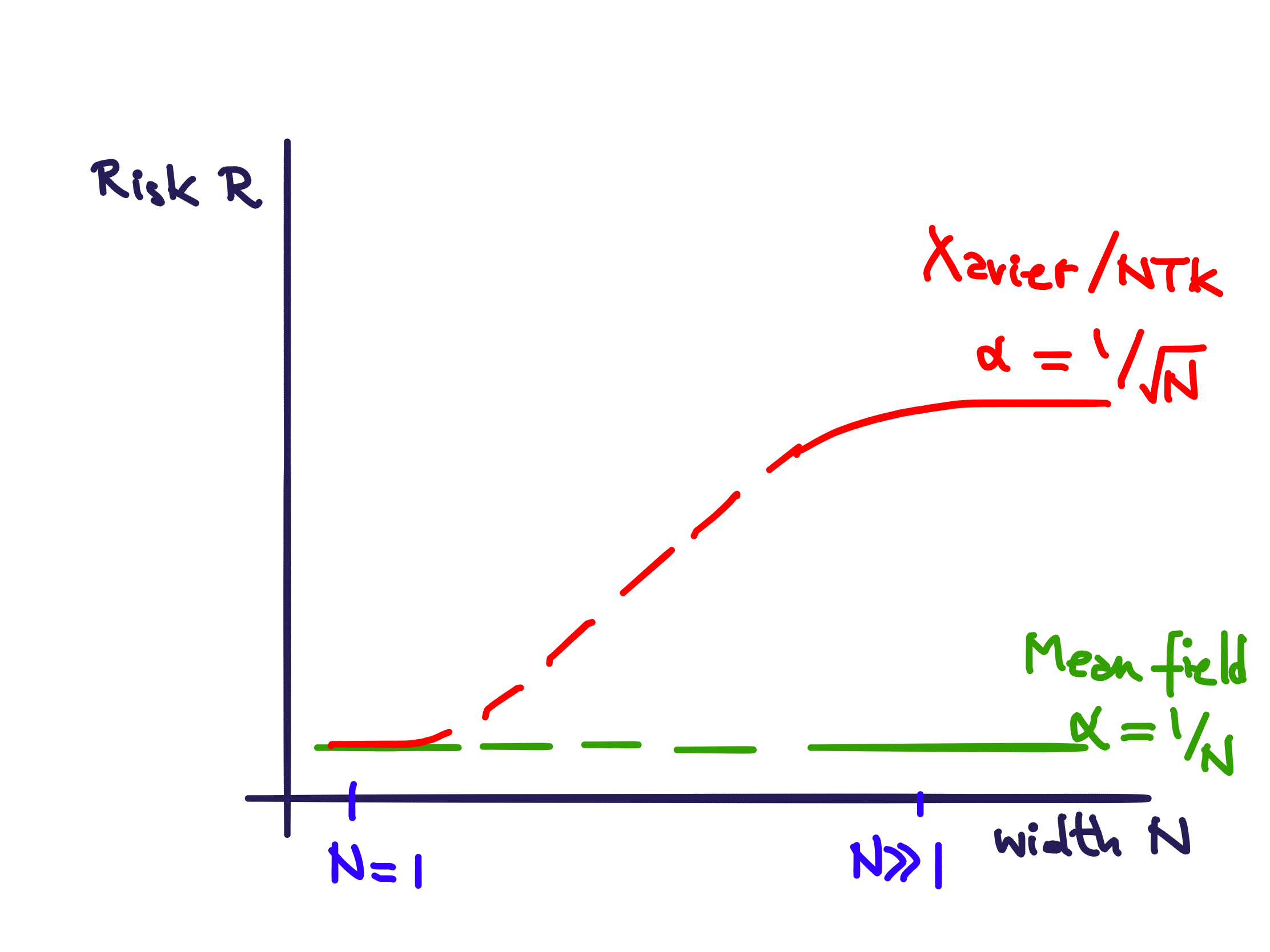}
 \end{center}
   \caption{Conjectured behavior of the prediction error as a function of width
   for different scalings of the network.}\label{fig:SketchRidgeFunction}
 \end{figure}
Let $R(\orho_t)$ be the prediction risk of Eq.~\eqref{eq:PopulationRiskRho},
written in terms of $\orho_t$ (using the fact that $\bw$ is uniformly random conditional on 
$a,s,r$). It follows from the general theory that $t\mapsto R(\orho_t)$ is non-decreasing.
If it can be shown that
\begin{align}
\lim_{t\to \infty} R(\orho_t)\le \eps_0\, , \label{eq:GeneralConsistencyPDE}
\end{align}
then this directly implies that, for any $\eps>0$, SGD achieves test error upper bounded 
(with high probability) by $\eps_0+\eps$, from $O(d)$ samples. 
Indeed, the PDE \eqref{eq:PDE-lim} can be shown to approximate SGD
provided the stepsize satisfies $\eta\le 1/(C_0d)$ and therefore a time of order one
translates in a number of iterations (or samples) of order $d$ (assuming online SGD).

Indeed we expect Eq.~\eqref{eq:GeneralConsistencyPDE} to hold with $\eps_0=0$ in many cases of
interest,  e.g. under the assumptions on $\sigma=\varphi$ in Proposition \ref{propo:OneNeuron}.
A proof of this fact would be too long for these notes. 
Partial evidence is provided by the following facts:
\begin{enumerate}
\item The reduced PDE \eqref{eq:ReducedPDE} is 3-dimensional and can be solved numerically,
confirming Eq.~\eqref{eq:GeneralConsistencyPDE}. 
\item For $\rP_A= \delta_1$ and $\gamma$ small, if we change SGD and do not update the second layer 
weights, the resulting learning dynamics is similar to the one of a single neuron, covered
by Proposition \ref{propo:OneNeuron}, and therefore an approximation argument can be employed 
to bound the risk.
\end{enumerate}

We conclude that, under the setting of Proposition \ref{propo:OneNeuron},
very wide neural networks with mean-field scaling learn ridge functions from $O(d)$
samples. 

While we do not have a precise characterization of the risk for
intermediate values of $N$, we sketch in Figure \ref{fig:SketchRidgeFunction} the (partly conjectural) behavior of the prediction errors for different scalings of the network, when learning a ridge function with $\varphi = \sigma$.

\appendix

\chapter{Summary of notations}
\label{app:Notations}
For a positive integer, we denote by $[n]$ the set $\{1 ,2 , \ldots , n \}$. For vectors $\bu,\bv \in \reals^d$, we denote $\< \bu, \bv \> = u_1 v_1 + \ldots + u_d v_d$ their scalar product, and $\| \bu \|_2 = \< \bu , \bu\>^{1/2}$ the $\ell_2$ norm. We denote by $\S^{d-1} (r) = \{ \bu \in \reals^d : \| \bu \|_2 = r \}$ the sphere of radius $r$ in $d$ dimensions (we will simply write for the unit ball $\S^{d-1} := \S^{d-1} (1)$).	

Given a matrix $\bA \in \reals^{n \times m}$, we denote $\| \bA \|_{\op} = \max_{\| \bu \|_2 = 1} \| \bA \bu \|_2$ its operator norm and by $\| \bA \|_{F} = \big( \sum_{i,j} A_{ij}^2 \big)^{1/2}$ its Frobenius norm. If $\bA \in \reals^{n \times n}$ is a square matrix, the trace of $\bA$ is denoted by $\Tr (\bA) = \sum_{i \in [n]} A_{ii}$. Further, given a positive semidefinite matrix $\bB \in \reals^{d \times d}$ and for vectors $\bu \in \reals^d$, we denote $\| \bu \|_{\bB} = \| \bB^{1/2} \bu \|_2 = \<\bu, \bB \bu \>^{1/2}$ the weighted $\ell_2$ norm.

Given a probability space $(\cX , \nu)$, we denote $L^2 ( \cX) := L^2 (\cX, \nu )$ the space of square integrable functions on $(\cX , \nu)$, and $\< f , g \>_{L^2 (\cX)} = \E_{\bx \sim \nu} \{ f(\bx) g (\bx)\}$ and $\| f \|_{L^2 (\cX) } = \< f , f \>_{L^2 (\cX)}^{1/2}$ respectively the scalar product and norm on $L^2 (\cX)$ (we will sometimes write $\< \cdot , \cdot \>_{L^2}$ when $(\cX , \nu)$ is clear from context).  The space $L^2(\reals , \gamma)$ plays an important role throughout these notes, where $\gamma (\de x) = e^{-x^2/2} \de x / \sqrt{2 \pi}$ is the standard Gaussian measure. We will often prefer the notation $G \sim \normal (0,1)$ to denote the standard normal distribution, and $L^2 ( \normal (0,1)) := L^2 (\reals , \gamma)$. The set of Hermite polynomials $\{ \He_k \}_{k \geq 0}$ forms an orthogonal basis of $L^2 ( \reals , \gamma)$, where $\He_k$ has degree $k$. We follow the classical normalization:
\[
\< \He_k , \He_j \>_{L^2 (\normal (0,1) )} = \E_{G \sim \normal (0,1)} \{ \He_k (G) \He_j (G) \} = k! \mathbbm{1}_{j=k} \, .
\]
In particular, for any function $g \in L^2 (\normal (0,1) )$, we have the decomposition
\[
g(x) = \sum_{k = 0}^\infty \frac{\mu_k (g)}{k!} \He_k (x)\, , \qquad \mu_k (g) := \E_{G \sim \normal (0,1)} \{ g(G) \He_k (G) \} \, , \qquad \|g \|_{L^2}^2 = \sum_{k = 0}^\infty \frac{(\mu_k (g))^2}{k!}\, , 
\]
where $\mu_k (g)$ is sometimes referred to as the $k$-th Hermite coefficient of $g$.

We use $O_d(\, \cdot \, )$  (resp. $o_d (\, \cdot \,)$) for the standard big-O (resp. little-o) relations, where the subscript $d$ emphasizes the asymptotic variable. Furthermore, we write $f = \Omega_d (g)$ if $g(d) = O_d (f(d) )$, and $f = \omega_d (g )$ if $g (d) = o_d (f (d))$. Finally, $f =\Theta_d (g)$ if we have both $f = O_d (g)$ and $f = \Omega_d (g)$.

We use $O_{d,\prob} (\, \cdot \,)$ (resp. $o_{d,\prob} (\, \cdot \,)$) the big-O (resp. little-o) in probability relations. Namely, for $h_1(d)$ and $h_2 (d)$ two sequences of random variables, $h_1 (d) = O_{d,\prob} ( h_2(d) )$ if for any $\eps > 0$, there exists $C_\eps > 0 $ and $d_\eps \in \Z_{>0}$, such that
\[
\begin{aligned}
\prob ( |h_1 (d) / h_2 (d) | > C_{\eps}  ) \le \eps, \qquad \forall d \ge d_{\eps},
\end{aligned}
\]
and respectively: $h_1 (d) = o_{d,\prob} ( h_2(d) )$, if $h_1 (d) / h_2 (d)$ converges to $0$ in probability.  Similarly, we will denote $h_1 (d) = \Omega_{d,\prob} (h_2 (d))$ if $h_2 (d) = O_{d,\prob} (h_1 (d))$, and $h_1 (d) = \omega_{d,\prob} (h_2 (d))$ if $h_2 (d) = o_{d,\prob} (h_1 (d))$. Finally, $h_1(d) =\Theta_{d,\prob} (h_2(d))$ if we have both $h_1(d) =O_{d,\prob} (h_2(d))$ and $h_1(d) =\Omega_{d,\prob} (h_2(d))$.

\chapter{Literature survey}
\label{app:Bibliography}
\paragraph*{Chapter 1 (The linear regime)}$\phantom{a}$

\indent The connection between neural networks trained by gradient descent and kernel methods 
was first\footnote{Several connections between neural networks and kernel methods have been 
discussed earlier in the literature, e.g. see \cite{neal1995bayesian,cho2009kernel,daniely2016toward,
lee2018deep,novak2018bayesian} and references therein. However, in these cases, the correspondence 
with kernel models only holds for random neural networks, i.e. at initialization of the SGD dynamics,
 while \cite{jacot2018neural} holds, under certain conditions,  for the entire training trajectory.} 
 elucidated in 
 \cite{jacot2018neural}. They showed that, under a specific scaling of weights at initialization 
 and learning rate, neural networks trained by gradient flow converge to a kernel regression
  solution in the infinite-width limit ($N \to \infty$). The specific kernel of this
   solution was termed the \textit{neural tangent kernel} (NTK). A striking implication 
   of this finding is that gradient flow converges to 
   zero training error ---a global minimizer--- despite the non-convexity of the 
   full optimization landscape. Following this intuition, \cite{li2018learning,du2018gradient}
    proved ---under the same scaling of parameters--- global convergence of gradient 
    descent for two-layer neural networks with ReLu activation and sufficiently
     large but finite number of neurons. Subsequent studies extended 
     the proof of global convergence to deep neural networks and more general architectures
      \cite{du2019gradient,allen2019convergence,allen2019convergence2,lee2019wide,zou2020gradient}.
However, these results require stringent overparametrization conditions on the network width.
        For example, \cite{du2018gradient} requires two-layer neural networks with a number of 
        neurons $N$ of at least $ \Tilde \Omega (n^6/ \lambda_0^4)$, where $n$ is the training
         sample size and $\lambda_0$ is the minimum eigenvalue of the empirical neural tangent
kernel matrix. With the goal of establishing global convergence for networks with
           realistic widths, a large body of work has steadily pushed down the overparametrization 
           condition on $N$  \cite{ji2019polylogarithmic,kawaguchi2019gradient,song2019quadratic,zou2019improved,oymak2019overparameterized,oymak2020toward,zou2020gradient}.
In particular, \cite{oymak2020toward} showed that under mild assumptions on the data distribution,
 quadratic overparametrization $Nd = \Omega (n^2)$ is sufficient for global convergence of 
 two-layer neural networks. This condition can be further improved to near optimal 
 overparametrization $Nd = \Omega (n \log (n) )$ by modifying the parameter scaling 
 and choosing a suitable initialization, as shown in \cite{bartlett2021deep}. In addition 
 to global convergence of the training error, a number of studies have considered
  the generalization properties of neural networks in this regime and proved that low 
  test error can be achieved under various data distribution assumptions 
  \cite{li2018learning,arora2019fine,ijcai2021-304,cao2019generalization,nitanda2019gradient,ji2019polylogarithmic}.

As shown in \cite{chizat2019lazy}, the connection to kernel methods is
 not restricted to neural networks, but extends to more general non-linear models
  trained by gradient descent. Indeed, it originates from a particular choice in the 
  scaling of the model parameters. With this specific scaling, a small change 
  in the weights produces a large change in the output of the non-linear model but only a small 
  change in its Hessian (see also discussion in \cite{liu2020linearity}). As a consequence,
   neural networks in this optimization regime have their weights barely moving, while 
   they converge to $0$ training loss. This led \cite{chizat2019lazy} to dub this regime the 
   \textit{lazy training regime}.  Notably, the network can be approximated throughout
    training by its linearization around its random weights at initialization, which is the 
    starting point for these notes.

For overparametrized neural networks, i.e. underconstrained optimization problems, 
there exist many global minimizers with zero training loss. These interpolators 
achieve very different generalization errors. Despite being trained with no 
explicit regularization that would promote `good' models, the solutions found 
in practice generalize well to test data \cite{zhang2021understanding}. A popular 
explanation for this performance postulates an `implicit regularization' from the training 
algorithm itself \cite{neyshabur2014search,neyshabur2017exploring}. In words, generalization is
 implicitly controlled by the dynamics of the optimization algorithm, which selects 
 a good minimizer, and not by adding an explicit regularization to the risk. 
 For example, in the linear regime, gradient descent selects the minimizer closest from 
 the initialization in weighted $\ell_2$ distance (see Remark \ref{rmk:implicit}). Major
  research activity has been devoted in recent years to characterizing implicit
   regularization for various algorithms, including gradient descent in matrix factorization 
   \cite{gunasekar2017implicit,li2020towards}, mirror descent \cite{gunasekar2018characterizing}, 
   gradient descent on separable data \cite{soudry2018implicit,ji2018risk}, linear neural networks 
   \cite{gunasekar2018implicit,ji2018gradient,yun2020unifying} and neural networks with homogeneous 
   activation functions \cite{lyu2019gradient,ji2020directional}. Despite these efforts,
    the precise complexity measure that is implicitly controlled by SGD in neural networks
     remains poorly understood, except in restricted settings.

\paragraph*{Chapter 2 (Linear regression under feature concentration)}$\phantom{a}$

\indent  In the case of Gaussian features, this model is also known as the 
`Gaussian design model' \cite{dobriban2018high}. The asymptotic risk of ridge regression with 
random features was computed by \cite{dicker2016ridge} for isotropic Gaussian features 
$\bx_i \sim \normal (0 , \id_p)$ in the proportional asymptotics $p,n \to \infty$ with 
$p/n \to \gamma \in (0, \infty)$. These results were generalized by \cite{dobriban2018high}
 to anisotropic features $\bx_i = \bSigma^{1/2} \bz_i$, where $\bz_i$ has independent 
 entries with bounded $12$th moment.
 \cite{hastie2022surprises} later derived similar 
  results under weaker assumptions on $(\bbeta_*, \bSigma)$. 
  
  The fact that ridge regression with isotropic random features displayed a 
  `double descent' pattern was pointed out in \cite{advani2017high}.
  However, reproducing the full phenomenology observed empirically,
  with the minimum risk achieved at large overparametrization,  
  require coefficient vectors 
  $\bbeta_*$ aligned with $\bSigma$.
  \cite{richards2021asymptotics,wu2020optimal} computed the asymptotic risk in 
  such settings. 
  \cite{hastie2022surprises} derived similar 
  results under weaker assumptions on $(\bbeta_*, \bSigma)$.
  The results in 
  \cite{hastie2022surprises} hold \textit{non-asymptotically}, with explicit error bounds,
   using the anisotropic local law proved in \cite{knowles2017anisotropic}.

The works listed above only considered the proportional parametrization scaling, i.e.,
 $C^{-1} \leq p/n \leq C$, which prohibits features coming from highly overparametrized
  ($p \gg n$) or kernel ($p= \infty$) models. A general  scaling
   (including $p=\infty$) was recently studied in \cite{cheng2022dimension}.  
   Using a novel martingale argument to prove deterministic equivalents,
   they derived non-asymptotic predictions for the test error, with nearly optimal rates 
   for the error bounds on the bias and variance terms, namely $\Tilde O (n^{-1/2})$ and 
   $\Tilde O (n^{-1})$ which match the fluctuations of respectively the local law and average
    law for the resolvent. 

 Despite the simplicity of this first model, it is connected to more complex
  models via universality. For example, it was shown in some settings that kernel regression 
  \cite{misiakiewicz2022spectrum} and random feature regression \cite{mei2019generalization} 
  have the same asymptotic test errors as some equivalent Gaussian models, where the non-linear
   feature maps are replaced by Gaussian vectors with matching first two moments.
 
 The recent interest in interpolators was prompted by the experimental results in 
 \cite{zhang2021understanding,belkin2018understand} which showed that deep neural 
 networks and kernel methods can generalize well even when they interpolate 
 \textit{noisy data}. This surprising phenomenon ---often referred to as \textit{benign overfitting}
  following \cite{bartlett2020benign}--- is at odds with the classical picture in statistics,
   where we expect interpolating noisy training data to lead to unreasonable models with 
   large test errors. Recent work has instead proved that several standard learning models 
   can indeed interpolate benignly under certain conditions, including the Nadaraya-Watson estimator \cite{belkin2019does}, kernel ridgeless regression
    \cite{liang2020just,ghorbani2021linearized} and max-margin linear classifiers 
    \cite{montanari2019generalization,shamir2022implicit}. As a testbed to understand 
    interpolation learning, much focus has been devoted to studying interpolation in
     linear regression with (sub-)Gaussian features 
     \cite{hastie2022surprises,bartlett2020benign,tsigler2020benign,muthukumar2020harmless,
     koehler2021uniform}.
 In particular, \cite{bartlett2020benign} showed sufficient conditions for consistency of 
 the minimum $\ell_2$-norm interpolating solution in terms of an effective rank, that
  were later refined in \cite{cheng2022dimension}. Further a number of studies have
   argued that benign overfitting in (kernel) ridge regression is a high dimensional 
   phenomenon \cite{rakhlin2019consistency,beaglehole2022kernel}.

\paragraph*{Chapter 3 (Kernel ridge regression)}$\phantom{a}$

\indent The test error of kernel ridge regression (KRR) was investigated in a number of
 papers in the past \cite{bartlett2005local,caponnetto2007optimal,wainwright2019high}. In particular, 
 \cite{caponnetto2007optimal} showed that KRR achieves minimax optimal rates over classes 
 of functions under certain capacity condition on the kernel operator and source 
 condition on the target function. However these earlier results have limitations. First, 
 they require a strictly positive ridge regularization and therefore do not 
 apply to interpolators. Second, they often only characterize tightly the decay rate of the error,
  i.e. as $n \to \infty$ for fixed dimension $d$. In these notes, we are instead interested
   in deriving the precise test error of KRR in high dimension where both
    $n$ and $d$ grow simultaneously. Finally, in contrasts to earlier work, 
    the results discussed in these chapters hold for a given target function $f_*$ (instead
     of worst case over a class of function).

The study of kernel models in high dimension was initiated by the seminal work 
\cite{el2010spectrum}. El Karoui analysed empirical kernel matrices of the form 
$\bK = ( h ( \< \bx_i , \bx_j \>/d ))_{i,j \leq n} \in \reals^{n \times n}$ in 
the proportional asymptotics $n \propto d$, with $\bx_i = \bSigma^{1/2} \bz_i 
$ where $\bz_i \in \reals^d$ has independent entries with bounded $4+\eps$ moment.
 \cite{el2010spectrum} showed that $\bK$ is well approximated by its linearization
  (here in the isotropic case $\bSigma = \id_d$)
\begin{equation}\label{eq:linearized_K}
\bK \approx (h(0) + h''(0)/(2d)) \boldsymbol{1} \boldsymbol{1}^\sT + h'(0) \bG + (h(1) - h(0) - h'(0)) \id_n \, ,
\end{equation}
where $\bG = (\< \bx_i , \bx_j \>/d)_{i,j \leq n}$ is the Gram matrix. This result was
later used to bound the asymptotic prediction error of KRR in the proportional regime
 \cite{liang2020just,liu2021kernel,bartlett2021deep}. In particular, KRR can learn at most a
  linear approximation to the target function in this regime. 

In order to study a more realistic scenario, where $n \gg d$ with $n,d$ large, 
several works have considered a more general \textit{polynomial asymptotic scaling},
 with $n \asymp d^\kappa$ for fixed $\kappa \in \reals_{>0}$ as $d,n \to \infty$ 
 \cite{ghorbani2021linearized,mei2022generalization,liang2020multiple,ghorbani2020neural,
 mei2021learning,misiakiewicz2021learning,xiao2022eigenspace}. In the case of covariates 
 uniformly distributed on the $d$-dimensional sphere, \cite{ghorbani2021linearized}
  generalized Eq.~\eqref{eq:linearized_K} and showed that the empirical kernel matrix 
  can be approximated by its degree-$\lfloor \kappa \rfloor$ polynomial approximation. 
  Using this decomposition, \cite{ghorbani2021linearized} showed that for $\kappa \not\in \naturals$,
   KRR fits the best degree-$\lfloor \kappa \rfloor$ polynomial approximation 
   (see Section \ref{sec:proof_sketch_KRR} in Chapter \ref{ch:random_matrix}).
    \cite{mei2022generalization} extended these results to a more general KRR 
    setting under abstract conditions on the kernel operator, 
    namely hypercontractivity of the top eigenfunctions and a spectral 
   gap property (the number of eigenvalues $\lambda_i$ such that $d^\delta/n \geq \lambda_i \geq d^{-\delta}/n$ 
 is
     smaller than $n^{1 - \delta}$ for some $\delta >0$). In this setting,
      \cite{mei2022generalization} showed that KRR effectively acts as a shrinkage operator
       with some effective regularization $\lambda^{{\rm eff}} > \lambda$. Specifically,
        denoting $(\psi_{d,j} )_{j \geq1}$ the eigenfunctions and $(\lambda_{d,j})_{j \geq 1}$ the egienvalues in nonincreasing order of the kernel operator $\Kop_d$, the KRR solution with target function $f_*$ and regularization $\lambda$ is given by
\begin{equation}\label{eq:shrinkage_app}
f_* (\bx) = \sum_{j = 1}^\infty c_j \psi_{d,j} (\bx) \qquad \Rightarrow \qquad \hat f_{\lambda} = \sum_{j = 1}^\infty \frac{\lambda_{d,j}}{\lambda_{d,j} + \frac{\lambda^{{\rm eff}}}{n}} c_j \psi_{d,j} (\bx) + \Delta \, , 
\end{equation}
where $\| \Delta \|_{L^2} = o_{d,\prob} (1)$. Hence, KRR essentially fits the projection of the target function on the top $O(n)$ eigenfunctions with $\lambda_{d,j} \gg \lambda^{{\rm eff}} /n$, and none of the components of $f_*$ with $\lambda_{d,j} \ll \lambda^{{\rm eff}} /n$. In the setting of inner product kernels on the sphere \cite{ghorbani2021linearized}, the spectral decay property is satisfied for $n \asymp d^\kappa$, $\kappa \not\in \naturals$. In this case, $\lambda^{{\rm eff}} = \Theta (1)$ and indeed all spherical harmonics of degree less or equal to $\lfloor \kappa \rfloor$ have $\lambda_{d,j} \gg \lambda^{{\rm eff}} /n$, while higher degrees spherical harmonics have $\lambda_{d,j} \ll \lambda^{{\rm eff}} /n$, and we recover that KRR fits the projection of $f_*$ onto degree-$\lfloor \kappa \rfloor$ polynomials.

If the spectral decay property is not satisfied, the estimator \eqref{eq:shrinkage_app} 
needs to be rescaled by a constant coming from random matrix theory effects, as shown in
 Chapter \ref{ch:linalg}. The case $n \asymp d^\ell$, $\ell \in \naturals$
  for data uniformly distributed on the sphere was studied recently in three concurrent papers 
  \cite{misiakiewicz2022spectrum,hu2022sharp,xiao2022precise}. They showed that the 
  contribution of degree-$\ell$ spherical harmonics to the empirical kernel matrix behaves 
  as a Wishart matrix. 
  In particular, as $n$ approaches $d^\ell / \ell! $, the condition number of 
  the Wishart matrix diverges and a  peak can appear in the risk curve.

Finally, note that asymptotics for the KRR test error were also 
heuristically derived either by conjecturing an equivalence with Gaussian feature models
 \cite{jacot2020kernel} or using statistical physics heuristics
  \cite{canatar2021spectral,cui2021generalization,cui2022error}.

\paragraph*{Chapter 4 (Random Feature model)}$\phantom{a}$

\indent The random feature (RF) model was initially introduced as a finite-rank randomized 
approximation to kernel methods \cite{balcan2006kernels,rahimi2007random}.  
 The connection between neural networks 
 and RF models was originally pointed out by \cite{neal1995bayesian,williams1996computing} and 
 has recently attracted renewed attention through the  NTK and Gaussian process descriptions 
 of wide neural networks \cite{novak2018bayesian,de2018gaussian,lee2018deep}. \cite{rahimi2007random} 
 showed that the empirical random feature kernel converges pointwise to the asymptotic kernel 
 ($N =\infty$). Note that this pointwise convergence does not provide any control 
 on the performance of RF when both $N,n$ are allowed to grow together. Subsequent works
  \cite{rahimi2008weighted,bach2017equivalence,rudi2017generalization} derived bounds
   on the approximation and generalization errors of RF models. \cite{rahimi2008weighted}
    proved a minimax upper bound $O (1/\sqrt{n} + 1/\sqrt{N})$ on the generalization error, 
    but their results are limited to Lipschitz losses and require a stringent condition
     on $\| \ba \|_{\infty}$. More recently, \cite{rudi2017generalization} studied the 
     case of the square loss and proved that for $f_*$ in the RKHS
     , $N = C \sqrt{n} \log n$ random features are sufficient to achieve the same error rate 
     $O(n^{-1/2})$ as the limiting KRR. This is in contrast to the requirement
      $N = O(n^{1+\delta})$ described in these notes, the core difference lying
       in the assumption on the target function $f_*$ (see discussion in \cite{mei2022generalization}).
        Furthermore, results in \cite{rudi2017generalization} required positive ridge 
        regularization and only characterized the minimax error rate as $n \to \infty$ for a
         fixed RKHS (fixed $d$). In these notes, we will consider instead studying the
          RF model in high dimension when $N,n,d$ all grow together.

 A number of works \cite{mei2019generalization,liao2020random,adlam2020neural,adlam2022random,mei2022generalization} 
 have studied the asymptotic risk of ridge regression with random features in high dimension.
  In particular, \cite{mei2019generalization} derives the complete analytical predictions 
  for the test error in the proportional regime $n \asymp d$ and $N \asymp d$, in the case of
   covariates and weights both uniformly distributed on the $d$-dimensional sphere.
    These asymptotics capture in detail the double descent phenomenon in this model. In particular, 
    \cite{mei2019generalization} provides precise conditions for the highly overaparametrized 
    ($N/n \to \infty$) and interpolating ($\lambda \to 0+$) solution to be optimal 
    among random feature models. The polynomial scaling for random feature models was
     first considered in \cite{ghorbani2021linearized} which focused on the approximation error 
     ($n = \infty$) with data and weights uniformly distributed on the sphere. They show that 
     RF models with $d^{\ell +\delta} \leq N \leq d^{\ell+1-\delta}$ random features can only
      approximate the projection of the target function on degree-$\ell$ polynomials. The generalization 
      error was studied in a subsequent work \cite{mei2022generalization}. They consider abstract
       conditions on the eigendecomposition of the activation function, namely hypercontractivity 
       of the top eigenfunctions, an eigenvalue gap (i.e. there exists $m \leq d^{-\delta} s$,
        where $s = \min (N,n)$, such that $\lambda_{d,m} \geq d^{\delta} / s $ and $\lambda_{d,m+1} 
        \leq d^{-\delta}/s$ for some constant $\delta >0$) and $\max (N/n , n/N) \geq d^\delta$. 
        Under these assumptions, \cite{mei2022generalization} shows that RF ridge regression 
        fits exactly the projection of the target function on the top $m \leq  \min (n, N)$
         eigenfunctions. Applying these results to covariates and weights uniformly distributed 
         on the sphere, this shows that for $d^{\ell_1 + \delta} \leq N \leq d^{\ell_1 - \delta}$, 
         $d^{\ell_2 + \delta} \leq n \leq d^{\ell_2 - \delta}$ and $\max (N/n , n/N) \geq d^\delta$, 
         RF ridge regression fits the degree-$\min (\ell_1 , \ell_2)$ polynomial approximation to $f_*$.

 Finally, in certain cases (for instance, ridge regression), the non-linear random feature model is 
 connected through universality to a simpler linear Gaussian covariate model 
 \cite{gerace2020generalisation,goldt2022gaussian,hu2022universality,montanari2022universality}.

\paragraph*{Chapter 5 (Neural Tangent model)}$\phantom{a}$

\indent In the linear regime, the optimization and generalization properties of
 neural networks depend crucially on the empirical neural tangent kernel matrix and 
 in particular, its smallest eigenvalue.  For example, \cite{du2018gradient} showed 
 that for two layer neural networks with width $N \geq Cn^6 / \lambda_{\min} (\bK_n)^4$, where 
 $\bK_n$ is the infinite-width empirical NTK matrix ($N = \infty$), the empirical kernel is 
 well conditioned throughout training and neural networks converge to zero training error.
  A number of works have studied the empirical kernel matrices arising from the neural tangent
   model \cite{fan2020spectra,liu2020linearity,oymak2020toward,zou2020gradient}.

The approximation error of the neural tangent model associated to 
two-layer neural networks was studied in \cite{ghorbani2021linearized} for covariates and 
first layer weights uniformly distributed on the $d$-dimensional sphere. They prove that for 
$d^{\ell -1 + \delta} \leq N \leq d^{\ell - \delta}$, the neural tangent model exactly fits 
the best degree-$\ell$ polynomial approximation to the target function. Namely,
 the neural tangent model has the same approximation power as the random feature model, 
 provided we match the number of parameters $p = Nd$ for NT model and $p = N$ for RF model.
  Note that this equivalence does not hold when data is anisotropic, e.g. see 
  \cite{ghorbani2020neural}. The generalization error of the neural tangent model was 
  studied in a subsequent paper \cite{montanari2022interpolation}. They first 
  show that as long as $Nd/ \log(Nd)^C \geq n$, the empirical neural tangent kernel 
  has eigenvalues bounded away from zero, and the neural tangent model can exactly 
  interpolate arbitrary labels. In the same regime,
   \cite{montanari2022interpolation} prove that ridge regression with the neural tangent model 
   is well approximated by kernel ridge regression with the infinite width kernel.

\paragraph*{Chapter 6 (Beyond the linear regime)}$\phantom{a}$

\indent  Several studies have empirically investigated the relevance of the 
linear regime to describe neural networks used in practice. They show that, in some settings,
the neural tangent model captures the behavior of SGD-trained neural networks, at least at
 the beginning of the dynamics \cite{lee2019wide,geiger2020disentangling,woodworth2020kernel}.
  Furthermore, it was shown that in some settings NTK can achieve superior performance compared to standard 
  kernels \cite{arora2019exact,li2019enhanced,shankar2020neural} and even state-of-the-art
   performance on some datasets \cite{arora2019harnessing}. However, neural networks in the 
   linear regime and kernel methods typically fall short in comparison to state-of-the-art 
   neural networks. The scenario in which weights move significantly away from their 
   random initialization is referred as \textit{rich} or \textit{feature learning regime}.

A major theoretical achievement of the nineties 
\cite{devore1989optimal,hornik1991approximation,barron1993universal,pinkus1999approximation} 
was to prove the approximation theoretic advantage of non-linear neural networks compared 
to fixed basis methods. In particular, Barron showed in his celebrated work
 \cite{barron1993universal} that two-layer neural networks can approximate, with few neurons,
  functions with fast decay of their high frequency components, while linear 
  methods need a number of parameters exponential in the dimension to approximate the
   same class of functions (in worst case). More recent works have established tighter lower bounds 
   ---that are cursed by dimensionality--- on the performance of kernel and random feature models 
   \cite{bach2017breaking,yehudai2019power,vempala2019gradient,ghorbani2021linearized}. 
   For example, \cite{yehudai2019power} proves that a super-polynomial in $d$ number of 
   random features is required to approximate a single neuron. 

While approximation theory deals with ideal representations ---which might not be 
tractable to find in practice---,  a recent line of work have sought to display theoretical
 settings where neural networks \textit{trained by gradient descent} provably outperform 
 neural tangent and kernel models \cite{bach2017breaking,woodworth2019kernel,allen2019can,allen2020backward,ghorbani2019limitations,ghorbani2020neural,fang2019over,chizat2020implicit,daniely2020learning}.
In particular, much attention has been devoted to the setting of learning ridge 
functions, i.e. functions that only depend on a small number of (unknown) relevant directions
 $f_* (\bx) = \psi ( \bU_*^\sT \bx) $ with $\bU_*^\sT \bU_* = \id_k$
\cite{chen2020towards,refinetti2021classifying,abbe2022merged,damian2022neural,mousavi2022neural,bietti2022learning,abbe2023sgd,berthier2023learning}. 
These functions are also known in the literature as single-index models in the case $k=1$ 
and multiple-index models for $k >1$. The reason for this interest is that learning ridge 
functions offer a simple setting where a clear separation exists between neural networks 
trained non-linearly and kernel methods. Indeed, linear methods are oblivious to latent
 linear structures \cite{bach2017breaking,ghorbani2021linearized,abbe2022merged}, while 
 gradient descent has the possibility to align the network weights with the sparse support. 
 The picture that emerges from these works is as follows. The complexity of learning single index models by gradient descent is driven by the \textit{information exponent} \cite{arous2021online}, which measures the strength of the correlation between a single neuron and the ridge function at initialization. This notion of information exponent was generalized to multi-index models via the `leap complexity' \cite{abbe2022merged,abbe2023sgd}.

Hence, while the lazy regime offers a setting where both the optimization and generalization 
properties of neural networks can be understood precisely, it does not capture the 
full power of deep learning. A number of approaches have been suggested to go beyond the linear regime. 
Several groups have proposed higher-order or finite-width corrections to the limiting neural tangent model \cite{huang2020dynamics,bai2019beyond,hanin2019finite,dou2020training}.
 Another approach examines other infinite-width limits for gradient-trained neural networks. A systematic analysis of the different infinite-widths limits was conducted in \cite{yang2020feature}, which characterized all non-trivial limits following an `abc-parametrization' of the scaling of learning rate, model parameters and initialization. In particular, they put forward a new \textit{maximal update parametrization ($\mu P$)} which maximizes the change in the network weights after one SGD step among all infinite-width limits.

 A popular alternative to the linear regime corresponds to two-layer neural networks trained by stochastic gradient descent in the mean-field regime \cite{chizat2018global,mei2018mean,rotskoff2018neural,sirignano2020mean,mei2019mean}. This limit coincides with the $\mu P$ limit for two-layer networks. Unlike the neural tangent approach, the evolution of network weights is now non-linear and described in terms of a Wasserstein gradient flow over the neurons' weight distribution. However, analyzing the training dynamics requires to track the evolution of a distribution and remains challenging, except in highly-symmetric settings where the PDE is effectively low-dimensional \cite{mei2018mean,abbe2023sgd,berthier2023learning}. Finally, several papers have proposed extensions of the mean-field limit to multilayer neural networks \cite{nguyen2020rigorous,lu2020mean}.

\backmatter

\bibliographystyle{amsalpha}
\bibliography{linear-nets.bbl}
\addcontentsline{toc}{section}{References}


\end{document}